%% file: cvrp.tex
\documentclass{article}
\usepackage{PRIMEarxiv}
\usepackage[utf8]{inputenc} % allow utf-8 input
\usepackage[T1]{fontenc}    % use 8-bit T1 fonts
\usepackage{hyperref}       % hyperlinks
\usepackage{url}            % simple URL typesetting
\usepackage{booktabs}       % professional-quality tables
\usepackage{amsfonts}       % blackboard math symbols
\usepackage{nicefrac}       % compact symbols for 1/2, etc.
\usepackage{microtype}      % microtypography
\usepackage{lipsum}
\usepackage{fancyhdr}       % header
\usepackage{graphicx}       % graphics
\graphicspath{{media/}}     % organize your images and other figures under media/ folder
\usepackage{tikz} % tikz library
\usepackage{amsmath}
\usepackage{pgf}
\usepackage{pgfplots}
\usepgfplotslibrary{statistics}
\usepgfplotslibrary{groupplots}
\pgfplotsset{compat=newest}
\usetikzlibrary{patterns} % tikz patterns library
\usepackage{bm}
\usepackage{float}
\usepackage{braket}
\usepackage{rotating}
\usepackage{multirow}
\usepackage{subcaption}
\usepackage{pdflscape,booktabs,graphicx}
\title{Quantum Reinforcement Learning with Transformers for the Capacitated Vehicle Routing Problem}

\author{
  Eva Andrés \\
  Department of Computer Science and Artificial Intelligence \\
  University of Granada \\
  C/ Periodista Daniel Saucedo Aranda, s/n, 18071, Granada, Spain \\
  \texttt{\ e.evaandres@go.ugr.es} \\
}

\begin{document}
\maketitle
\begin{abstract}
This paper addresses the Capacitated Vehicle Routing Problem (CVRP) by comparing classical and quantum Reinforcement Learning (RL) approaches. An Advantage Actor–Critic (A2C) agent is implemented in classical, full quantum, and hybrid variants, integrating transformer architectures to capture the relationships between vehicles, clients, and the depot through self- and cross-attention mechanisms. The experiments focus on multi-vehicle scenarios with capacity constraints, considering 20 clients and 4 vehicles, and are conducted over ten independent runs. Performance is assessed using routing distance, route compactness, and route overlap.

The results show that all three approaches are capable of learning effective routing policies. However, quantum-enhanced models outperform the classical baseline and produce more robust route organization, with the hybrid architecture achieving the best overall performance across distance, compactness, and route overlap. In addition to quantitative improvements, qualitative visualizations reveal that quantum-based models generate more structured and coherent routing solutions. These findings highlight the potential of hybrid quantum–classical reinforcement learning models for addressing complex combinatorial optimization problems such as the CVRP.
\end{abstract}

% keywords 
\keywords{
Capacitated Vehicle Routing Problem \and Multi-Vehicle Routing \and Quantum Reinforcement Learning \and Advantage Actor–Critic \and Quantum Transformers
}

\section{Introduction}

The Capacitated Vehicle Routing Problem (CVRP) is a well-known combinatorial optimization problem classified as NP-hard \cite{Karp1972}. It involves finding the optimal set of routes for a fleet of vehicles with limited capacity to serve a group of customers, minimizing the total travel cost while satisfying all capacity and demand constraints. Over the years, numerous heuristic and learning-based approaches have been proposed to address the CVRP and its multiple variants, such as the dynamic VRP, the stochastic VRP, and the multi-depot VRP \cite{toth2002vehicle, laporte2009fifty, pillac2013review}.

Recent advances in deep learning have introduced new perspectives for solving combinatorial optimization problems through sequence modeling and attention mechanisms. The seminal work \cite{vaswani2023attentionneed} inspired the development of architectures capable of learning complex dependencies among input elements, giving rise to the family of transformer-based models. Building on this idea, \cite{bello2017neuralcombinatorialoptimizationreinforcement} proposed the use of Pointer Networks to tackle routing problems, extending their framework to general combinatorial tasks such as the Traveling Salesman Problem (TSP) and the VRP. However, a key limitation of their approach lies in the assumption of a static environment, whereas in the VRP the system is inherently dynamic: once a customer is served, its demand effectively becomes zero, altering the feasible solution space at each decision step.

Subsequent research has explored attention-based Reinforcement Learning (RL) methods that combine recurrent neural networks (RNNs) with dynamic embeddings, as in Reinforcement Learning for Solving the Vehicle Routing Problem \cite{nazari2018reinforcementlearningsolvingvehicle}. In that model, an RNN decoder is coupled with an attention mechanism to compute a probability distribution over the feasible next locations based on both static and dynamic embeddings, effectively capturing the evolution of the problem state as routes are constructed.

Quantum approaches have also been proposed to tackle the VRP, including Variational Quantum Algorithms (VQA) such as the Quantum Approximate Optimization Algorithm (QAOA) \cite{farhi2014qaoa} and the Variational Quantum Eigensolver (VQE) \cite{Peruzzo2014vqe}, as well as Grover Adaptive Search (GAS) \cite{durr1999grover}. For instance, \cite{Palackal2023} investigates decomposition strategies and performance aspects of VQA-based approaches, highlighting limitations in achieving high-quality solutions under Noisy Intermediate-Scale Quantum (NISQ) constraints, and focusing exclusively on single-vehicle scenarios. Similarly, \cite{Liu2025} GAS-based methods offer a quadratic speedup and can efficiently handle linear inequality constraints without gradient-based optimization, but are limited to small-scale VRP instances and do not incorporate sequential decision-making or dynamic customer demands.

In contrast, Reinforcement Learning frameworks—particularly those leveraging attention mechanisms—can model the dynamic nature of routing problems more effectively. By combining the representational power of transformers with the optimization capabilities of quantum-enhanced RL, the present work proposes a new direction for addressing the CVRP beyond static variational formulations. Specifically, a quantum Advantage Actor-Critic (A2C) agent with self- and cross-attention mechanisms is designed to capture both static and dynamic features, such as changing customer demands after service. This approach allows the model to learn adaptive routing policies capable of handling multiple vehicles and capacity constraints, while exploiting quantum representations to improve policy expressivity and reward optimization.

The main contributions of this work are summarized as follows:
\begin{itemize}
\item
Proposal of a quantum Advantage Actor-Critic (A2C) framework with transformer-based attention for solving the CVRP.

\item Modeling of multi-vehicle and dynamically changing demand scenarios, beyond static single-vehicle formulations.

\item Comparative evaluation of classical and quantum A2C models with transformer attention, demonstrating improved reward metrics and route quality for the quantum version.

\item Experimental validation over randomly generated scenarios with multiple runs to assess robustness and generalization.

\end{itemize}

\section{Background}\label{Background}

\subsection{Mathematical Formulation of the Capacitated Vehicle Routing Problem (CVRP)}
\label{sec:CVRP}

The Capacitated Vehicle Routing Problem (CVRP) is one of the most classical and widely studied problems in logistics and combinatorial optimisation. It generalises the Travelling Salesman Problem (TSP) by introducing a fleet of vehicles with limited capacity that must serve a set of customers and return to a depot.

Over the years, many variants have emerged to incorporate real-world constraints and modern operational requirements. Among the most notable variants are:

\begin{itemize}
    \item \textbf{VRP with Time Windows (VRPTW):} Customers must be visited within specific time intervals, introducing scheduling constraints.
    \item \textbf{Multi-Depot VRP (MDVRP):} Vehicles may start and end at different depots, increasing distribution flexibility.
    \item \textbf{Multi-Objective VRP:} The objective includes not only distance minimisation but also fleet size reduction, load balancing, or service level optimisation.
    \item \textbf{Pickup and Delivery VRP (PDP/VRPPD):} Some nodes require pickups before deliveries, imposing precedence constraints.
    \item \textbf{Electric VRP (EVRP):} Routing must consider battery capacity, charging stations, and energy consumption.
    \item \textbf{Green VRP (G-VRP):} Includes environmental objectives such as carbon emissions or fuel consumption.
    \item \textbf{Hybrid/Fuel-Mix VRP:} Fleets composed of multiple vehicle technologies (ICE, EVs, hybrids).
    \item \textbf{Dynamic, Stochastic, or Real-Time VRP:} Parameters such as demand or travel times are uncertain or evolve dynamically.
    \item \textbf{Sustainable Multi-Objective VRP:} Integrates both operational efficiency and environmental or service-quality metrics.
\end{itemize}

In this work, we focus on the classical CVRP formulation and extend it with additional objectives that encode spatial coordination and service-oriented metrics (vehicle overlap, zone coherence, and customer satisfaction). Our formulation can thus be seen as a variant of the CVRP which, by inheritance, also generalises the TSP.

\paragraph{Problem Definition.}
Let $C = \{1,2,\dots,n\}$ be the set of customers, each with demand $d_i$, and let $V = \{1,2,\dots,m\}$ be a fleet of vehicles, each with capacity $Q$. All routes start and end at a central depot $D$. The goal is to minimise the total distance travelled while ensuring all demands are served without exceeding vehicle capacity.

\paragraph{Decision Variables.}
\begin{itemize}
    \item $x^{k}_{i,j}$: binary variable equal to 1 if vehicle $k$ travels from node $i$ to node $j$.
    \item $u_i$: remaining load of the vehicle after visiting customer $i$.
\end{itemize}

\paragraph{Classical Objective Function.}
The standard CVRP objective minimises total distance travelled:

\[
\min_{x} 
\sum_{k \in V} 
\sum_{i \in C \cup D} 
\sum_{j \in C \cup D} 
d_{ij} \, x_{ij}^k
\]

where $d_{ij}$ is the distance between nodes $i$ and $j$.

\paragraph{Extended Objective.}
In our formulation, we introduce three additional components to encourage spatial coordination and service quality:

\begin{itemize}
    \item \textbf{Overlap penalty}: discourages vehicles from serving customers too close to those served by others.
    \item \textbf{Zone (soft-clustering) penalty}: encourages each vehicle to remain close to its spatial anchor.
    \item \textbf{Customer service reward}: provides a positive reward when a customer is successfully served.
\end{itemize}

The full objective becomes:

\begin{equation} \label{eq:objective}
\min_{x} 
\sum_{k \in V} \sum_{i \in C \cup D} \sum_{j \in C \cup D}
\Big[
    \alpha \frac{\|p_j - p_i\|}{d_{\max}} x_{ij}^k
    + \lambda_{\text{overlap}} O_j^k x_{ij}^k
    + \lambda_{\text{zone}} Z_j^k x_{ij}^k
    - \beta S_j^k x_{ij}^k
\Big]
\end{equation}

where $p_i$ is the position of node $i$, $d_{\max}$ is the map diagonal for normalisation, and all $\lambda$ and $\alpha,\beta$ are scalar weights.

\paragraph{Additional Terms.}

\textbf{1. Overlap Penalty.}  
\[
O_j^k = 
\frac{1}{|V|-1} 
\sum_{v \neq k} 
\exp\left( -\frac{\|p^{(k)} - p^{(v)}\|}{\gamma d_{\max}} \right)
\]

\textbf{2. Zone Penalty.}
\[
Z_j^k = 
1 - \exp\left(
-\left(
\frac{\delta_k(j)}{\delta_{\min}(j)} - 1
\right)
\right)
\]
where  
\[
\delta_k(j) = \|c_j - a_k\| + \varepsilon, 
\quad 
\delta_{\min}(j) = \min_{v \in V} (\|c_j - a_v\| + \varepsilon),
\]
$c_j$ is the customer position, and $a_k$ is the centroid (anchor) of the customers already served by vehicle $k$.

\textbf{3. Incremental Reward.}

\[
r_{\text{inc}}(k, i \to j) =
\beta
- \alpha \frac{\|p_j - p_i\|}{d_{\max}}
- \lambda_{\text{overlap}} O_j^k
- \lambda_{\text{zone}} Z_j^k
\]

Each term corresponds to:
\begin{itemize}
    \item $\beta$: customer service reward.
    \item $\displaystyle \alpha \frac{\|p_j - p_i\|}{d_{\max}}$: distance penalty.
    \item $\lambda_{\text{overlap}} O_j^k$: vehicle overlap penalty.
    \item $\lambda_{\text{zone}} Z_j^k$: zone penalty encouraging spatial coherence.
\end{itemize}

Thus, the agent receives a positive reward for serving customers while being penalised for inefficient routing or spatial incoherence.

\subsection{Quantum Reinforcement Learning}\label{sec:QRL}

Quantum Reinforcement Learning (QRL) aims to extend classical reinforcement learning by exploiting fundamental properties of quantum mechanics—superposition, entanglement, and unitary evolution—to enhance representational capacity, exploration efficiency, and decision-making in high-dimensional or combinatorial environments. In recent years, QRL has attracted growing theoretical and experimental interest. Early work primarily explored conceptual adaptations of classical RL to quantum settings, often evaluated on reduced, toy environments—such as small OpenAI Gym tasks—due to hardware limitations \cite{dunjko2016quantumrlsurvey, jerbi2021variationalrl, chen2020vqcrl}. Classical value-based methods were initially extended via amplitude amplification or Grover-inspired updates \cite{skolik2021quantumenhancedoptimization, paparo2014quantumgroverrl}, while policy-based approaches employed parameterized quantum circuits (PQCs) as function approximators for policies or value functions \cite{schuld2020circuitcentric, lockwood2020vqpg}.

A recent comprehensive survey \cite{IEEE2025Survey} provides a unified view of QRL and identifies four key aspects distinguishing it from classical RL: quantum representations of states and actions, quantum-parameterized policies, intrinsic parallelism, and quantum state-update operations. These ideas have motivated a broad family of hybrid quantum–classical reinforcement learning algorithms built around the variational paradigm. In these approaches, VQCs replace or augment neural networks in the RL pipeline—either partially or fully—leading to quantum variants of Deep Q-Learning \cite{chen2020vqcrl}, quantum Actor–Critic methods \cite{jerbi2021variationalrl}, and quantum policy gradient algorithms \cite{lockwood2020vqpg}. Trainability issues such as barren plateaus \cite{cerezo2021costfunctiondependentbp} have also received significant attention, with recent work exploring quantum natural gradients, geometry-aware ansätze, and tensor-network-inspired circuit constructions to improve scalability \cite{stokes2020quantumnaturalgradient, Holmes_2022, arrasmith2021layerwise, cerezo2021variational, anschuetz2022geometry}.

More recent research has started to address key limitations of QRL, such as scalability, high dimensionality, and domain-specific structure. For instance, Efficient Dimensionality Reduction Strategies for QRL \cite{EffDimRedQRL} propose quantum-aware state-compression techniques that embed large classical state spaces into low-qubit latent representations—crucial for combinatorial optimization problems like CVRP, where agents must encode vehicle positions, remaining capacities, and dynamic customer demands. In addition, QRL for Energy-Efficiency Scenarios \cite{EnergyQRL} shows that hybrid quantum–classical agents can outperform classical baselines in different energy consumption scenarios, demonstrating practical benefits beyond simple benchmark tasks. Finally, Brain-Inspired Agents for QRL \cite{BrainQRL} introduce biologically motivated components such as structured memory, hierarchical information processing, and cognitively inspired attention mechanisms. These features enable richer inductive biases for sequential decision-making and mitigate catastrophic forgetting, while remaining compatible with VQC-based architectures.

Another active line of research focuses on multi-agent QRL, where entanglement can serve as a coordination resource and improve joint policy stability \cite{yang2022multiagentqrl}. QRL has also gained significant traction in quantum control, with RL agents learning pulse sequences and gate-synthesis strategies for real quantum hardware \cite{bukov2018rlquantumcontrol, nau2023rlcontrol}. These works highlight a natural synergy between RL and quantum systems, where the environment itself may be quantum.

Quantum-based approaches for routing problems—such as Variational Quantum Algorithms (VQAs, e.g., QAOA and VQE) and Grover Adaptive Search (GAS)—have been explored previously. While promising, these methods typically treat VRP as a static combinatorial problem, are limited to single-vehicle scenarios, and cannot handle dynamic customer demands or multi-step routing decisions \cite{Palackal2023, Liu2025}.

In contrast, QRL naturally accommodates dynamic environments, as it updates state representations and selects actions sequentially. Recent QRL research has demonstrated improved expressivity and sample efficiency when modeling tasks with combinatorial structure or high-order correlations \cite{skolik2021quantumenhancedoptimization, zhang2023quantumpointer}. Moreover, hybrid RL architectures that incorporate attention mechanisms with quantum modules have been identified as a promising direction for scalable routing policies capable of handling multi-vehicle interactions, dynamic constraints, and nonstationary demands.

However, none of the existing quantum approaches to routing incorporate transformer-based architectures, multi-head attention, or quantum attention mechanisms. Prior QRL efforts—such as the quantum attention heads introduced for SDVRP—use shallow PQC modules embedded in classical models, do not implement self-attention or cross-attention, and remain limited to static, single-vehicle settings without evolving demands. Crucially, no existing work applies Quantum Transformers, or any form of quantum multi-head self-attention or cross-attention, to CVRP or any VRP variant.

To the best of current knowledge, this is the first study to introduce a quantum Advantage Actor–Critic (A2C) agent equipped with Quantum Transformer modules—including both self-attention and cross-attention mechanisms—for solving dynamic multi-vehicle CVRP scenarios. This establishes a novel and significant bridge between transformer-based sequence modelling, quantum-enhanced reinforcement learning, and real-world combinatorial routing problems.

Overall, QRL has evolved from conceptual demonstrations to a field addressing structurally rich applications in quantum control, robotics, energy management, and combinatorial optimization. The integration of Quantum Transformers with actor–critic methods for dynamic CVRP represents a timely and natural extension of current QRL research, directly addressing limitations of previous approaches and enabling more expressive, scalable, and adaptive routing policies.

\section{Methodology}
\label{sec:methodology}

In this work, we develop \textbf{three complementary reinforcement-learning approaches} to tackle the Capacitated Vehicle Routing Problem (CVRP):  

\begin{enumerate}
    \item \textbf{Classical model (CPN)}: based on an Advantage Actor--Critic (A2C) architecture enhanced with a Pointer Network.  
    \item \textbf{Hybrid Quantum Pointer Network (HQP)}: where the policy module leverages variational quantum circuits for encoder-decoder relational processing, while input embeddings and post-processing remain classical.  
    \item \textbf{Full Quantum Pointer Network (FQP)}: in which the encoder, decoder and input embeddings are all implemented via quantum circuits, maximizing quantum relational modeling capabilities.
\end{enumerate}

All three agents share the same RL formulation and interact with a custom environment specifically designed for the multi-vehicle CVRP. At each decision step, the actor produces a separate probability distribution for each vehicle, indicating the next client that vehicle should visit—or the special ``return-to-depot'' action—while the critic estimates the value of the current state.  

Thus, the key differences among the three approaches lie in how they process the internal state: entirely classical in CPN, hybrid quantum–classical in HQP, and fully quantum in FQP. The environment, state representation, transition dynamics, and reward structure remain identical for all models.

\subsection{Environment}
\label{sec:environment}

The environment is implemented as an OpenAI Gym class (\texttt{CVRPEnvironmentGym}) and simulates a stochastic Capacitated Vehicle Routing Problem (CVRP) with $N_c$ clients and $N_v$ vehicles. Clients are randomly placed on a 2D map with integer demands; all vehicles start at a central depot with full capacity. The depot is explicitly modelled as an entity with a shared remaining capacity, enabling resource-constrained reloading operations.

\paragraph{State representation.}
At each decision step, the environment provides a flattened continuous observation vector composed of three components:
(i) the depot state, including its normalised position and remaining capacity;
(ii) the client states, where each client contributes its normalised coordinates and remaining demand; and
(iii) the vehicle states, where each vehicle contributes its normalised position and remaining capacity.
This yields a state vector of dimension $3 + 3N_c + 3N_v$, which is shared by both classical and quantum agents.

\paragraph{Action space and execution.}
The action space follows a \emph{vehicle-wise sequential decision scheme}. At every logical timestep of the agent, the policy outputs one action per vehicle,
\begin{equation}
a_t = \{ a_t^1, a_t^2, \dots, a_t^{N_v} \},
\end{equation}
where each action indicates either the next client to be served by the corresponding vehicle or a return to the depot.

These actions are executed sequentially within the environment, one vehicle at a time. After each individual vehicle transition, the shared global state is updated before executing the next vehicle action. This design preserves a standard single-step Gym interface while enabling multi-vehicle reasoning and stable on-policy learning.

\paragraph{Action feasibility.}
Action feasibility is enforced through a per-vehicle valid-action mask, which prevents selecting already served clients or clients whose demand exceeds the remaining vehicle capacity. The depot action is enabled only when no feasible client remains for the selected vehicle. Any attempt to execute an invalid action results in an immediate negative reward.

\paragraph{State transitions.}
State transitions implement the structural constraints of the CVRP. When a vehicle serves a client, it moves to the client’s location, reduces its remaining capacity, sets the client’s demand to zero, and appends the movement to its route history. Returning to the depot restores vehicle capacity subject to the depot’s shared resource constraint and incurs a distance-based penalty.

\paragraph{Reward function.}
The reward function combines multiple components designed to encourage efficient and spatially coherent routing. Serving a client yields a positive service bonus and a normalised distance penalty proportional to the travelled distance. Additional geometric regularisation discourages undesirable routing patterns, including local route crossings between vehicles.

\paragraph{Zonification.}

To encourage spatial coherence among vehicle routes, a soft zonification mechanism based on dynamic vehicle anchors is introduced.

Each vehicle $v$ maintains a dynamic anchor point computed from the set of clients it has already served. Let $P_v$ denote the set of positions of clients served by vehicle $v$. The anchor of vehicle $v$ is defined as
\[
\mathrm{Anchor}(v) =
\begin{cases}
\frac{1}{|P_v|} \sum_{x \in P_v} x, & \text{if } |P_v| > 0, \\
p_v, & \text{otherwise},
\end{cases}
\]
where $p_v$ denotes the current position of the vehicle.

Zonification is evaluated only for the currently selected client. Let $c$ be the client under consideration, with spatial position $x_c$. The distance between client $c$ and the anchor of vehicle $v$ is given by
\[
d(v,c) = \lVert x_c - \mathrm{Anchor}(v) \rVert.
\]

This distance is compared against the minimum such distance across all vehicles:
\[
d_{\mathrm{best}}(c) = \min_{u \in \{1,\dots,N_v\}} \lVert x_c - \mathrm{Anchor}(u) \rVert.
\]

A smooth relative zonification cost is then defined as
\[
\mathrm{ZoneCost}(v,c)
= 1 - \exp\!\left[-\left(\frac{d(v,c)}{d_{\mathrm{best}}(c)} - 1\right)\right],
\]
which evaluates to zero for the vehicle whose anchor is closest to the client and increases smoothly as a vehicle becomes less spatially suitable relative to others.

The zonification contribution is incorporated into the reward function as a penalisation term:
\[
r_{\text{zone}}(v,c) = - \lambda_{\text{zonification}} \, \mathrm{ZoneCost}(v,c),
\]
where $\lambda_{\text{zonification}} > 0$ is a tunable hyperparameter controlling the strength of the zonification regularisation.

This formulation softly biases vehicles toward serving geographically coherent subsets of clients while preserving flexibility, allowing cross-region assignments when they are globally beneficial.

\paragraph{Vehicle overlap.}
To discourage spatial interference between vehicle routes, a local overlap penalisation based on trajectory crossings is introduced.

When a vehicle $v$ attempts to move from its current position $p_v^{\text{prev}}$ to a new position $p_v^{\text{new}}$ (corresponding to a client assignment), the newly formed segment is tested for geometric intersection against all previously executed route segments of all vehicles. Let $\mathcal{S}$ denote the set of all existing route segments in the environment.

An overlap event is detected if the segment $(p_v^{\text{prev}}, p_v^{\text{new}})$ intersects any segment in $\mathcal{S}$, excluding contiguous segments belonging to the same vehicle.

If such an intersection occurs, a penalty proportional to the travelled distance is applied:
\[
r_{\text{overlap}}(v)
=
- \lambda_{\text{overlap}} \cdot \frac{\lVert p_v^{\text{new}} - p_v^{\text{prev}} \rVert}{d_{\max}},
\]
where $d_{\max}$ denotes the diagonal length of the map and $\lambda_{\text{overlap}} > 0$ controls the strength of the overlap regularisation.

This formulation penalises route crossings in a local and action-dependent manner, encouraging the emergence of spatially separated and non-interfering vehicle trajectories without imposing explicit global coordination constraints.

\paragraph{Episode termination.}
Episodes terminate once all client demands have been satisfied or when a maximum number of steps is reached, in which case remaining unmet demands are penalised.

Both the classical Transformer-based Pointer Network and the Quantum Pointer Network are trained within the same environment and reward formulation. This shared setup ensures a fair comparison between classical and quantum policies under identical routing dynamics and constraints.

\subsection{Classical Pointer Network (CPN)}

\begin{figure*}[!ht]
  \centering
  \includegraphics[scale=0.75]{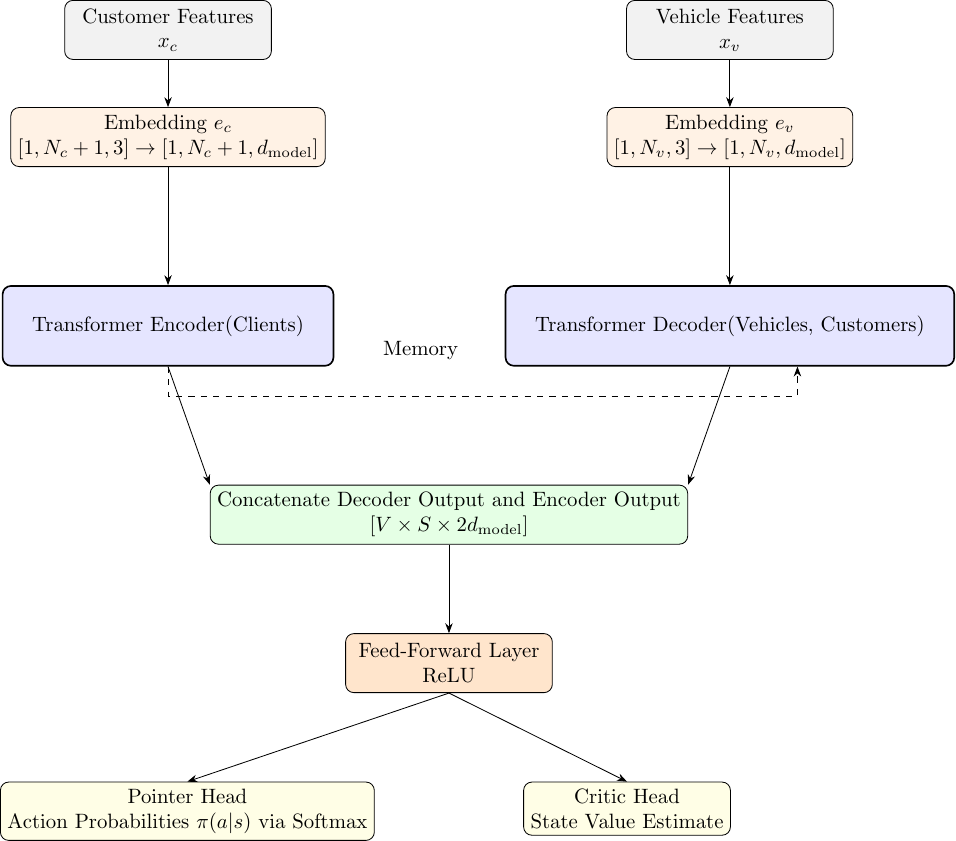}
  \caption{CPN architecture}
  \label{fig:CPN_arq}
\end{figure*}

The \textit{Classical Pointer Network} (CPN) is implemented as a Transformer-based architecture specifically adapted for the multi-vehicle CVRP. Unlike the fully quantum model, all computations are performed classically, but the architecture preserves the separation between customer and vehicle information while leveraging self-attention and cross-attention mechanisms. An overview of the proposed architecture is presented in Figure~\ref{fig:CPN_arq}.

\subsubsection*{Input Representation}\label{ClassicalInputRepresentation}
The environment state is decomposed into two complementary components. The first is the customer feature vector,
\[
x_c \in \mathbb{R}^{d_c}, \qquad d_c = 3 \,(n_{\text{clients}} + 1),
\]
which encodes the spatial coordinates $(x, y)$ and demand of all customers, including the depot. The second is the vehicle feature vector,
\[
x_v \in \mathbb{R}^{d_v}, \qquad d_v = 3\,n_{\text{vehicles}},
\]
representing vehicle-specific attributes such as position and remaining capacity. These representations are processed independently at the embedding stage to respect their distinct semantic roles in the routing problem.

\subsubsection*{Embedding Layer}\label{ClassicalEmbeddingLayer}
Each feature vector is mapped to a common latent space of dimension $d_{\text{model}}$ through separate linear layers for customers and vehicles:
\[
\text{Embedding}_c : \mathbb{R}^{3} \rightarrow \mathbb{R}^{d_{\text{model}}}, \qquad
\text{Embedding}_v : \mathbb{R}^{3} \rightarrow \mathbb{R}^{d_{\text{model}}}.
\]
This embedding projects both entities into a compatible space, enabling subsequent attention operations.

\subsubsection*{Transformer Encoder (Customer Self-Attention)}
The embedded customer vectors are processed by a Transformer encoder composed of $L$ layers, each employing multi-head self-attention and feed-forward networks. The encoder produces contextualized customer embeddings that capture pairwise correlations and global relational information among all customers. Formally, given the embedded customer matrix $X_c \in \mathbb{R}^{\text{batch} \times S \times d_{\text{model}}}$, the encoder outputs
\[
\tilde{X}_c = \text{Encoder}(X_c) \in \mathbb{R}^{\text{batch} \times S \times d_{\text{model}}}.
\]
where $S = N_c + 1$ denotes the number of service candidates (including the depot), the encoder outputs.

\subsubsection*{Transformer Decoder (Vehicle Cross-Attention)}
In parallel, the embedded vehicle vectors serve as queries to a Transformer decoder that attends to the encoded customer embeddings. This cross-attention mechanism allows each vehicle to aggregate information from all customers, producing a latent representation that combines its own state with global client context:
\[
\tilde{X}_v = \text{Decoder}(X_v, \tilde{X}_c) \in \mathbb{R}^{\text{batch} \times V \times d_{\text{model}}}.
\]
where $V = N_v$ denotes the number of vehicles.

Subsequently, to construct comprehensive vehicle--customer representations, the decoder outputs are expanded along the customer dimension and concatenated with the corresponding encoder outputs. 
\[
\text{joint} = \text{concat}(\tilde{X}_v \text{ expanded}, \tilde{X}_c\text{ expanded}) 
\in \mathbb{R}^{\text{batch} \times V \times S \times 2 d_\text{model}}.
\]

This operation results in a tensor of shape $[\text{batch}, V, S, 2 d_{\text{model}}]$ and ensures that each vehicle--customer pair integrates: (i) the contextualized customer information from the encoder, and (ii) the vehicle-specific representation that has already incorporated cross-attention over all customers. While this concatenation may appear redundant, it provides the pointer and critic heads with a complete representation for evaluating each vehicle--customer assignment.

\subsubsection*{Classical Post-Processing}
The concatenated representations are projected through a feed-forward layer with ReLU activation,
\[
h = \text{ReLU}(W \, [\tilde{X}_v \oplus \tilde{X}_c]),
\]
yielding a hidden representation of size $[\text{batch}, V, S, d_{\text{hidden}}]$. Two linear heads are then applied. The pointer head produces logits over customers for each vehicle, which are masked to enforce feasibility constraints and normalized with a softmax to yield a probability distribution:
\[
\pi(a_v = c \mid s) = \frac{\exp(\mathrm{score}_{v,c})}{\sum_{c' \in \mathcal{C}_v} \exp(\mathrm{score}_{v,c'})},
\]
where $\mathcal{C}_v$ denotes the set of feasible customers (including the depot) for vehicle $v$ given the current state $s$.

The critic head outputs a scalar state-value estimate for actor--critic training. This post-processing stage converts the latent multi-head attention representations into actionable outputs while maintaining differentiability for gradient-based optimization.

In summary, the CPN leverages self-attention among customers and cross-attention between vehicles and customers to capture the combinatorial structure of the CVRP. It produces probability distributions over feasible customer selections for each vehicle and integrates seamlessly with action-masking logic, providing a fully classical alternative to the FQP and HQP while retaining the advantages of attention-based relational reasoning.

\subsection{Hybrid Quantum Pointer Network (HQP)}

\begin{figure*}[!ht]
  \centering
 \includegraphics[scale=0.75]{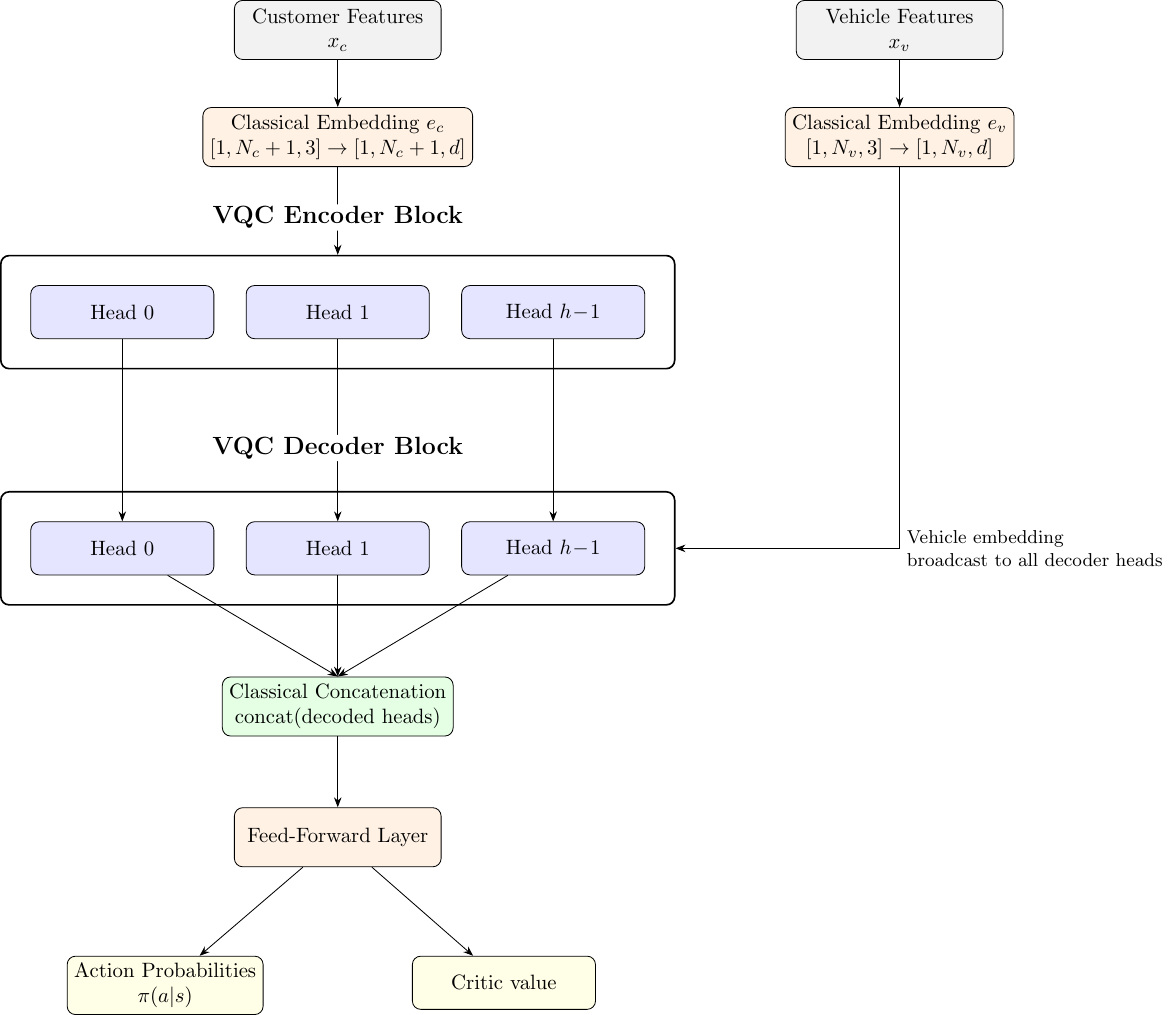}
  \caption{HQP architecture}
  \label{fig:HQP_arq}
\end{figure*}

The \textit{Hybrid Quantum Pointer Network} (HQP) is designed to leverage quantum circuits for modeling relational patterns in the Capacitated Vehicle Routing Problem (CVRP) while minimizing quantum resource usage. Unlike the fully quantum model, the HQP implements a hybrid architecture: quantum circuits are used for encoding and processing customer-vehicle interactions, whereas input embeddings, output aggregation, and decision layers remain classical. The notion of multi-head in HQP refers to parallel, independent quantum processing paths, rather than attention heads as in classical Transformers. This design ensures that each quantum head can learn distinct relational subspaces while keeping the overall qubit and entanglement requirements within practical limits. Figure \ref{fig:HQP_arq} presents an overview of the HQP architecture.

\subsubsection*{Input Representation and Embedding}

The input representation and classical embedding layers for both customers and vehicles are identical to those used in the classical pointer network (see Section \ref{ClassicalInputRepresentation} and \ref{ClassicalEmbeddingLayer}). Customer and vehicle features are linearly projected into higher-dimensional spaces before entering the quantum encoder and decoder. 

\subsubsection*{Quantum Multi-Head Encoder (Self-Attention)}

The quantum encoder consists of $h$ independent variational quantum heads. Each head receives only the full customer embedding $e_c$, and processes it through its own parameterized quantum circuit. The number of qubits per encoder head is determined by the customer embedding dimension:
\[
n_q^{\text{enc}} = \left\lceil \log_2(d_c) \right\rceil.
\]
Since no cross-head interaction occurs in the encoder, each head operates independently, reducing entanglement overhead and ensuring modularity. Correlations among customers are captured implicitly through amplitude encoding and entangling gates within each head.

\subsubsection*{Quantum Multi-Head Decoder (Encoder–Decoder Paired Heads)}

In the decoder, each quantum head follows a strict one-to-one correspondence with an encoder head. For head $i$, the decoder receives as input the complete vehicle embedding $e_v$ together with only the quantum-encoded output $z^{(i)}$ from the corresponding encoder head. The combined input is therefore:
\[
x_{\text{dec}}^{(i)} = [e_v \mid z^{(i)}],
\]
with the number of qubits for each decoder head given by
\[
n_q^{\text{dec}} = \left\lceil \log_2(d_v + d_c) \right\rceil.
\]
This "Encoder–Decoder Paired Heads" structure restricts quantum information flow, avoiding cross-head mixing, and keeping the effective Hilbert space manageable. At the same time, each vehicle interacts with its dedicated encoded customer representation.

For each decoder head, the vehicle embeddings are expanded along the customer dimension to match the shape of the encoded customer states, producing a tensor of shape $[ \text{batch}, N_v, N_c + 1, d_v + d_c ]$. This ensures that each vehicle--customer pair contains both its specific vehicle information and the latent features of the corresponding encoder head.

\subsubsection*{Classical Post-Processing}

The outputs of all decoder heads are concatenated along the feature dimension and passed through a classical feed-forward layer, yielding a hidden representation
\[
h = \text{ReLU}\Big(W \, [x_{\text{dec}}^{(1)} \mid \dots \mid x_{\text{dec}}^{(h)}]\Big).
\]
From this representation, two separate heads are applied: a pointer head producing logits over all customers for each vehicle, which are masked according to the environment's feasible actions and normalized through a softmax to yield per-vehicle probability distributions, and a critic head estimating the state-value function. This classical aggregation allows the HQP to leverage quantum relational encoding while relying on classical computation for large-scale joint decision-making.

Although no explicit self- or cross-attention mechanism is implemented, relational patterns among customers and between customers and vehicles are captured implicitly by the variational quantum circuits. Each quantum head learns a distinct subspace of interactions, analogous to attention heads in classical Transformers, but operating via amplitude encoding and entanglement rather than explicit weighted sums.

\subsection{Fully Quantum Pointer Network (FQP)}

\begin{figure*}[!ht]
  \centering
  \includegraphics[scale=0.75]{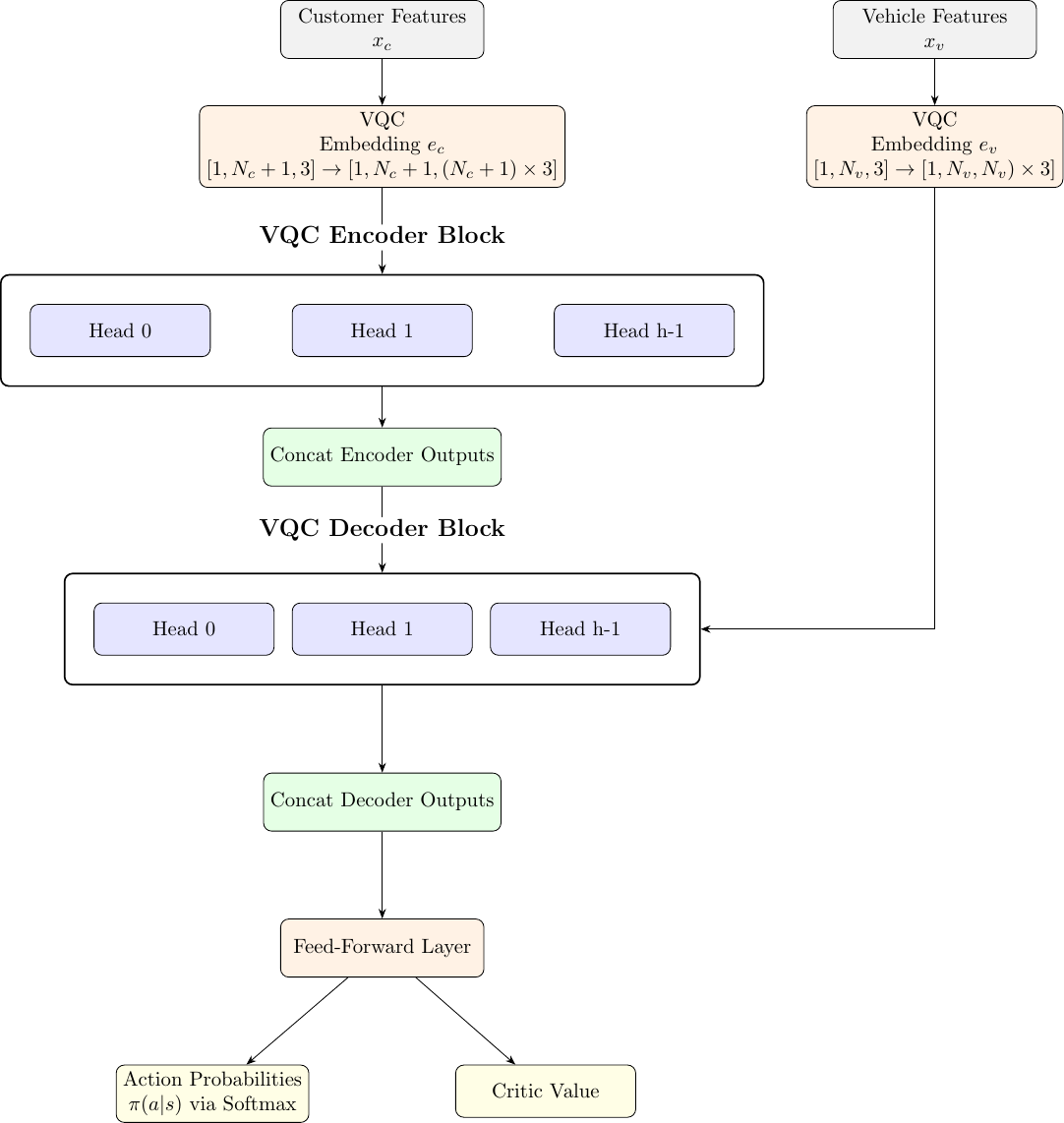}
  \caption{FQP architecture}
  \label{fig:FQP_arq}
\end{figure*}

The \textit{Fully Quantum Pointer Network} (FQP) is designed to maximize quantum 
expressivity by allowing each quantum head to process the entire customer state and by jointly exposing all encoded heads to the decoder.
To this end, the model employs amplitude embedding in both the encoder and the decoder, ensuring that all available information is compactly encoded in the quantum state.
An overview of the proposed architecture is shown in Figure~\ref{fig:FQP_arq}.

\subsubsection*{Input Representation}
The environment state is first decomposed into two complementary components:
\begin{itemize}
    \item \textbf{Customer feature vector:}
    \[
        e_c \in \mathbb{R}^{d_c}, \qquad d_c = 3\,(n_{\text{clients}} + 1),
    \]
    encoding spatial coordinates and demand-related information for all customers, including the depot.
    
    \item \textbf{Vehicle feature vector:}
    \[
        e_v \in \mathbb{R}^{d_v}, \qquad d_v = 3\,n_{\text{vehicles}},
    \]
    representing vehicle-specific attributes such as position and remaining capacity.
\end{itemize}

These two representations are treated independently in the initial embedding stage, reflecting their distinct semantic roles in the routing problem.

\subsubsection*{Quantum Embedding Layer}
Next, both customer and vehicle features are embedded into quantum states using amplitude encoding.
This choice allows all classical information to be compactly stored within the amplitudes of a quantum state.

\begin{itemize}
    \item \textbf{Customer embedding.}
    The customer feature vector is encoded as
    \[
        \text{AmplitudeEmbedding}(e_c) \;\longrightarrow\; \ket{\psi_c},
    \]
    where the required number of qubits is
    \[
        n_q^{\text{enc}} = \left\lceil \log_2(d_c) \right\rceil.
    \]

    \item \textbf{Vehicle embedding.}
    Similarly, the vehicle feature vector is encoded as
    \[
        \text{AmplitudeEmbedding}(e_v) \;\longrightarrow\; \ket{\psi_v}.
    \]
\end{itemize}

By construction, this separation enables customer and vehicle information to be processed independently before any interaction is introduced.

\subsubsection*{Quantum Multi-Head Encoder (Self-Attention over Customers)}

Subsequently, the embedded customer state is processed by a quantum multi-head encoder composed of $h$ parallel quantum heads.
Importantly, all encoder heads receive the same full customer embedding $e_c$, rather than disjoint subsets of customers.

Each encoder head performs the transformation
\[
z^{(i)} = Q_{\text{enc}}^{(i)}\!\left( \ket{\psi_c} \right),
\qquad i = 1, \ldots, h,
\]
where $Q_{\text{enc}}^{(i)}$ denotes a variational quantum circuit acting on
\[
n_q^{\text{enc}} = \left\lceil \log_2(d_c) \right\rceil
\]
qubits.

This stage can be interpreted as a form of \emph{quantum self-attention}.
Since each head operates on the complete customer state, all customers are observed simultaneously.
Correlations and dependencies between customers are therefore encoded directly into the quantum state, while different heads are able to learn complementary relational patterns.

The outputs of the $h$ encoder heads are concatenated into a single global customer representation,
\[
z = \big[ z^{(1)} \mid z^{(2)} \mid \cdots \mid z^{(h)} \big]
\;\in\; \mathbb{R}^{h \cdot d_c}.
\]

\subsubsection*{Quantum Multi-Head Decoder (Cross-Attention between Vehicles and Customers)}
The decoder explicitly models interactions between vehicles and the entire customer representation.
For each vehicle, the decoder receives:
    \[
    x_{dec} =e_v \oplus z,
\]
with dimensionality
\[
d_{\text{dec}} = d_v + h \cdot d_c.
\]
This construction is applied independently for each vehicle, producing a vehicle-specific decoder input that combines the global customer context with the current vehicle state. This vector is then amplitude-encoded into a quantum state:
\[
\text{AmplitudeEmbedding}(x_{\text{dec}}) \;\longrightarrow\; \ket{\phi}.
\]

The number of qubits used in the decoder is fixed across all heads and is given by
\[
n_q^{\text{dec}} = \left\lceil \log_2(d_{\text{dec}}) \right\rceil.
\]

Each decoder head operates on the concatenation of the outputs from all encoder heads, rather than being restricted to a single corresponding encoder representation. This design enables cross-head interaction directly at the quantum level, allowing entanglement to form between features from different encoder heads. As a result, the decoder implements a form of quantum cross-attention, in which vehicle states attend to the complete multi-head customer representation, jointly capturing customer--vehicle dependencies and ensuring that no decoder head is isolated from the global customer context.

While this fully quantum design significantly increases expressivity, it also introduces potential trade-offs. Operating in a higher-dimensional and more entangled quantum space may lead to optimization challenges, such as barren plateaus or slower convergence during training. In contrast, hybrid architectures in which each decoder head attends only to its corresponding encoder output exhibit a more localized information flow. These models offer a less global representation but may benefit from improved training stability and faster convergence due to their reduced expressivity.

\subsubsection*{Classical Post-Processing}
The outputs of all decoder heads are concatenated and subsequently projected through a classical feed-forward layer, producing a hidden representation 
\[
h = \text{ReLU}(W \, z_{\text{dec}}).
\] 
From this hidden representation, two separate linear heads are applied. The first, the pointer head, generates a probability distribution over all customers for each vehicle, with action masking and a softmax applied to ensure that only feasible actions are selected. The second, the critic head, estimates the state value required for actor--critic training. This classical post-processing stage effectively maps the high-dimensional quantum representation into actionable outputs while preserving differentiability for gradient-based optimization.

\section{Experimentation} \label{experimentation}
\subsection{Problem Statement} \label{ProblemStatement}

For the experimental evaluation, we consider a multi-vehicle Capacitated Vehicle Routing Problem (CVRP), as introduced in Section~\ref{sec:CVRP}. In our experiments, we focus on an environment with 
20 clients and 4 vehicles. Each vehicle has a maximum capacity, and all routes start and end at a central depot. The agent's objective is to serve all customer demands efficiently while minimising total travel distance and respecting capacity constraints.

In our reinforcement learning setting, the environment encodes additional spatial and service-oriented considerations, including overlap penalties between vehicles, soft zone coherence, and incremental rewards for customer service. Vehicles can move either to a client or return to the depot to reload, with valid actions determined by remaining vehicle capacity and outstanding customer demands.

The state observed by the agent consists of the normalized positions and capacities of the depot, all customers, and all vehicles. Actions are represented as discrete vehicle-client assignments, where the depot is treated as a special client index. Rewards combine distance penalties, overlap and zonification costs, and bonuses for serving nearby customers or improving local routing efficiency. 

This formulation enables a controlled and direct comparison between classical, hybrid, and fully quantum Pointer Network-based agents under identical environment dynamics.

\subsection{Experimental Settings}\label{experimentalsettings}

To evaluate the proposed approaches, we conducted experiments using three Pointer Network-based agents: classical, hybrid, and fully quantum. All agents were trained using an Advantage Actor--Critic (A2C) algorithm in the same custom CVRP environment, sharing identical dynamics, state representations, and reward functions.

The classical model was trained for 1000 episodes, while the hybrid and fully quantum models were trained for 500 episodes each. This choice reflects empirical observations that the classical Transformer-based model required a larger number of training iterations to reach stable convergence, whereas the hybrid and fully quantum models exhibited faster convergence in terms of policy stability and reward saturation. Every 50 episodes, agents were evaluated using a deterministic policy. To ensure statistical robustness, each experiment was repeated 10 times with different random seeds.

For each agent, we measured: (i) total travelled distance, (ii) spatial compactness of the client sets served by each vehicle, and (iii) vehicle overlap, measured through route crossings. In addition, the routes followed by vehicles in each test episode were stored as plots for qualitative inspection. All three models share the same environment configuration, including the number of clients, the number of vehicles, the maximum demand per client, and the maximum vehicle capacity. The environment hyperparameters were set as follows: $\alpha = 1.0$ for the distance penalty, $\beta = 1.0$ for the service reward, $\lambda_\text{overlap} = 1.0$ to penalize vehicle overlap, $\lambda_\text{zone} = 1.0$ to encourage spatial coherence, \texttt{neighbor\_threshold} = 0.2, and \texttt{nearby\_client\_bonus} = 5.0. 

AdamW was used as the optimizer for all models to ensure stable and consistent optimization across classical, hybrid, and fully quantum architectures. Model configurations and additional hyperparameters are summarized in Table~\ref{tab:model_config}.

\begin{sidewaystable}[!ht]
\centering
\caption{Architectural and training configuration of the three evaluated models.}
\label{tab:model_config}
\resizebox{\textwidth}{!}{
\begin{tabular}{lccc}
\toprule
\textbf{Component} & \textbf{Classical PN} & \textbf{Hybrid PN} & \textbf{Full Quantum PN} \\
\midrule
\multicolumn{4}{l}{\textit{Input and Embeddings}} \\
Customer embedding        & $Linear(3, 32)$ & $Linear(3, 32)$ & $VQCEmbedding(log_2((Nc+1)*3), L_{emb})$ \\
Vehicle embedding         & $Linear(3, 32)$ & $Linear(3, 32)$ & $VQCEmbedding(log_2(Nv*3), L_{emb})$ \\
Embedding dimension & $d_{\text{model}}=32$ & Input dimension (amplitude encoded) & Input dimension (amplitude encoded)\\
\midrule
\multicolumn{4}{l}{\textit{Encoder / Decoder}} \\
Encoder                   & $TransformerEncoderLayer(d_{\text{model}},num_{\text{heads}},
                                                   hidden_{\text{dim}})$ & $VQCEncoder(n_{qenc}, L_q, H)$&  \\
                          & $TransformerEncoder(encoder_{\text{layer}}, l_c)$ & & \\                         
Decoder                   & $TransformerDecoderLayer(d_{\text{model}}, num_{\text{heads}}, hidden_{\text{dim}})$ & $VQCDecoder(n_{qdec}, L_q, H)$ & VQC Decoder \\
                          & $TransformerDecoder(decoder_{\text{layer}}, l_c)$& &              \\ 
Encoder heads             & $4$ & $4$ & $4$ \\
Decoder heads             & $4$ & $4$ & $4$ \\
Number of Layers          & $l_c=7$ & $L_{q}=1$ & $L_{emb}=1, L_{enc}=1$     \\ 

\midrule
\multicolumn{4}{l}{\textit{Quantum Configuration}} \\
Number of qubits          & -- & $n_{qenc}=6,n_{qdec}=7$ & $n_{qemb_c}=6, n_{qemb_v}=4, n_{qenc}=6,n_{qdec}=7$ \\
Quantum encoding          & -- & $Amplitude-encoded$ & $Amplitude-encoded$ \\
Quantum ansatz            & -- & $RX-RY-RZ + CZ-ring$ & $RX-RY-RZ + CZ-ring$ \\

\midrule
\multicolumn{4}{l}{\textit{RL and Optimisation}} \\
Optimizer                 & AdamW & AdamW & AdamW \\
Learning rate             & $1e-5$ & $1e-6$ & $1e{-6}$ \\
Discount factor $\gamma$  & 0.97 & 0.97 & 0.97 \\
Entropy coefficient       & $0.1-0.01$ & $0.5-0.03$ & $0.5-0.03$ \\
Number of episodes        & $1000$ & $500$ & $500$ \\

\bottomrule
\end{tabular}}
\end{sidewaystable}

\subsection{Results}\label{Results}
We evaluate the performance of the three models—Classical Pointer Network (CPN), Hybrid Quantum Pointer Network (HQP), and Full Quantum Pointer Network (FQP)—on the Capacitated Vehicle Routing Problem (CVRP). All experiments are conducted in an environment with 20 clients and 4 vehicles, which provides a controlled setting to compare model behaviors while maintaining computational tractability.

The following metrics are used:

\begin{itemize}
    \item \textbf{Total Distance:} The sum of all routes in a solution, reflecting the efficiency of the routing.
    \item \textbf{Compactness:} A measure of how geographically concentrated the routes are, indicating how well clients are grouped per vehicle.
    \item \textbf{Overlap:} The degree of route intersection among vehicles, quantifying conflicts or inefficiencies in the solution.
\end{itemize}

In addition to quantitative evaluation, we visualize the routing solutions on selected seeds for all three models. These graphs illustrate how each model constructs routes, showing the evolution of the solution and the strategies used to resolve CVRP in specific instances. This qualitative analysis helps interpret differences in metric performance and highlights patterns not captured by numerical metrics alone.

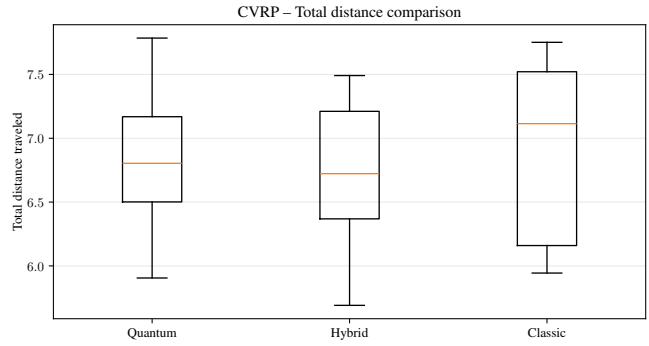
\begin{figure}[H]
        \centering
            %\resizebox{1.0\textwidth}{!}{\input{figures/DistanceBoxPlot.pgf}}
            \scalebox{0.5}{\input{images/DistanceBoxPlot.pgf}}
\caption{Boxplots illustrating the distribution of average distance obtained from quantum (HPN, QPN) and from classical model (CPN).}
\label{fig:BoxplotDist}
\end{figure}

\begin{figure}[H]
        \centering
            %\resizebox{1.0\textwidth}{!}{\input{figures/CompactationBoxPlot.pgf}}
            \scalebox{0.5}{\input{images/CompactationBoxPlot.pgf}}
\caption{Boxplots illustrating the distribution of average compactation obtained from quantum (HPN, QPN) and from classical model (CPN).}
\label{fig:BoxplotCompactation}
\end{figure}
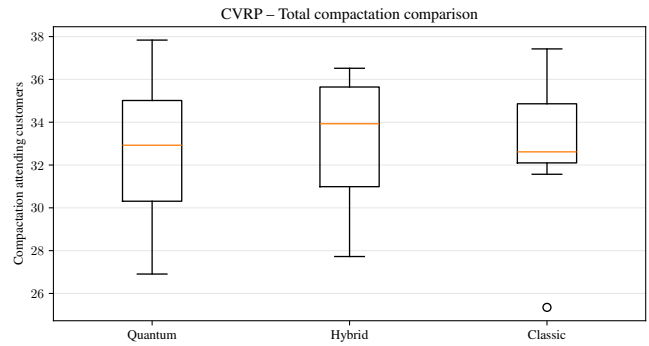

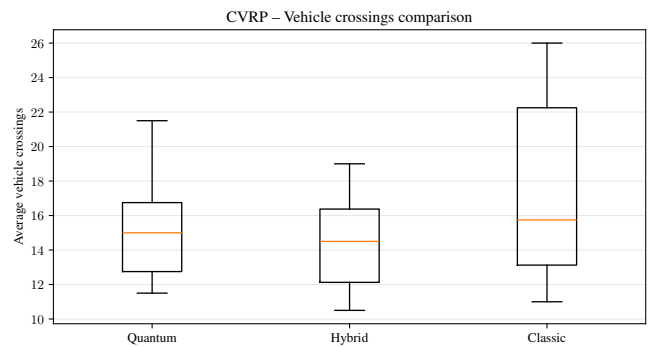
\begin{figure}[H]
        \centering
            %\resizebox{1.0\textwidth}{!}{\input{figures/DistanceBoxPlot.pgf}}
            \scalebox{0.5}{\input{images/CrossingBoxPlot.pgf}}
\caption{Boxplots illustrating the distribution of average overlap obtained from quantum (HPN, QPN) and from classical model (CPN).}
\label{fig:BoxplotCrossing}
\end{figure}

Table \ref{tab:metrics_results} presents the values of the three metrics (Total Distance, Compactness, Overlap) for each model across the evaluated instances. To complement the table, Figure \ref{fig:BoxplotDist}, \ref{fig:BoxplotCompactation}, \ref{fig:BoxplotCrossing} shows boxplots for each metric, comparing the distributions between the three models. This combination of table and boxplots provides both summary statistics and insight into the variability and robustness of each approach.

Based on the results reported in Table~\ref{tab:metrics_results}, the Hybrid Pointer Network consistently achieves the best performance in terms of routing distance. It obtains the lowest average distance, as well as the best minimum and maximum distance values, indicating not only shorter routes on average but also more robust behavior in both best- and worst-case scenarios compared to the Classical and Full Quantum models.

Regarding route compactness (zonification), the results are more balanced across models. The Full Quantum Pointer Network achieves the best average compactness, suggesting a stronger tendency to generate spatially coherent routes on average. In contrast, the Classical Pointer Network obtains the best minimum compactness value, while the Hybrid model shows the best worst-case compactness, highlighting complementary strengths across architectures depending on the evaluation criterion.

Finally, in terms of route overlap (crossings), the Hybrid Pointer Network clearly outperforms the other approaches across all statistics. It achieves the lowest average number of crossings, as well as the best minimum and maximum values, indicating a consistent reduction in route intersections and a more structured allocation of delivery paths.

Figures~\ref{fig:BoxplotDist}–\ref{fig:BoxplotCrossing} illustrate the distribution of the results obtained by the Classical, Full Quantum, and Hybrid Pointer Networks for distance, compactness, and route overlap, respectively.

For total distance (Figure~\ref{fig:BoxplotDist}), the Hybrid Pointer Network presents a lower median compared to the Classical model, with the distribution shifted towards shorter routes. The Full Quantum model also improves over the Classical baseline, although with slightly higher variability. Overall, the Hybrid approach shows a favorable trade-off between median performance and dispersion, indicating more consistent distance optimization across instances.

Regarding route compactness (Figure~\ref{fig:BoxplotCompactation}), the three models exhibit largely overlapping distributions, confirming that none of the approaches clearly dominates this metric. The Classical model shows the lowest median compactness; however, it also presents an outlier corresponding to a notably poor compactness case, indicating occasional instability. The Full Quantum model reaches lower minimum values, as reflected by its lower whisker, while the Hybrid model displays an intermediate median and a more balanced distribution, suggesting more consistent behavior across instances.

In terms of route overlap (Figure~\ref{fig:BoxplotCrossing}), the Hybrid Pointer Network clearly outperforms the other approaches. It achieves the lowest median number of crossings and a reduced interquartile range, indicating greater stability and more structured routing solutions. In contrast, the Classical model exhibits higher variability and more extreme values, while the Full Quantum model provides moderate improvements without reaching the consistency observed in the Hybrid approach.

\begin{figure}[H]
  \centering
  \begin{subfigure}{0.48\textwidth}
    \centering
    \includegraphics[width=\linewidth]{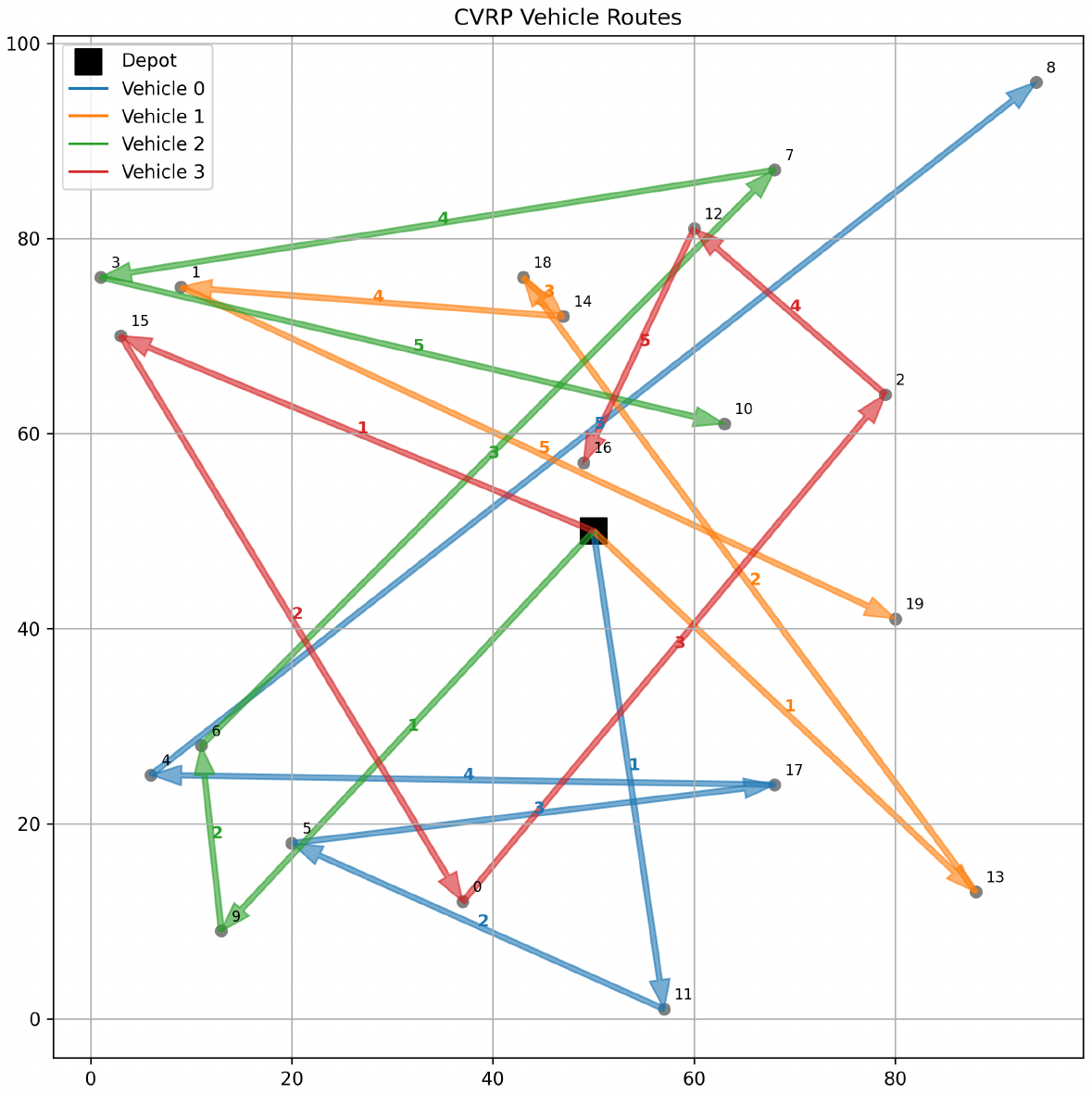}
    \caption{Iteration 50 (mid-training)}
    \label{fig:CPN_grafo1}
  \end{subfigure}
  \hfill
  \begin{subfigure}{0.48\textwidth}
    \centering
    \includegraphics[width=\linewidth]{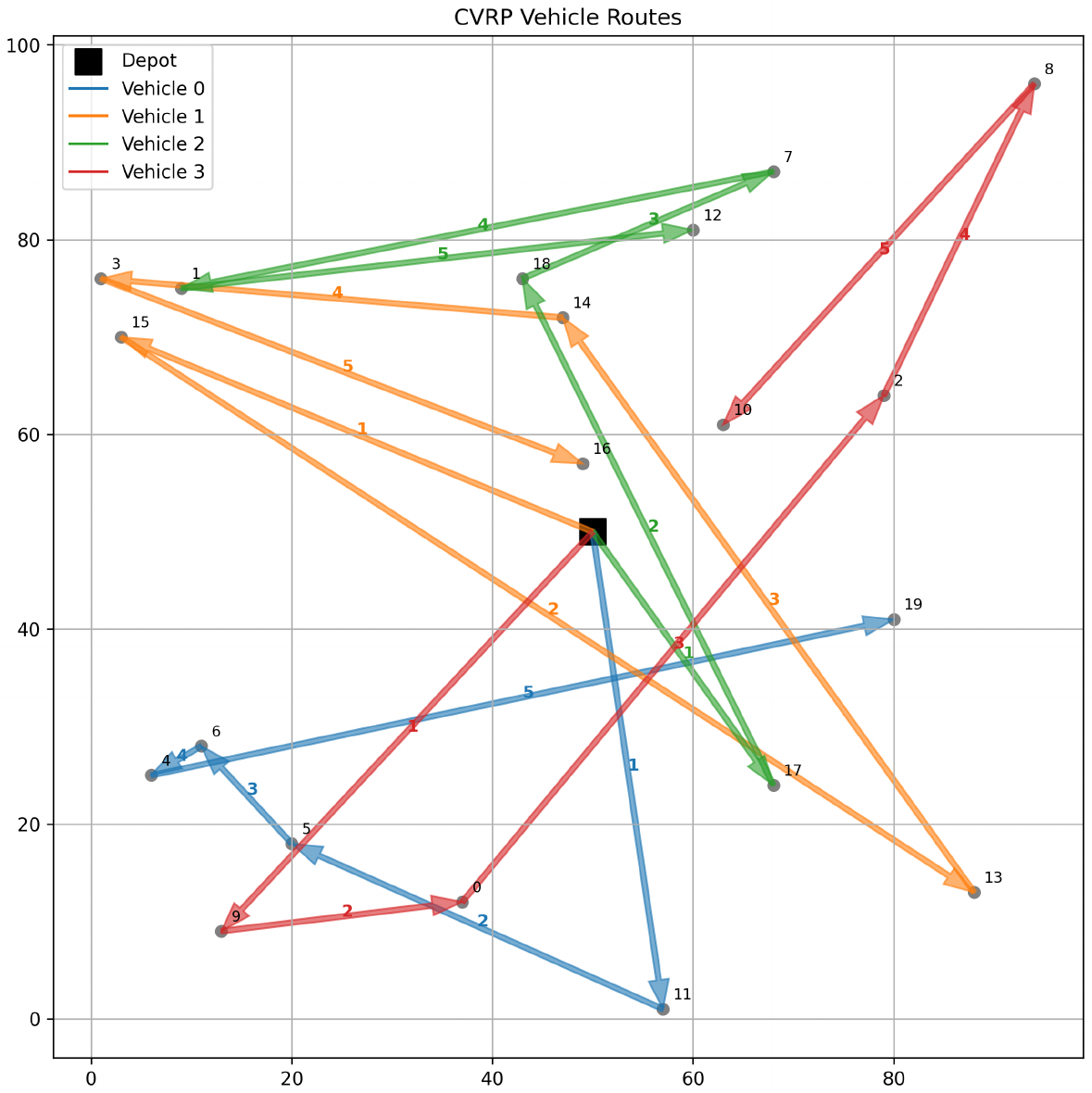}
    \caption{Final iteration (after convergence)}
    \label{fig:CPN_grafo2}
  \end{subfigure}
\caption{Evolution of the CPN routing graph for a CVRP instance with 20 clients and 4 vehicles, comparing an intermediate solution at iteration 50 with the final solution at iteration 500 for seed 1. The final solution shows an improved spatial organization of routes compared to the intermediate stage, reflecting the learning process, although service regions remain partially overlapping and some route crossings persist.}
  \label{fig:CPN_grafos}
\end{figure}

\begin{figure}[H]
  \centering
  \begin{subfigure}{0.48\textwidth}
    \centering
    \includegraphics[width=\linewidth]{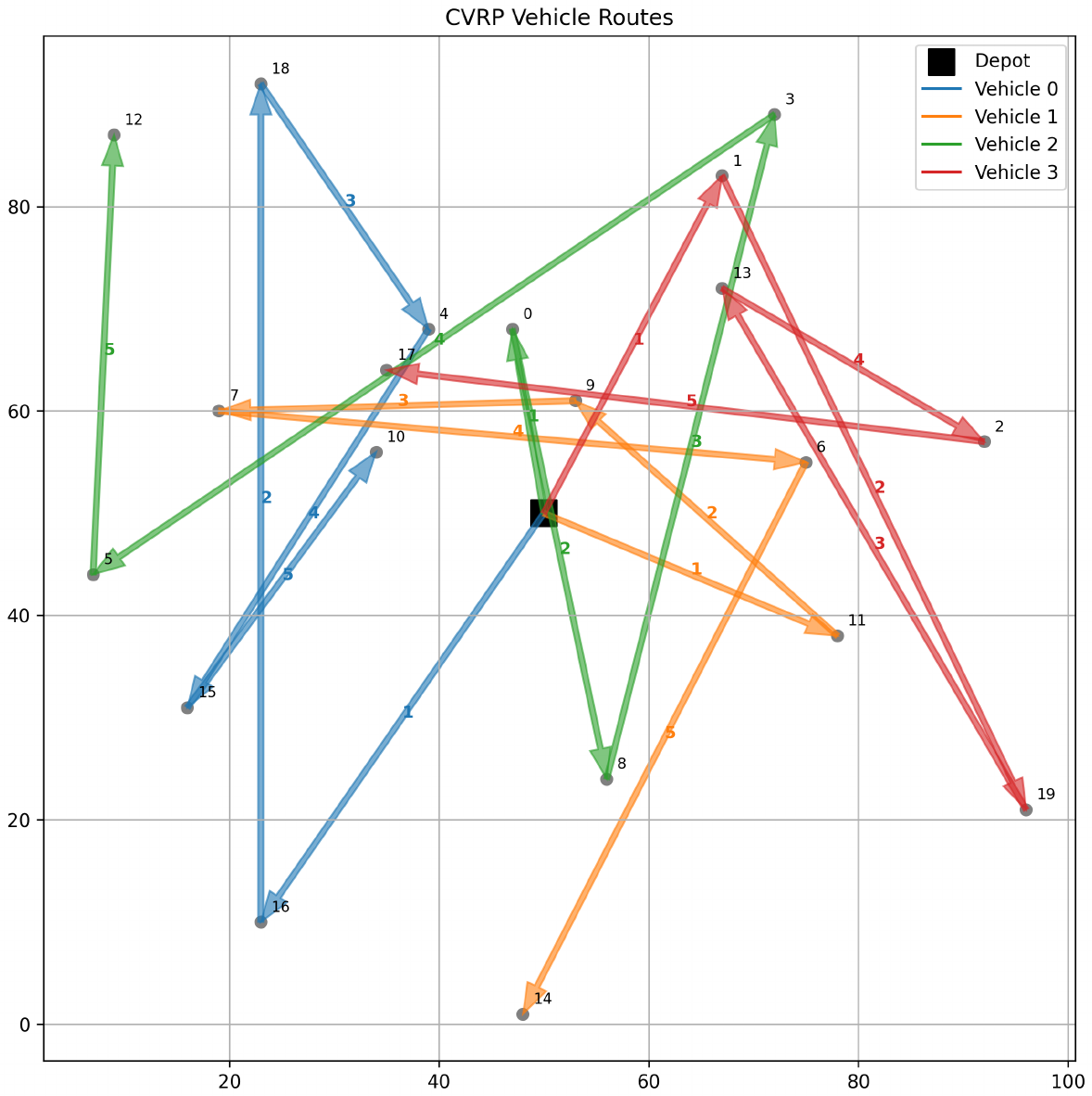}
    \caption{Iteration 50 (mid-training)}
    \label{fig:FQP_grafo1}
  \end{subfigure}
  \hfill
  \begin{subfigure}{0.48\textwidth}
    \centering
    \includegraphics[width=\linewidth]{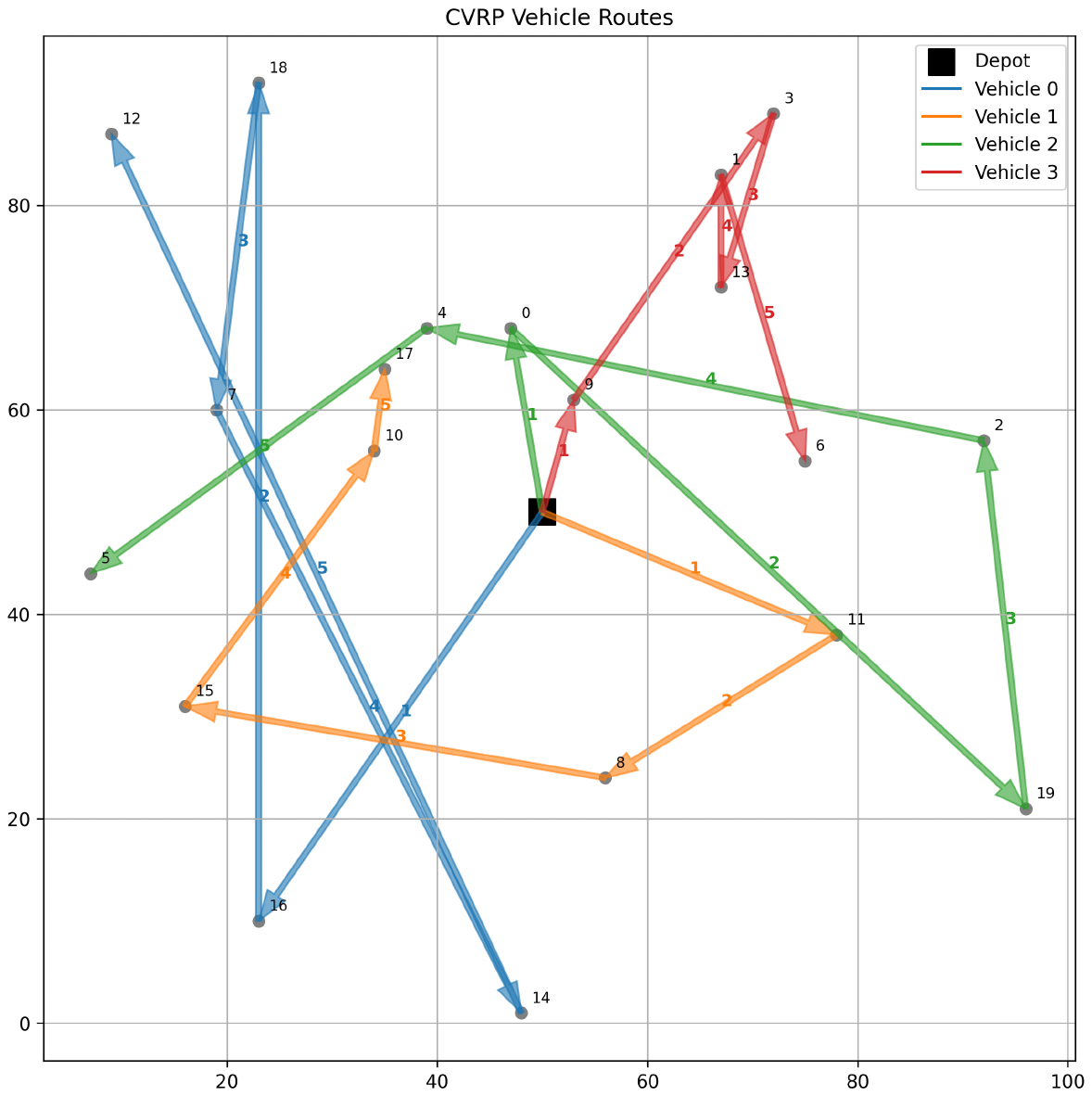}
    \caption{Final iteration (after convergence)}
    \label{fig:FQP_grafo2}
  \end{subfigure}

  \caption{Evolution of the FQP routing graph for a CVRP instance with 20 clients and 4 vehicles, comparing an intermediate solution at iteration 50 with the final solution at iteration 500 for seed 7. In the final solution, an emergent spatial organization into four regions can be observed, with vehicles predominantly serving left, central, northern, and southern areas. }
  \label{fig:FQP_grafos}
\end{figure}

\begin{figure}[H]
  \centering
  \begin{subfigure}{0.48\textwidth}
    \centering
    \includegraphics[width=\linewidth]{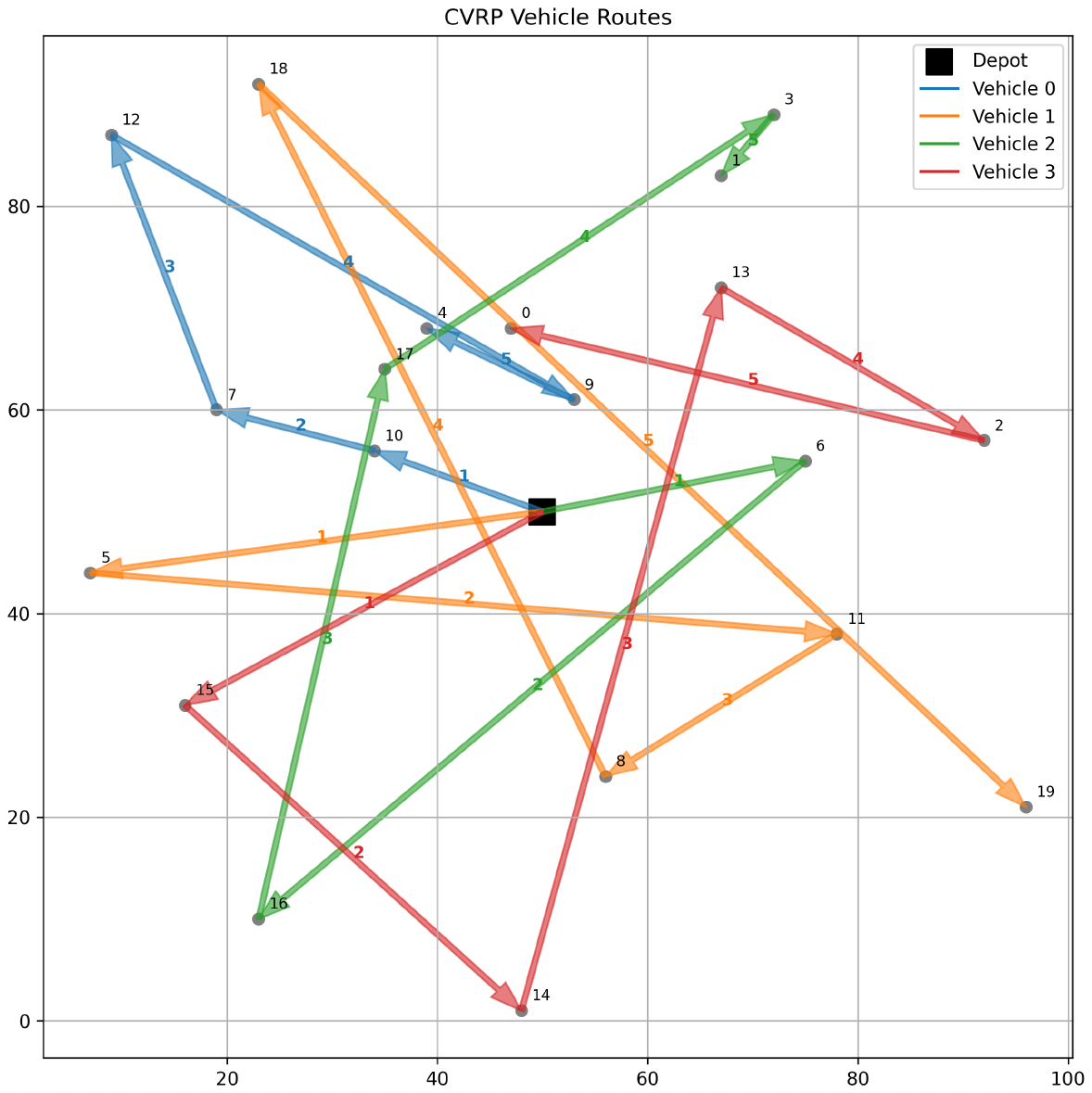}
    \caption{Iteration 50 (mid-training)}
    \label{fig:HPN_grafo1}
  \end{subfigure}
  \hfill
  \begin{subfigure}{0.48\textwidth}
    \centering
    \includegraphics[width=\linewidth]{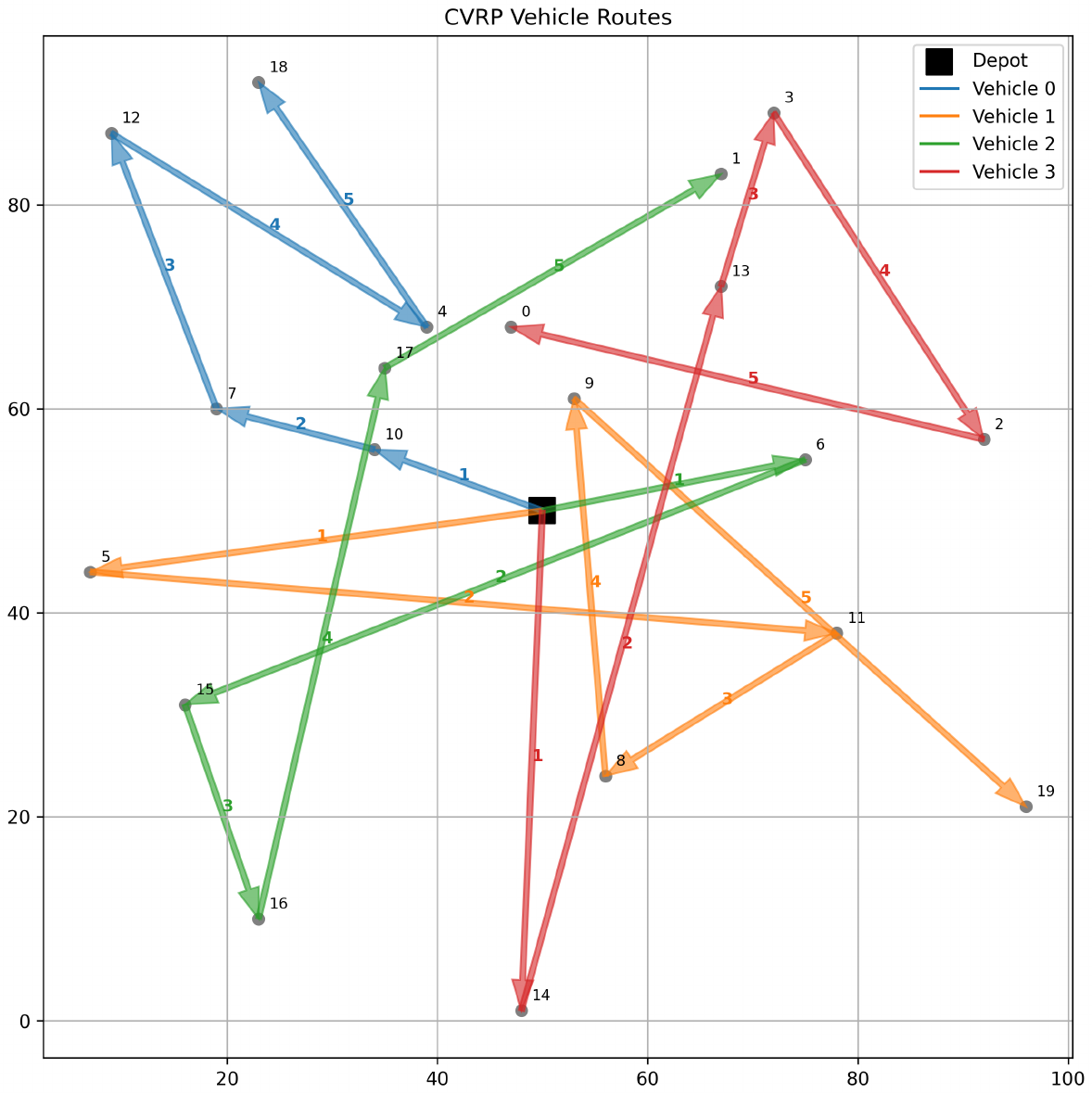}
    \caption{Final iteration (after convergence)}
    \label{fig:HPN_grafo2}
  \end{subfigure}

  \caption{HPN routing graph evolution for a CVRP instance with 20 clients and 4 vehicles, from iteration 50 to iteration 500 (seed 7), showing increased route organization into four zones and reduced crossings and distance.}
  \label{fig:HPN_grafos}
\end{figure}

Finally, in Figures \ref{fig:CPN_grafos}, \ref{fig:FQP_grafos} y \ref{fig:HPN_grafos} examples of route solutions for three different seeds are presented, comparing the initial and final iterations. For each of the three models, seeds were selected that yielded the best route organization. The evolution of the learning process can be observed in these figures, complementing the performance metrics reported above.

\begin{table}[H]
\centering
\begin{tabular}{llccc}
\toprule
\textbf{Metric} & \textbf{Statistic} & \textbf{Classical PN} & \textbf{Quantum PN} & \textbf{Hybrid PN} \\
\midrule
\multirow{3}{*}{Distance} 
 & Avg & 6.89 & 6.80 & 6.76 \\
 & Min & 5.94 & 5.91 & 5.69 \\
 & Max & 7.75 & 7.78 & 7.49 \\
\midrule
\multirow{3}{*}{Compactness} 
 & Avg & 32.86 & 32.77 & 33.03 \\
 & Min & 25.35 & 26.91 & 27.73 \\
 & Max & 37.42 & 37.84 & 36.52 \\
\midrule
\multirow{3}{*}{Overlap} 
 & Avg & 17.35 & 15.15 & 14.50 \\
 & Min & 11.00 & 11.50 & 10.50 \\
 & Max & 26.00 & 21.50 & 19.00 \\
\bottomrule
\end{tabular}
\caption{Performance comparison for distance, compactness, and route crossings in the CVRP environment with 20 clients and 4 vehicles.}
\label{tab:metrics_results}
\end{table}

\section{Discussion}

The results highlight clear differences among the three models in solving the CVRP. The Hybrid Quantum Pointer Network consistently achieves the best performance in total routing distance and route overlap, indicating efficient and well-structured solutions (\ref{fig:BoxplotDist}, \ref{fig:BoxplotCompactation}, \ref{fig:BoxplotCrossing}).

The Full Quantum Pointer Network demonstrates the highest route compactness, producing geographically more coherent routes than the Classical Pointer Network. This qualitative improvement is evident in Figures \ref{fig:CPN_grafos}, \ref{fig:FQP_grafos}, and \ref{fig:HPN_grafos}, where quantum models show better-organized routes from the initial to the final iteration.

The Classical Pointer Network, while competitive in certain metrics, exhibits higher variability and less consistent route organization. A drawback of the quantum models is the significantly longer training times required when using simulators, which may limit scalability to very large instances.

Overall, combining quantitative metrics with visualizations shows that quantum-enhanced models not only optimize distances but also generate more interpretable and spatially coherent routing patterns. These findings suggest that integrating quantum processing into pointer networks can improve both efficiency and route organization, although practical application depends on balancing performance gains with computational cost.

\section{Conclutions and Future Work}

This study demonstrates the potential of integrating Quantum Transformer modules within an Advantage Actor–Critic framework for dynamic multi-vehicle CVRP scenarios. Our results show that quantum-enhanced architectures can improve route efficiency, compactness, and organization, providing both quantitative and qualitative advantages over classical approaches.

A key research direction involves geometry-aware and structure-preserving quantum architectures. Variational circuits adapted to the symmetry, topology, and dynamic structure of routing problems could constrain quantum state evolution to meaningful, low-dimensional manifolds, improving trainability and mitigating barren plateaus \cite{anschuetz2022geometry, volkoff2023dynamic, zhang2023laqa}. Techniques such as layerwise ansatz growth, dynamic circuit morphing, and adaptive architecture search allow circuits to evolve according to task complexity, enabling more efficient exploration of the exponentially large CVRP solution space.

Insights from neuroscience suggest strategies for efficient exploration and decision-making in quantum reinforcement learning (QRL). Low-dimensional latent dynamics, structured trajectory selection, and selective amplification—as observed in cortical networks \cite{Genkin2025} and glial modulation \cite{HODEBOURG20252399}—can inspire mechanisms for trajectory pruning, structured exploration, and context-dependent reasoning. Integrating foundation-style cognitive architectures \cite{liu2025advanceschallengesfoundationagents, rosenbloom2025mappingneuraltheoriesconsciousness} into QRL may enhance working memory, attention, and long-horizon coordination, providing robustness against catastrophic forgetting and enabling more adaptive multi-vehicle decision-making.

Another promising avenue is Quantum Spiking Neural Networks (QSNNs), which introduce temporal dynamics and spike-like events that guide quantum states along structured trajectories, analogous to neuronal tuning functions. Combined with geometry-aware circuits, these constraints could help quantum agents identify promising solutions without exploring irrelevant regions of the combinatorial space.

A further long-term direction involves fully quantum architectures, where both the model and the optimization process are purely quantum, without classical layers or pre-/post-processing. Such models could reveal additional advantages of end-to-end quantum learning, though realizing them will require significant advances in quantum hardware and optimization techniques.

Limitations include high computational costs when training quantum models on simulators, which may restrict scalability. High-performance computing and, eventually, fault-tolerant quantum hardware will be essential to unlock the full potential of these architectures.

Overall, the convergence of quantum computing, reinforcement learning, and neuroscience-inspired design opens a fertile path toward next-generation routing algorithms. Continued progress in hardware, simulation, and quantum algorithm design will be vital to scaling these approaches to larger, real-world CVRP instances and achieving expressive, adaptive, and resource-efficient quantum routing agents.

%Bibliography
\bibliographystyle{unsrt}  
\bibliography{references}

\end{document}

%% file: images/DistanceBoxPlot.pgf
\begingroup%
\makeatletter%
\begin{pgfpicture}%
\pgfpathrectangle{\pgfpointorigin}{\pgfqpoint{8.000000in}{4.000000in}}%
\pgfusepath{use as bounding box, clip}%
\begin{pgfscope}%
\pgfsetbuttcap%
\pgfsetmiterjoin%
\definecolor{currentfill}{rgb}{1.000000,1.000000,1.000000}%
\pgfsetfillcolor{currentfill}%
\pgfsetlinewidth{0.000000pt}%
\definecolor{currentstroke}{rgb}{1.000000,1.000000,1.000000}%
\pgfsetstrokecolor{currentstroke}%
\pgfsetdash{}{0pt}%
\pgfpathmoveto{\pgfqpoint{0.000000in}{0.000000in}}%
\pgfpathlineto{\pgfqpoint{8.000000in}{0.000000in}}%
\pgfpathlineto{\pgfqpoint{8.000000in}{4.000000in}}%
\pgfpathlineto{\pgfqpoint{0.000000in}{4.000000in}}%
\pgfpathlineto{\pgfqpoint{0.000000in}{0.000000in}}%
\pgfpathclose%
\pgfusepath{fill}%
\end{pgfscope}%
\begin{pgfscope}%
\pgfsetbuttcap%
\pgfsetmiterjoin%
\definecolor{currentfill}{rgb}{1.000000,1.000000,1.000000}%
\pgfsetfillcolor{currentfill}%
\pgfsetlinewidth{0.000000pt}%
\definecolor{currentstroke}{rgb}{0.000000,0.000000,0.000000}%
\pgfsetstrokecolor{currentstroke}%
\pgfsetstrokeopacity{0.000000}%
\pgfsetdash{}{0pt}%
\pgfpathmoveto{\pgfqpoint{1.000000in}{0.440000in}}%
\pgfpathlineto{\pgfqpoint{7.200000in}{0.440000in}}%
\pgfpathlineto{\pgfqpoint{7.200000in}{3.520000in}}%
\pgfpathlineto{\pgfqpoint{1.000000in}{3.520000in}}%
\pgfpathlineto{\pgfqpoint{1.000000in}{0.440000in}}%
\pgfpathclose%
\pgfusepath{fill}%
\end{pgfscope}%
\begin{pgfscope}%
\pgfsetbuttcap%
\pgfsetroundjoin%
\definecolor{currentfill}{rgb}{0.000000,0.000000,0.000000}%
\pgfsetfillcolor{currentfill}%
\pgfsetlinewidth{0.803000pt}%
\definecolor{currentstroke}{rgb}{0.000000,0.000000,0.000000}%
\pgfsetstrokecolor{currentstroke}%
\pgfsetdash{}{0pt}%
\pgfsys@defobject{currentmarker}{\pgfqpoint{0.000000in}{-0.048611in}}{\pgfqpoint{0.000000in}{0.000000in}}{%
\pgfpathmoveto{\pgfqpoint{0.000000in}{0.000000in}}%
\pgfpathlineto{\pgfqpoint{0.000000in}{-0.048611in}}%
\pgfusepath{stroke,fill}%
}%
\begin{pgfscope}%
\pgfsys@transformshift{2.033333in}{0.440000in}%
\pgfsys@useobject{currentmarker}{}%
\end{pgfscope}%
\end{pgfscope}%
\begin{pgfscope}%
\definecolor{textcolor}{rgb}{0.000000,0.000000,0.000000}%
\pgfsetstrokecolor{textcolor}%
\pgfsetfillcolor{textcolor}%
\pgftext[x=2.033333in,y=0.342778in,,top]{\color{textcolor}{\rmfamily\fontsize{10.000000}{12.000000}\selectfont\catcode`\^=\active\def^{\ifmmode\sp\else\^{}\fi}\catcode`\%=\active\def%{\%}Quantum}}%
\end{pgfscope}%
\begin{pgfscope}%
\pgfsetbuttcap%
\pgfsetroundjoin%
\definecolor{currentfill}{rgb}{0.000000,0.000000,0.000000}%
\pgfsetfillcolor{currentfill}%
\pgfsetlinewidth{0.803000pt}%
\definecolor{currentstroke}{rgb}{0.000000,0.000000,0.000000}%
\pgfsetstrokecolor{currentstroke}%
\pgfsetdash{}{0pt}%
\pgfsys@defobject{currentmarker}{\pgfqpoint{0.000000in}{-0.048611in}}{\pgfqpoint{0.000000in}{0.000000in}}{%
\pgfpathmoveto{\pgfqpoint{0.000000in}{0.000000in}}%
\pgfpathlineto{\pgfqpoint{0.000000in}{-0.048611in}}%
\pgfusepath{stroke,fill}%
}%
\begin{pgfscope}%
\pgfsys@transformshift{4.100000in}{0.440000in}%
\pgfsys@useobject{currentmarker}{}%
\end{pgfscope}%
\end{pgfscope}%
\begin{pgfscope}%
\definecolor{textcolor}{rgb}{0.000000,0.000000,0.000000}%
\pgfsetstrokecolor{textcolor}%
\pgfsetfillcolor{textcolor}%
\pgftext[x=4.100000in,y=0.342778in,,top]{\color{textcolor}{\rmfamily\fontsize{10.000000}{12.000000}\selectfont\catcode`\^=\active\def^{\ifmmode\sp\else\^{}\fi}\catcode`\%=\active\def%{\%}Hybrid}}%
\end{pgfscope}%
\begin{pgfscope}%
\pgfsetbuttcap%
\pgfsetroundjoin%
\definecolor{currentfill}{rgb}{0.000000,0.000000,0.000000}%
\pgfsetfillcolor{currentfill}%
\pgfsetlinewidth{0.803000pt}%
\definecolor{currentstroke}{rgb}{0.000000,0.000000,0.000000}%
\pgfsetstrokecolor{currentstroke}%
\pgfsetdash{}{0pt}%
\pgfsys@defobject{currentmarker}{\pgfqpoint{0.000000in}{-0.048611in}}{\pgfqpoint{0.000000in}{0.000000in}}{%
\pgfpathmoveto{\pgfqpoint{0.000000in}{0.000000in}}%
\pgfpathlineto{\pgfqpoint{0.000000in}{-0.048611in}}%
\pgfusepath{stroke,fill}%
}%
\begin{pgfscope}%
\pgfsys@transformshift{6.166667in}{0.440000in}%
\pgfsys@useobject{currentmarker}{}%
\end{pgfscope}%
\end{pgfscope}%
\begin{pgfscope}%
\definecolor{textcolor}{rgb}{0.000000,0.000000,0.000000}%
\pgfsetstrokecolor{textcolor}%
\pgfsetfillcolor{textcolor}%
\pgftext[x=6.166667in,y=0.342778in,,top]{\color{textcolor}{\rmfamily\fontsize{10.000000}{12.000000}\selectfont\catcode`\^=\active\def^{\ifmmode\sp\else\^{}\fi}\catcode`\%=\active\def%{\%}Classic}}%
\end{pgfscope}%
\begin{pgfscope}%
\pgfpathrectangle{\pgfqpoint{1.000000in}{0.440000in}}{\pgfqpoint{6.200000in}{3.080000in}}%
\pgfusepath{clip}%
\pgfsetrectcap%
\pgfsetroundjoin%
\pgfsetlinewidth{0.803000pt}%
\definecolor{currentstroke}{rgb}{0.690196,0.690196,0.690196}%
\pgfsetstrokecolor{currentstroke}%
\pgfsetstrokeopacity{0.300000}%
\pgfsetdash{}{0pt}%
\pgfpathmoveto{\pgfqpoint{1.000000in}{0.993316in}}%
\pgfpathlineto{\pgfqpoint{7.200000in}{0.993316in}}%
\pgfusepath{stroke}%
\end{pgfscope}%
\begin{pgfscope}%
\pgfsetbuttcap%
\pgfsetroundjoin%
\definecolor{currentfill}{rgb}{0.000000,0.000000,0.000000}%
\pgfsetfillcolor{currentfill}%
\pgfsetlinewidth{0.803000pt}%
\definecolor{currentstroke}{rgb}{0.000000,0.000000,0.000000}%
\pgfsetstrokecolor{currentstroke}%
\pgfsetdash{}{0pt}%
\pgfsys@defobject{currentmarker}{\pgfqpoint{-0.048611in}{0.000000in}}{\pgfqpoint{-0.000000in}{0.000000in}}{%
\pgfpathmoveto{\pgfqpoint{-0.000000in}{0.000000in}}%
\pgfpathlineto{\pgfqpoint{-0.048611in}{0.000000in}}%
\pgfusepath{stroke,fill}%
}%
\begin{pgfscope}%
\pgfsys@transformshift{1.000000in}{0.993316in}%
\pgfsys@useobject{currentmarker}{}%
\end{pgfscope}%
\end{pgfscope}%
\begin{pgfscope}%
\definecolor{textcolor}{rgb}{0.000000,0.000000,0.000000}%
\pgfsetstrokecolor{textcolor}%
\pgfsetfillcolor{textcolor}%
\pgftext[x=0.725308in, y=0.945091in, left, base]{\color{textcolor}{\rmfamily\fontsize{10.000000}{12.000000}\selectfont\catcode`\^=\active\def^{\ifmmode\sp\else\^{}\fi}\catcode`\%=\active\def%{\%}$\mathdefault{6.0}$}}%
\end{pgfscope}%
\begin{pgfscope}%
\pgfpathrectangle{\pgfqpoint{1.000000in}{0.440000in}}{\pgfqpoint{6.200000in}{3.080000in}}%
\pgfusepath{clip}%
\pgfsetrectcap%
\pgfsetroundjoin%
\pgfsetlinewidth{0.803000pt}%
\definecolor{currentstroke}{rgb}{0.690196,0.690196,0.690196}%
\pgfsetstrokecolor{currentstroke}%
\pgfsetstrokeopacity{0.300000}%
\pgfsetdash{}{0pt}%
\pgfpathmoveto{\pgfqpoint{1.000000in}{1.661971in}}%
\pgfpathlineto{\pgfqpoint{7.200000in}{1.661971in}}%
\pgfusepath{stroke}%
\end{pgfscope}%
\begin{pgfscope}%
\pgfsetbuttcap%
\pgfsetroundjoin%
\definecolor{currentfill}{rgb}{0.000000,0.000000,0.000000}%
\pgfsetfillcolor{currentfill}%
\pgfsetlinewidth{0.803000pt}%
\definecolor{currentstroke}{rgb}{0.000000,0.000000,0.000000}%
\pgfsetstrokecolor{currentstroke}%
\pgfsetdash{}{0pt}%
\pgfsys@defobject{currentmarker}{\pgfqpoint{-0.048611in}{0.000000in}}{\pgfqpoint{-0.000000in}{0.000000in}}{%
\pgfpathmoveto{\pgfqpoint{-0.000000in}{0.000000in}}%
\pgfpathlineto{\pgfqpoint{-0.048611in}{0.000000in}}%
\pgfusepath{stroke,fill}%
}%
\begin{pgfscope}%
\pgfsys@transformshift{1.000000in}{1.661971in}%
\pgfsys@useobject{currentmarker}{}%
\end{pgfscope}%
\end{pgfscope}%
\begin{pgfscope}%
\definecolor{textcolor}{rgb}{0.000000,0.000000,0.000000}%
\pgfsetstrokecolor{textcolor}%
\pgfsetfillcolor{textcolor}%
\pgftext[x=0.725308in, y=1.613745in, left, base]{\color{textcolor}{\rmfamily\fontsize{10.000000}{12.000000}\selectfont\catcode`\^=\active\def^{\ifmmode\sp\else\^{}\fi}\catcode`\%=\active\def%{\%}$\mathdefault{6.5}$}}%
\end{pgfscope}%
\begin{pgfscope}%
\pgfpathrectangle{\pgfqpoint{1.000000in}{0.440000in}}{\pgfqpoint{6.200000in}{3.080000in}}%
\pgfusepath{clip}%
\pgfsetrectcap%
\pgfsetroundjoin%
\pgfsetlinewidth{0.803000pt}%
\definecolor{currentstroke}{rgb}{0.690196,0.690196,0.690196}%
\pgfsetstrokecolor{currentstroke}%
\pgfsetstrokeopacity{0.300000}%
\pgfsetdash{}{0pt}%
\pgfpathmoveto{\pgfqpoint{1.000000in}{2.330625in}}%
\pgfpathlineto{\pgfqpoint{7.200000in}{2.330625in}}%
\pgfusepath{stroke}%
\end{pgfscope}%
\begin{pgfscope}%
\pgfsetbuttcap%
\pgfsetroundjoin%
\definecolor{currentfill}{rgb}{0.000000,0.000000,0.000000}%
\pgfsetfillcolor{currentfill}%
\pgfsetlinewidth{0.803000pt}%
\definecolor{currentstroke}{rgb}{0.000000,0.000000,0.000000}%
\pgfsetstrokecolor{currentstroke}%
\pgfsetdash{}{0pt}%
\pgfsys@defobject{currentmarker}{\pgfqpoint{-0.048611in}{0.000000in}}{\pgfqpoint{-0.000000in}{0.000000in}}{%
\pgfpathmoveto{\pgfqpoint{-0.000000in}{0.000000in}}%
\pgfpathlineto{\pgfqpoint{-0.048611in}{0.000000in}}%
\pgfusepath{stroke,fill}%
}%
\begin{pgfscope}%
\pgfsys@transformshift{1.000000in}{2.330625in}%
\pgfsys@useobject{currentmarker}{}%
\end{pgfscope}%
\end{pgfscope}%
\begin{pgfscope}%
\definecolor{textcolor}{rgb}{0.000000,0.000000,0.000000}%
\pgfsetstrokecolor{textcolor}%
\pgfsetfillcolor{textcolor}%
\pgftext[x=0.725308in, y=2.282400in, left, base]{\color{textcolor}{\rmfamily\fontsize{10.000000}{12.000000}\selectfont\catcode`\^=\active\def^{\ifmmode\sp\else\^{}\fi}\catcode`\%=\active\def%{\%}$\mathdefault{7.0}$}}%
\end{pgfscope}%
\begin{pgfscope}%
\pgfpathrectangle{\pgfqpoint{1.000000in}{0.440000in}}{\pgfqpoint{6.200000in}{3.080000in}}%
\pgfusepath{clip}%
\pgfsetrectcap%
\pgfsetroundjoin%
\pgfsetlinewidth{0.803000pt}%
\definecolor{currentstroke}{rgb}{0.690196,0.690196,0.690196}%
\pgfsetstrokecolor{currentstroke}%
\pgfsetstrokeopacity{0.300000}%
\pgfsetdash{}{0pt}%
\pgfpathmoveto{\pgfqpoint{1.000000in}{2.999279in}}%
\pgfpathlineto{\pgfqpoint{7.200000in}{2.999279in}}%
\pgfusepath{stroke}%
\end{pgfscope}%
\begin{pgfscope}%
\pgfsetbuttcap%
\pgfsetroundjoin%
\definecolor{currentfill}{rgb}{0.000000,0.000000,0.000000}%
\pgfsetfillcolor{currentfill}%
\pgfsetlinewidth{0.803000pt}%
\definecolor{currentstroke}{rgb}{0.000000,0.000000,0.000000}%
\pgfsetstrokecolor{currentstroke}%
\pgfsetdash{}{0pt}%
\pgfsys@defobject{currentmarker}{\pgfqpoint{-0.048611in}{0.000000in}}{\pgfqpoint{-0.000000in}{0.000000in}}{%
\pgfpathmoveto{\pgfqpoint{-0.000000in}{0.000000in}}%
\pgfpathlineto{\pgfqpoint{-0.048611in}{0.000000in}}%
\pgfusepath{stroke,fill}%
}%
\begin{pgfscope}%
\pgfsys@transformshift{1.000000in}{2.999279in}%
\pgfsys@useobject{currentmarker}{}%
\end{pgfscope}%
\end{pgfscope}%
\begin{pgfscope}%
\definecolor{textcolor}{rgb}{0.000000,0.000000,0.000000}%
\pgfsetstrokecolor{textcolor}%
\pgfsetfillcolor{textcolor}%
\pgftext[x=0.725308in, y=2.951054in, left, base]{\color{textcolor}{\rmfamily\fontsize{10.000000}{12.000000}\selectfont\catcode`\^=\active\def^{\ifmmode\sp\else\^{}\fi}\catcode`\%=\active\def%{\%}$\mathdefault{7.5}$}}%
\end{pgfscope}%
\begin{pgfscope}%
\definecolor{textcolor}{rgb}{0.000000,0.000000,0.000000}%
\pgfsetstrokecolor{textcolor}%
\pgfsetfillcolor{textcolor}%
\pgftext[x=0.669752in,y=1.980000in,,bottom,rotate=90.000000]{\color{textcolor}{\rmfamily\fontsize{10.000000}{12.000000}\selectfont\catcode`\^=\active\def^{\ifmmode\sp\else\^{}\fi}\catcode`\%=\active\def%{\%}Total distance traveled}}%
\end{pgfscope}%
\begin{pgfscope}%
\pgfpathrectangle{\pgfqpoint{1.000000in}{0.440000in}}{\pgfqpoint{6.200000in}{3.080000in}}%
\pgfusepath{clip}%
\pgfsetrectcap%
\pgfsetroundjoin%
\pgfsetlinewidth{1.003750pt}%
\definecolor{currentstroke}{rgb}{0.000000,0.000000,0.000000}%
\pgfsetstrokecolor{currentstroke}%
\pgfsetdash{}{0pt}%
\pgfpathmoveto{\pgfqpoint{1.723333in}{1.663616in}}%
\pgfpathlineto{\pgfqpoint{2.343333in}{1.663616in}}%
\pgfpathlineto{\pgfqpoint{2.343333in}{2.556258in}}%
\pgfpathlineto{\pgfqpoint{1.723333in}{2.556258in}}%
\pgfpathlineto{\pgfqpoint{1.723333in}{1.663616in}}%
\pgfusepath{stroke}%
\end{pgfscope}%
\begin{pgfscope}%
\pgfpathrectangle{\pgfqpoint{1.000000in}{0.440000in}}{\pgfqpoint{6.200000in}{3.080000in}}%
\pgfusepath{clip}%
\pgfsetrectcap%
\pgfsetroundjoin%
\pgfsetlinewidth{1.003750pt}%
\definecolor{currentstroke}{rgb}{0.000000,0.000000,0.000000}%
\pgfsetstrokecolor{currentstroke}%
\pgfsetdash{}{0pt}%
\pgfpathmoveto{\pgfqpoint{2.033333in}{1.663616in}}%
\pgfpathlineto{\pgfqpoint{2.033333in}{0.866732in}}%
\pgfusepath{stroke}%
\end{pgfscope}%
\begin{pgfscope}%
\pgfpathrectangle{\pgfqpoint{1.000000in}{0.440000in}}{\pgfqpoint{6.200000in}{3.080000in}}%
\pgfusepath{clip}%
\pgfsetrectcap%
\pgfsetroundjoin%
\pgfsetlinewidth{1.003750pt}%
\definecolor{currentstroke}{rgb}{0.000000,0.000000,0.000000}%
\pgfsetstrokecolor{currentstroke}%
\pgfsetdash{}{0pt}%
\pgfpathmoveto{\pgfqpoint{2.033333in}{2.556258in}}%
\pgfpathlineto{\pgfqpoint{2.033333in}{3.380000in}}%
\pgfusepath{stroke}%
\end{pgfscope}%
\begin{pgfscope}%
\pgfpathrectangle{\pgfqpoint{1.000000in}{0.440000in}}{\pgfqpoint{6.200000in}{3.080000in}}%
\pgfusepath{clip}%
\pgfsetrectcap%
\pgfsetroundjoin%
\pgfsetlinewidth{1.003750pt}%
\definecolor{currentstroke}{rgb}{0.000000,0.000000,0.000000}%
\pgfsetstrokecolor{currentstroke}%
\pgfsetdash{}{0pt}%
\pgfpathmoveto{\pgfqpoint{1.878333in}{0.866732in}}%
\pgfpathlineto{\pgfqpoint{2.188333in}{0.866732in}}%
\pgfusepath{stroke}%
\end{pgfscope}%
\begin{pgfscope}%
\pgfpathrectangle{\pgfqpoint{1.000000in}{0.440000in}}{\pgfqpoint{6.200000in}{3.080000in}}%
\pgfusepath{clip}%
\pgfsetrectcap%
\pgfsetroundjoin%
\pgfsetlinewidth{1.003750pt}%
\definecolor{currentstroke}{rgb}{0.000000,0.000000,0.000000}%
\pgfsetstrokecolor{currentstroke}%
\pgfsetdash{}{0pt}%
\pgfpathmoveto{\pgfqpoint{1.878333in}{3.380000in}}%
\pgfpathlineto{\pgfqpoint{2.188333in}{3.380000in}}%
\pgfusepath{stroke}%
\end{pgfscope}%
\begin{pgfscope}%
\pgfpathrectangle{\pgfqpoint{1.000000in}{0.440000in}}{\pgfqpoint{6.200000in}{3.080000in}}%
\pgfusepath{clip}%
\pgfsetrectcap%
\pgfsetroundjoin%
\pgfsetlinewidth{1.003750pt}%
\definecolor{currentstroke}{rgb}{0.000000,0.000000,0.000000}%
\pgfsetstrokecolor{currentstroke}%
\pgfsetdash{}{0pt}%
\pgfpathmoveto{\pgfqpoint{3.790000in}{1.486170in}}%
\pgfpathlineto{\pgfqpoint{4.410000in}{1.486170in}}%
\pgfpathlineto{\pgfqpoint{4.410000in}{2.612608in}}%
\pgfpathlineto{\pgfqpoint{3.790000in}{2.612608in}}%
\pgfpathlineto{\pgfqpoint{3.790000in}{1.486170in}}%
\pgfusepath{stroke}%
\end{pgfscope}%
\begin{pgfscope}%
\pgfpathrectangle{\pgfqpoint{1.000000in}{0.440000in}}{\pgfqpoint{6.200000in}{3.080000in}}%
\pgfusepath{clip}%
\pgfsetrectcap%
\pgfsetroundjoin%
\pgfsetlinewidth{1.003750pt}%
\definecolor{currentstroke}{rgb}{0.000000,0.000000,0.000000}%
\pgfsetstrokecolor{currentstroke}%
\pgfsetdash{}{0pt}%
\pgfpathmoveto{\pgfqpoint{4.100000in}{1.486170in}}%
\pgfpathlineto{\pgfqpoint{4.100000in}{0.580000in}}%
\pgfusepath{stroke}%
\end{pgfscope}%
\begin{pgfscope}%
\pgfpathrectangle{\pgfqpoint{1.000000in}{0.440000in}}{\pgfqpoint{6.200000in}{3.080000in}}%
\pgfusepath{clip}%
\pgfsetrectcap%
\pgfsetroundjoin%
\pgfsetlinewidth{1.003750pt}%
\definecolor{currentstroke}{rgb}{0.000000,0.000000,0.000000}%
\pgfsetstrokecolor{currentstroke}%
\pgfsetdash{}{0pt}%
\pgfpathmoveto{\pgfqpoint{4.100000in}{2.612608in}}%
\pgfpathlineto{\pgfqpoint{4.100000in}{2.987501in}}%
\pgfusepath{stroke}%
\end{pgfscope}%
\begin{pgfscope}%
\pgfpathrectangle{\pgfqpoint{1.000000in}{0.440000in}}{\pgfqpoint{6.200000in}{3.080000in}}%
\pgfusepath{clip}%
\pgfsetrectcap%
\pgfsetroundjoin%
\pgfsetlinewidth{1.003750pt}%
\definecolor{currentstroke}{rgb}{0.000000,0.000000,0.000000}%
\pgfsetstrokecolor{currentstroke}%
\pgfsetdash{}{0pt}%
\pgfpathmoveto{\pgfqpoint{3.945000in}{0.580000in}}%
\pgfpathlineto{\pgfqpoint{4.255000in}{0.580000in}}%
\pgfusepath{stroke}%
\end{pgfscope}%
\begin{pgfscope}%
\pgfpathrectangle{\pgfqpoint{1.000000in}{0.440000in}}{\pgfqpoint{6.200000in}{3.080000in}}%
\pgfusepath{clip}%
\pgfsetrectcap%
\pgfsetroundjoin%
\pgfsetlinewidth{1.003750pt}%
\definecolor{currentstroke}{rgb}{0.000000,0.000000,0.000000}%
\pgfsetstrokecolor{currentstroke}%
\pgfsetdash{}{0pt}%
\pgfpathmoveto{\pgfqpoint{3.945000in}{2.987501in}}%
\pgfpathlineto{\pgfqpoint{4.255000in}{2.987501in}}%
\pgfusepath{stroke}%
\end{pgfscope}%
\begin{pgfscope}%
\pgfpathrectangle{\pgfqpoint{1.000000in}{0.440000in}}{\pgfqpoint{6.200000in}{3.080000in}}%
\pgfusepath{clip}%
\pgfsetrectcap%
\pgfsetroundjoin%
\pgfsetlinewidth{1.003750pt}%
\definecolor{currentstroke}{rgb}{0.000000,0.000000,0.000000}%
\pgfsetstrokecolor{currentstroke}%
\pgfsetdash{}{0pt}%
\pgfpathmoveto{\pgfqpoint{5.856667in}{1.206565in}}%
\pgfpathlineto{\pgfqpoint{6.476667in}{1.206565in}}%
\pgfpathlineto{\pgfqpoint{6.476667in}{3.027362in}}%
\pgfpathlineto{\pgfqpoint{5.856667in}{3.027362in}}%
\pgfpathlineto{\pgfqpoint{5.856667in}{1.206565in}}%
\pgfusepath{stroke}%
\end{pgfscope}%
\begin{pgfscope}%
\pgfpathrectangle{\pgfqpoint{1.000000in}{0.440000in}}{\pgfqpoint{6.200000in}{3.080000in}}%
\pgfusepath{clip}%
\pgfsetrectcap%
\pgfsetroundjoin%
\pgfsetlinewidth{1.003750pt}%
\definecolor{currentstroke}{rgb}{0.000000,0.000000,0.000000}%
\pgfsetstrokecolor{currentstroke}%
\pgfsetdash{}{0pt}%
\pgfpathmoveto{\pgfqpoint{6.166667in}{1.206565in}}%
\pgfpathlineto{\pgfqpoint{6.166667in}{0.918433in}}%
\pgfusepath{stroke}%
\end{pgfscope}%
\begin{pgfscope}%
\pgfpathrectangle{\pgfqpoint{1.000000in}{0.440000in}}{\pgfqpoint{6.200000in}{3.080000in}}%
\pgfusepath{clip}%
\pgfsetrectcap%
\pgfsetroundjoin%
\pgfsetlinewidth{1.003750pt}%
\definecolor{currentstroke}{rgb}{0.000000,0.000000,0.000000}%
\pgfsetstrokecolor{currentstroke}%
\pgfsetdash{}{0pt}%
\pgfpathmoveto{\pgfqpoint{6.166667in}{3.027362in}}%
\pgfpathlineto{\pgfqpoint{6.166667in}{3.335463in}}%
\pgfusepath{stroke}%
\end{pgfscope}%
\begin{pgfscope}%
\pgfpathrectangle{\pgfqpoint{1.000000in}{0.440000in}}{\pgfqpoint{6.200000in}{3.080000in}}%
\pgfusepath{clip}%
\pgfsetrectcap%
\pgfsetroundjoin%
\pgfsetlinewidth{1.003750pt}%
\definecolor{currentstroke}{rgb}{0.000000,0.000000,0.000000}%
\pgfsetstrokecolor{currentstroke}%
\pgfsetdash{}{0pt}%
\pgfpathmoveto{\pgfqpoint{6.011667in}{0.918433in}}%
\pgfpathlineto{\pgfqpoint{6.321667in}{0.918433in}}%
\pgfusepath{stroke}%
\end{pgfscope}%
\begin{pgfscope}%
\pgfpathrectangle{\pgfqpoint{1.000000in}{0.440000in}}{\pgfqpoint{6.200000in}{3.080000in}}%
\pgfusepath{clip}%
\pgfsetrectcap%
\pgfsetroundjoin%
\pgfsetlinewidth{1.003750pt}%
\definecolor{currentstroke}{rgb}{0.000000,0.000000,0.000000}%
\pgfsetstrokecolor{currentstroke}%
\pgfsetdash{}{0pt}%
\pgfpathmoveto{\pgfqpoint{6.011667in}{3.335463in}}%
\pgfpathlineto{\pgfqpoint{6.321667in}{3.335463in}}%
\pgfusepath{stroke}%
\end{pgfscope}%
\begin{pgfscope}%
\pgfpathrectangle{\pgfqpoint{1.000000in}{0.440000in}}{\pgfqpoint{6.200000in}{3.080000in}}%
\pgfusepath{clip}%
\pgfsetbuttcap%
\pgfsetroundjoin%
\pgfsetlinewidth{1.003750pt}%
\definecolor{currentstroke}{rgb}{1.000000,0.498039,0.054902}%
\pgfsetstrokecolor{currentstroke}%
\pgfsetdash{}{0pt}%
\pgfpathmoveto{\pgfqpoint{1.723333in}{2.067982in}}%
\pgfpathlineto{\pgfqpoint{2.343333in}{2.067982in}}%
\pgfusepath{stroke}%
\end{pgfscope}%
\begin{pgfscope}%
\pgfpathrectangle{\pgfqpoint{1.000000in}{0.440000in}}{\pgfqpoint{6.200000in}{3.080000in}}%
\pgfusepath{clip}%
\pgfsetbuttcap%
\pgfsetroundjoin%
\pgfsetlinewidth{1.003750pt}%
\definecolor{currentstroke}{rgb}{1.000000,0.498039,0.054902}%
\pgfsetstrokecolor{currentstroke}%
\pgfsetdash{}{0pt}%
\pgfpathmoveto{\pgfqpoint{3.790000in}{1.959617in}}%
\pgfpathlineto{\pgfqpoint{4.410000in}{1.959617in}}%
\pgfusepath{stroke}%
\end{pgfscope}%
\begin{pgfscope}%
\pgfpathrectangle{\pgfqpoint{1.000000in}{0.440000in}}{\pgfqpoint{6.200000in}{3.080000in}}%
\pgfusepath{clip}%
\pgfsetbuttcap%
\pgfsetroundjoin%
\pgfsetlinewidth{1.003750pt}%
\definecolor{currentstroke}{rgb}{1.000000,0.498039,0.054902}%
\pgfsetstrokecolor{currentstroke}%
\pgfsetdash{}{0pt}%
\pgfpathmoveto{\pgfqpoint{5.856667in}{2.483663in}}%
\pgfpathlineto{\pgfqpoint{6.476667in}{2.483663in}}%
\pgfusepath{stroke}%
\end{pgfscope}%
\begin{pgfscope}%
\pgfsetrectcap%
\pgfsetmiterjoin%
\pgfsetlinewidth{0.803000pt}%
\definecolor{currentstroke}{rgb}{0.000000,0.000000,0.000000}%
\pgfsetstrokecolor{currentstroke}%
\pgfsetdash{}{0pt}%
\pgfpathmoveto{\pgfqpoint{1.000000in}{0.440000in}}%
\pgfpathlineto{\pgfqpoint{1.000000in}{3.520000in}}%
\pgfusepath{stroke}%
\end{pgfscope}%
\begin{pgfscope}%
\pgfsetrectcap%
\pgfsetmiterjoin%
\pgfsetlinewidth{0.803000pt}%
\definecolor{currentstroke}{rgb}{0.000000,0.000000,0.000000}%
\pgfsetstrokecolor{currentstroke}%
\pgfsetdash{}{0pt}%
\pgfpathmoveto{\pgfqpoint{7.200000in}{0.440000in}}%
\pgfpathlineto{\pgfqpoint{7.200000in}{3.520000in}}%
\pgfusepath{stroke}%
\end{pgfscope}%
\begin{pgfscope}%
\pgfsetrectcap%
\pgfsetmiterjoin%
\pgfsetlinewidth{0.803000pt}%
\definecolor{currentstroke}{rgb}{0.000000,0.000000,0.000000}%
\pgfsetstrokecolor{currentstroke}%
\pgfsetdash{}{0pt}%
\pgfpathmoveto{\pgfqpoint{1.000000in}{0.440000in}}%
\pgfpathlineto{\pgfqpoint{7.200000in}{0.440000in}}%
\pgfusepath{stroke}%
\end{pgfscope}%
\begin{pgfscope}%
\pgfsetrectcap%
\pgfsetmiterjoin%
\pgfsetlinewidth{0.803000pt}%
\definecolor{currentstroke}{rgb}{0.000000,0.000000,0.000000}%
\pgfsetstrokecolor{currentstroke}%
\pgfsetdash{}{0pt}%
\pgfpathmoveto{\pgfqpoint{1.000000in}{3.520000in}}%
\pgfpathlineto{\pgfqpoint{7.200000in}{3.520000in}}%
\pgfusepath{stroke}%
\end{pgfscope}%
\begin{pgfscope}%
\definecolor{textcolor}{rgb}{0.000000,0.000000,0.000000}%
\pgfsetstrokecolor{textcolor}%
\pgfsetfillcolor{textcolor}%
\pgftext[x=4.100000in,y=3.603333in,,base]{\color{textcolor}{\rmfamily\fontsize{12.000000}{14.400000}\selectfont\catcode`\^=\active\def^{\ifmmode\sp\else\^{}\fi}\catcode`\%=\active\def%{\%}CVRP – Total distance comparison}}%
\end{pgfscope}%
\end{pgfpicture}%
\makeatother%
\endgroup%

%% file: images/CompactationBoxPlot.pgf
\begingroup%
\makeatletter%
\begin{pgfpicture}%
\pgfpathrectangle{\pgfpointorigin}{\pgfqpoint{8.000000in}{4.000000in}}%
\pgfusepath{use as bounding box, clip}%
\begin{pgfscope}%
\pgfsetbuttcap%
\pgfsetmiterjoin%
\definecolor{currentfill}{rgb}{1.000000,1.000000,1.000000}%
\pgfsetfillcolor{currentfill}%
\pgfsetlinewidth{0.000000pt}%
\definecolor{currentstroke}{rgb}{1.000000,1.000000,1.000000}%
\pgfsetstrokecolor{currentstroke}%
\pgfsetdash{}{0pt}%
\pgfpathmoveto{\pgfqpoint{0.000000in}{0.000000in}}%
\pgfpathlineto{\pgfqpoint{8.000000in}{0.000000in}}%
\pgfpathlineto{\pgfqpoint{8.000000in}{4.000000in}}%
\pgfpathlineto{\pgfqpoint{0.000000in}{4.000000in}}%
\pgfpathlineto{\pgfqpoint{0.000000in}{0.000000in}}%
\pgfpathclose%
\pgfusepath{fill}%
\end{pgfscope}%
\begin{pgfscope}%
\pgfsetbuttcap%
\pgfsetmiterjoin%
\definecolor{currentfill}{rgb}{1.000000,1.000000,1.000000}%
\pgfsetfillcolor{currentfill}%
\pgfsetlinewidth{0.000000pt}%
\definecolor{currentstroke}{rgb}{0.000000,0.000000,0.000000}%
\pgfsetstrokecolor{currentstroke}%
\pgfsetstrokeopacity{0.000000}%
\pgfsetdash{}{0pt}%
\pgfpathmoveto{\pgfqpoint{1.000000in}{0.440000in}}%
\pgfpathlineto{\pgfqpoint{7.200000in}{0.440000in}}%
\pgfpathlineto{\pgfqpoint{7.200000in}{3.520000in}}%
\pgfpathlineto{\pgfqpoint{1.000000in}{3.520000in}}%
\pgfpathlineto{\pgfqpoint{1.000000in}{0.440000in}}%
\pgfpathclose%
\pgfusepath{fill}%
\end{pgfscope}%
\begin{pgfscope}%
\pgfsetbuttcap%
\pgfsetroundjoin%
\definecolor{currentfill}{rgb}{0.000000,0.000000,0.000000}%
\pgfsetfillcolor{currentfill}%
\pgfsetlinewidth{0.803000pt}%
\definecolor{currentstroke}{rgb}{0.000000,0.000000,0.000000}%
\pgfsetstrokecolor{currentstroke}%
\pgfsetdash{}{0pt}%
\pgfsys@defobject{currentmarker}{\pgfqpoint{0.000000in}{-0.048611in}}{\pgfqpoint{0.000000in}{0.000000in}}{%
\pgfpathmoveto{\pgfqpoint{0.000000in}{0.000000in}}%
\pgfpathlineto{\pgfqpoint{0.000000in}{-0.048611in}}%
\pgfusepath{stroke,fill}%
}%
\begin{pgfscope}%
\pgfsys@transformshift{2.033333in}{0.440000in}%
\pgfsys@useobject{currentmarker}{}%
\end{pgfscope}%
\end{pgfscope}%
\begin{pgfscope}%
\definecolor{textcolor}{rgb}{0.000000,0.000000,0.000000}%
\pgfsetstrokecolor{textcolor}%
\pgfsetfillcolor{textcolor}%
\pgftext[x=2.033333in,y=0.342778in,,top]{\color{textcolor}{\rmfamily\fontsize{10.000000}{12.000000}\selectfont\catcode`\^=\active\def^{\ifmmode\sp\else\^{}\fi}\catcode`\%=\active\def%{\%}Quantum}}%
\end{pgfscope}%
\begin{pgfscope}%
\pgfsetbuttcap%
\pgfsetroundjoin%
\definecolor{currentfill}{rgb}{0.000000,0.000000,0.000000}%
\pgfsetfillcolor{currentfill}%
\pgfsetlinewidth{0.803000pt}%
\definecolor{currentstroke}{rgb}{0.000000,0.000000,0.000000}%
\pgfsetstrokecolor{currentstroke}%
\pgfsetdash{}{0pt}%
\pgfsys@defobject{currentmarker}{\pgfqpoint{0.000000in}{-0.048611in}}{\pgfqpoint{0.000000in}{0.000000in}}{%
\pgfpathmoveto{\pgfqpoint{0.000000in}{0.000000in}}%
\pgfpathlineto{\pgfqpoint{0.000000in}{-0.048611in}}%
\pgfusepath{stroke,fill}%
}%
\begin{pgfscope}%
\pgfsys@transformshift{4.100000in}{0.440000in}%
\pgfsys@useobject{currentmarker}{}%
\end{pgfscope}%
\end{pgfscope}%
\begin{pgfscope}%
\definecolor{textcolor}{rgb}{0.000000,0.000000,0.000000}%
\pgfsetstrokecolor{textcolor}%
\pgfsetfillcolor{textcolor}%
\pgftext[x=4.100000in,y=0.342778in,,top]{\color{textcolor}{\rmfamily\fontsize{10.000000}{12.000000}\selectfont\catcode`\^=\active\def^{\ifmmode\sp\else\^{}\fi}\catcode`\%=\active\def%{\%}Hybrid}}%
\end{pgfscope}%
\begin{pgfscope}%
\pgfsetbuttcap%
\pgfsetroundjoin%
\definecolor{currentfill}{rgb}{0.000000,0.000000,0.000000}%
\pgfsetfillcolor{currentfill}%
\pgfsetlinewidth{0.803000pt}%
\definecolor{currentstroke}{rgb}{0.000000,0.000000,0.000000}%
\pgfsetstrokecolor{currentstroke}%
\pgfsetdash{}{0pt}%
\pgfsys@defobject{currentmarker}{\pgfqpoint{0.000000in}{-0.048611in}}{\pgfqpoint{0.000000in}{0.000000in}}{%
\pgfpathmoveto{\pgfqpoint{0.000000in}{0.000000in}}%
\pgfpathlineto{\pgfqpoint{0.000000in}{-0.048611in}}%
\pgfusepath{stroke,fill}%
}%
\begin{pgfscope}%
\pgfsys@transformshift{6.166667in}{0.440000in}%
\pgfsys@useobject{currentmarker}{}%
\end{pgfscope}%
\end{pgfscope}%
\begin{pgfscope}%
\definecolor{textcolor}{rgb}{0.000000,0.000000,0.000000}%
\pgfsetstrokecolor{textcolor}%
\pgfsetfillcolor{textcolor}%
\pgftext[x=6.166667in,y=0.342778in,,top]{\color{textcolor}{\rmfamily\fontsize{10.000000}{12.000000}\selectfont\catcode`\^=\active\def^{\ifmmode\sp\else\^{}\fi}\catcode`\%=\active\def%{\%}Classic}}%
\end{pgfscope}%
\begin{pgfscope}%
\pgfpathrectangle{\pgfqpoint{1.000000in}{0.440000in}}{\pgfqpoint{6.200000in}{3.080000in}}%
\pgfusepath{clip}%
\pgfsetrectcap%
\pgfsetroundjoin%
\pgfsetlinewidth{0.803000pt}%
\definecolor{currentstroke}{rgb}{0.690196,0.690196,0.690196}%
\pgfsetstrokecolor{currentstroke}%
\pgfsetstrokeopacity{0.300000}%
\pgfsetdash{}{0pt}%
\pgfpathmoveto{\pgfqpoint{1.000000in}{0.725699in}}%
\pgfpathlineto{\pgfqpoint{7.200000in}{0.725699in}}%
\pgfusepath{stroke}%
\end{pgfscope}%
\begin{pgfscope}%
\pgfsetbuttcap%
\pgfsetroundjoin%
\definecolor{currentfill}{rgb}{0.000000,0.000000,0.000000}%
\pgfsetfillcolor{currentfill}%
\pgfsetlinewidth{0.803000pt}%
\definecolor{currentstroke}{rgb}{0.000000,0.000000,0.000000}%
\pgfsetstrokecolor{currentstroke}%
\pgfsetdash{}{0pt}%
\pgfsys@defobject{currentmarker}{\pgfqpoint{-0.048611in}{0.000000in}}{\pgfqpoint{-0.000000in}{0.000000in}}{%
\pgfpathmoveto{\pgfqpoint{-0.000000in}{0.000000in}}%
\pgfpathlineto{\pgfqpoint{-0.048611in}{0.000000in}}%
\pgfusepath{stroke,fill}%
}%
\begin{pgfscope}%
\pgfsys@transformshift{1.000000in}{0.725699in}%
\pgfsys@useobject{currentmarker}{}%
\end{pgfscope}%
\end{pgfscope}%
\begin{pgfscope}%
\definecolor{textcolor}{rgb}{0.000000,0.000000,0.000000}%
\pgfsetstrokecolor{textcolor}%
\pgfsetfillcolor{textcolor}%
\pgftext[x=0.763888in, y=0.677473in, left, base]{\color{textcolor}{\rmfamily\fontsize{10.000000}{12.000000}\selectfont\catcode`\^=\active\def^{\ifmmode\sp\else\^{}\fi}\catcode`\%=\active\def%{\%}$\mathdefault{26}$}}%
\end{pgfscope}%
\begin{pgfscope}%
\pgfpathrectangle{\pgfqpoint{1.000000in}{0.440000in}}{\pgfqpoint{6.200000in}{3.080000in}}%
\pgfusepath{clip}%
\pgfsetrectcap%
\pgfsetroundjoin%
\pgfsetlinewidth{0.803000pt}%
\definecolor{currentstroke}{rgb}{0.690196,0.690196,0.690196}%
\pgfsetstrokecolor{currentstroke}%
\pgfsetstrokeopacity{0.300000}%
\pgfsetdash{}{0pt}%
\pgfpathmoveto{\pgfqpoint{1.000000in}{1.174203in}}%
\pgfpathlineto{\pgfqpoint{7.200000in}{1.174203in}}%
\pgfusepath{stroke}%
\end{pgfscope}%
\begin{pgfscope}%
\pgfsetbuttcap%
\pgfsetroundjoin%
\definecolor{currentfill}{rgb}{0.000000,0.000000,0.000000}%
\pgfsetfillcolor{currentfill}%
\pgfsetlinewidth{0.803000pt}%
\definecolor{currentstroke}{rgb}{0.000000,0.000000,0.000000}%
\pgfsetstrokecolor{currentstroke}%
\pgfsetdash{}{0pt}%
\pgfsys@defobject{currentmarker}{\pgfqpoint{-0.048611in}{0.000000in}}{\pgfqpoint{-0.000000in}{0.000000in}}{%
\pgfpathmoveto{\pgfqpoint{-0.000000in}{0.000000in}}%
\pgfpathlineto{\pgfqpoint{-0.048611in}{0.000000in}}%
\pgfusepath{stroke,fill}%
}%
\begin{pgfscope}%
\pgfsys@transformshift{1.000000in}{1.174203in}%
\pgfsys@useobject{currentmarker}{}%
\end{pgfscope}%
\end{pgfscope}%
\begin{pgfscope}%
\definecolor{textcolor}{rgb}{0.000000,0.000000,0.000000}%
\pgfsetstrokecolor{textcolor}%
\pgfsetfillcolor{textcolor}%
\pgftext[x=0.763888in, y=1.125978in, left, base]{\color{textcolor}{\rmfamily\fontsize{10.000000}{12.000000}\selectfont\catcode`\^=\active\def^{\ifmmode\sp\else\^{}\fi}\catcode`\%=\active\def%{\%}$\mathdefault{28}$}}%
\end{pgfscope}%
\begin{pgfscope}%
\pgfpathrectangle{\pgfqpoint{1.000000in}{0.440000in}}{\pgfqpoint{6.200000in}{3.080000in}}%
\pgfusepath{clip}%
\pgfsetrectcap%
\pgfsetroundjoin%
\pgfsetlinewidth{0.803000pt}%
\definecolor{currentstroke}{rgb}{0.690196,0.690196,0.690196}%
\pgfsetstrokecolor{currentstroke}%
\pgfsetstrokeopacity{0.300000}%
\pgfsetdash{}{0pt}%
\pgfpathmoveto{\pgfqpoint{1.000000in}{1.622708in}}%
\pgfpathlineto{\pgfqpoint{7.200000in}{1.622708in}}%
\pgfusepath{stroke}%
\end{pgfscope}%
\begin{pgfscope}%
\pgfsetbuttcap%
\pgfsetroundjoin%
\definecolor{currentfill}{rgb}{0.000000,0.000000,0.000000}%
\pgfsetfillcolor{currentfill}%
\pgfsetlinewidth{0.803000pt}%
\definecolor{currentstroke}{rgb}{0.000000,0.000000,0.000000}%
\pgfsetstrokecolor{currentstroke}%
\pgfsetdash{}{0pt}%
\pgfsys@defobject{currentmarker}{\pgfqpoint{-0.048611in}{0.000000in}}{\pgfqpoint{-0.000000in}{0.000000in}}{%
\pgfpathmoveto{\pgfqpoint{-0.000000in}{0.000000in}}%
\pgfpathlineto{\pgfqpoint{-0.048611in}{0.000000in}}%
\pgfusepath{stroke,fill}%
}%
\begin{pgfscope}%
\pgfsys@transformshift{1.000000in}{1.622708in}%
\pgfsys@useobject{currentmarker}{}%
\end{pgfscope}%
\end{pgfscope}%
\begin{pgfscope}%
\definecolor{textcolor}{rgb}{0.000000,0.000000,0.000000}%
\pgfsetstrokecolor{textcolor}%
\pgfsetfillcolor{textcolor}%
\pgftext[x=0.763888in, y=1.574483in, left, base]{\color{textcolor}{\rmfamily\fontsize{10.000000}{12.000000}\selectfont\catcode`\^=\active\def^{\ifmmode\sp\else\^{}\fi}\catcode`\%=\active\def%{\%}$\mathdefault{30}$}}%
\end{pgfscope}%
\begin{pgfscope}%
\pgfpathrectangle{\pgfqpoint{1.000000in}{0.440000in}}{\pgfqpoint{6.200000in}{3.080000in}}%
\pgfusepath{clip}%
\pgfsetrectcap%
\pgfsetroundjoin%
\pgfsetlinewidth{0.803000pt}%
\definecolor{currentstroke}{rgb}{0.690196,0.690196,0.690196}%
\pgfsetstrokecolor{currentstroke}%
\pgfsetstrokeopacity{0.300000}%
\pgfsetdash{}{0pt}%
\pgfpathmoveto{\pgfqpoint{1.000000in}{2.071213in}}%
\pgfpathlineto{\pgfqpoint{7.200000in}{2.071213in}}%
\pgfusepath{stroke}%
\end{pgfscope}%
\begin{pgfscope}%
\pgfsetbuttcap%
\pgfsetroundjoin%
\definecolor{currentfill}{rgb}{0.000000,0.000000,0.000000}%
\pgfsetfillcolor{currentfill}%
\pgfsetlinewidth{0.803000pt}%
\definecolor{currentstroke}{rgb}{0.000000,0.000000,0.000000}%
\pgfsetstrokecolor{currentstroke}%
\pgfsetdash{}{0pt}%
\pgfsys@defobject{currentmarker}{\pgfqpoint{-0.048611in}{0.000000in}}{\pgfqpoint{-0.000000in}{0.000000in}}{%
\pgfpathmoveto{\pgfqpoint{-0.000000in}{0.000000in}}%
\pgfpathlineto{\pgfqpoint{-0.048611in}{0.000000in}}%
\pgfusepath{stroke,fill}%
}%
\begin{pgfscope}%
\pgfsys@transformshift{1.000000in}{2.071213in}%
\pgfsys@useobject{currentmarker}{}%
\end{pgfscope}%
\end{pgfscope}%
\begin{pgfscope}%
\definecolor{textcolor}{rgb}{0.000000,0.000000,0.000000}%
\pgfsetstrokecolor{textcolor}%
\pgfsetfillcolor{textcolor}%
\pgftext[x=0.763888in, y=2.022988in, left, base]{\color{textcolor}{\rmfamily\fontsize{10.000000}{12.000000}\selectfont\catcode`\^=\active\def^{\ifmmode\sp\else\^{}\fi}\catcode`\%=\active\def%{\%}$\mathdefault{32}$}}%
\end{pgfscope}%
\begin{pgfscope}%
\pgfpathrectangle{\pgfqpoint{1.000000in}{0.440000in}}{\pgfqpoint{6.200000in}{3.080000in}}%
\pgfusepath{clip}%
\pgfsetrectcap%
\pgfsetroundjoin%
\pgfsetlinewidth{0.803000pt}%
\definecolor{currentstroke}{rgb}{0.690196,0.690196,0.690196}%
\pgfsetstrokecolor{currentstroke}%
\pgfsetstrokeopacity{0.300000}%
\pgfsetdash{}{0pt}%
\pgfpathmoveto{\pgfqpoint{1.000000in}{2.519718in}}%
\pgfpathlineto{\pgfqpoint{7.200000in}{2.519718in}}%
\pgfusepath{stroke}%
\end{pgfscope}%
\begin{pgfscope}%
\pgfsetbuttcap%
\pgfsetroundjoin%
\definecolor{currentfill}{rgb}{0.000000,0.000000,0.000000}%
\pgfsetfillcolor{currentfill}%
\pgfsetlinewidth{0.803000pt}%
\definecolor{currentstroke}{rgb}{0.000000,0.000000,0.000000}%
\pgfsetstrokecolor{currentstroke}%
\pgfsetdash{}{0pt}%
\pgfsys@defobject{currentmarker}{\pgfqpoint{-0.048611in}{0.000000in}}{\pgfqpoint{-0.000000in}{0.000000in}}{%
\pgfpathmoveto{\pgfqpoint{-0.000000in}{0.000000in}}%
\pgfpathlineto{\pgfqpoint{-0.048611in}{0.000000in}}%
\pgfusepath{stroke,fill}%
}%
\begin{pgfscope}%
\pgfsys@transformshift{1.000000in}{2.519718in}%
\pgfsys@useobject{currentmarker}{}%
\end{pgfscope}%
\end{pgfscope}%
\begin{pgfscope}%
\definecolor{textcolor}{rgb}{0.000000,0.000000,0.000000}%
\pgfsetstrokecolor{textcolor}%
\pgfsetfillcolor{textcolor}%
\pgftext[x=0.763888in, y=2.471493in, left, base]{\color{textcolor}{\rmfamily\fontsize{10.000000}{12.000000}\selectfont\catcode`\^=\active\def^{\ifmmode\sp\else\^{}\fi}\catcode`\%=\active\def%{\%}$\mathdefault{34}$}}%
\end{pgfscope}%
\begin{pgfscope}%
\pgfpathrectangle{\pgfqpoint{1.000000in}{0.440000in}}{\pgfqpoint{6.200000in}{3.080000in}}%
\pgfusepath{clip}%
\pgfsetrectcap%
\pgfsetroundjoin%
\pgfsetlinewidth{0.803000pt}%
\definecolor{currentstroke}{rgb}{0.690196,0.690196,0.690196}%
\pgfsetstrokecolor{currentstroke}%
\pgfsetstrokeopacity{0.300000}%
\pgfsetdash{}{0pt}%
\pgfpathmoveto{\pgfqpoint{1.000000in}{2.968223in}}%
\pgfpathlineto{\pgfqpoint{7.200000in}{2.968223in}}%
\pgfusepath{stroke}%
\end{pgfscope}%
\begin{pgfscope}%
\pgfsetbuttcap%
\pgfsetroundjoin%
\definecolor{currentfill}{rgb}{0.000000,0.000000,0.000000}%
\pgfsetfillcolor{currentfill}%
\pgfsetlinewidth{0.803000pt}%
\definecolor{currentstroke}{rgb}{0.000000,0.000000,0.000000}%
\pgfsetstrokecolor{currentstroke}%
\pgfsetdash{}{0pt}%
\pgfsys@defobject{currentmarker}{\pgfqpoint{-0.048611in}{0.000000in}}{\pgfqpoint{-0.000000in}{0.000000in}}{%
\pgfpathmoveto{\pgfqpoint{-0.000000in}{0.000000in}}%
\pgfpathlineto{\pgfqpoint{-0.048611in}{0.000000in}}%
\pgfusepath{stroke,fill}%
}%
\begin{pgfscope}%
\pgfsys@transformshift{1.000000in}{2.968223in}%
\pgfsys@useobject{currentmarker}{}%
\end{pgfscope}%
\end{pgfscope}%
\begin{pgfscope}%
\definecolor{textcolor}{rgb}{0.000000,0.000000,0.000000}%
\pgfsetstrokecolor{textcolor}%
\pgfsetfillcolor{textcolor}%
\pgftext[x=0.763888in, y=2.919998in, left, base]{\color{textcolor}{\rmfamily\fontsize{10.000000}{12.000000}\selectfont\catcode`\^=\active\def^{\ifmmode\sp\else\^{}\fi}\catcode`\%=\active\def%{\%}$\mathdefault{36}$}}%
\end{pgfscope}%
\begin{pgfscope}%
\pgfpathrectangle{\pgfqpoint{1.000000in}{0.440000in}}{\pgfqpoint{6.200000in}{3.080000in}}%
\pgfusepath{clip}%
\pgfsetrectcap%
\pgfsetroundjoin%
\pgfsetlinewidth{0.803000pt}%
\definecolor{currentstroke}{rgb}{0.690196,0.690196,0.690196}%
\pgfsetstrokecolor{currentstroke}%
\pgfsetstrokeopacity{0.300000}%
\pgfsetdash{}{0pt}%
\pgfpathmoveto{\pgfqpoint{1.000000in}{3.416728in}}%
\pgfpathlineto{\pgfqpoint{7.200000in}{3.416728in}}%
\pgfusepath{stroke}%
\end{pgfscope}%
\begin{pgfscope}%
\pgfsetbuttcap%
\pgfsetroundjoin%
\definecolor{currentfill}{rgb}{0.000000,0.000000,0.000000}%
\pgfsetfillcolor{currentfill}%
\pgfsetlinewidth{0.803000pt}%
\definecolor{currentstroke}{rgb}{0.000000,0.000000,0.000000}%
\pgfsetstrokecolor{currentstroke}%
\pgfsetdash{}{0pt}%
\pgfsys@defobject{currentmarker}{\pgfqpoint{-0.048611in}{0.000000in}}{\pgfqpoint{-0.000000in}{0.000000in}}{%
\pgfpathmoveto{\pgfqpoint{-0.000000in}{0.000000in}}%
\pgfpathlineto{\pgfqpoint{-0.048611in}{0.000000in}}%
\pgfusepath{stroke,fill}%
}%
\begin{pgfscope}%
\pgfsys@transformshift{1.000000in}{3.416728in}%
\pgfsys@useobject{currentmarker}{}%
\end{pgfscope}%
\end{pgfscope}%
\begin{pgfscope}%
\definecolor{textcolor}{rgb}{0.000000,0.000000,0.000000}%
\pgfsetstrokecolor{textcolor}%
\pgfsetfillcolor{textcolor}%
\pgftext[x=0.763888in, y=3.368503in, left, base]{\color{textcolor}{\rmfamily\fontsize{10.000000}{12.000000}\selectfont\catcode`\^=\active\def^{\ifmmode\sp\else\^{}\fi}\catcode`\%=\active\def%{\%}$\mathdefault{38}$}}%
\end{pgfscope}%
\begin{pgfscope}%
\definecolor{textcolor}{rgb}{0.000000,0.000000,0.000000}%
\pgfsetstrokecolor{textcolor}%
\pgfsetfillcolor{textcolor}%
\pgftext[x=0.708333in,y=1.980000in,,bottom,rotate=90.000000]{\color{textcolor}{\rmfamily\fontsize{10.000000}{12.000000}\selectfont\catcode`\^=\active\def^{\ifmmode\sp\else\^{}\fi}\catcode`\%=\active\def%{\%}Compactation attending customers}}%
\end{pgfscope}%
\begin{pgfscope}%
\pgfpathrectangle{\pgfqpoint{1.000000in}{0.440000in}}{\pgfqpoint{6.200000in}{3.080000in}}%
\pgfusepath{clip}%
\pgfsetrectcap%
\pgfsetroundjoin%
\pgfsetlinewidth{1.003750pt}%
\definecolor{currentstroke}{rgb}{0.000000,0.000000,0.000000}%
\pgfsetstrokecolor{currentstroke}%
\pgfsetdash{}{0pt}%
\pgfpathmoveto{\pgfqpoint{1.723333in}{1.691986in}}%
\pgfpathlineto{\pgfqpoint{2.343333in}{1.691986in}}%
\pgfpathlineto{\pgfqpoint{2.343333in}{2.747239in}}%
\pgfpathlineto{\pgfqpoint{1.723333in}{2.747239in}}%
\pgfpathlineto{\pgfqpoint{1.723333in}{1.691986in}}%
\pgfusepath{stroke}%
\end{pgfscope}%
\begin{pgfscope}%
\pgfpathrectangle{\pgfqpoint{1.000000in}{0.440000in}}{\pgfqpoint{6.200000in}{3.080000in}}%
\pgfusepath{clip}%
\pgfsetrectcap%
\pgfsetroundjoin%
\pgfsetlinewidth{1.003750pt}%
\definecolor{currentstroke}{rgb}{0.000000,0.000000,0.000000}%
\pgfsetstrokecolor{currentstroke}%
\pgfsetdash{}{0pt}%
\pgfpathmoveto{\pgfqpoint{2.033333in}{1.691986in}}%
\pgfpathlineto{\pgfqpoint{2.033333in}{0.929159in}}%
\pgfusepath{stroke}%
\end{pgfscope}%
\begin{pgfscope}%
\pgfpathrectangle{\pgfqpoint{1.000000in}{0.440000in}}{\pgfqpoint{6.200000in}{3.080000in}}%
\pgfusepath{clip}%
\pgfsetrectcap%
\pgfsetroundjoin%
\pgfsetlinewidth{1.003750pt}%
\definecolor{currentstroke}{rgb}{0.000000,0.000000,0.000000}%
\pgfsetstrokecolor{currentstroke}%
\pgfsetdash{}{0pt}%
\pgfpathmoveto{\pgfqpoint{2.033333in}{2.747239in}}%
\pgfpathlineto{\pgfqpoint{2.033333in}{3.380000in}}%
\pgfusepath{stroke}%
\end{pgfscope}%
\begin{pgfscope}%
\pgfpathrectangle{\pgfqpoint{1.000000in}{0.440000in}}{\pgfqpoint{6.200000in}{3.080000in}}%
\pgfusepath{clip}%
\pgfsetrectcap%
\pgfsetroundjoin%
\pgfsetlinewidth{1.003750pt}%
\definecolor{currentstroke}{rgb}{0.000000,0.000000,0.000000}%
\pgfsetstrokecolor{currentstroke}%
\pgfsetdash{}{0pt}%
\pgfpathmoveto{\pgfqpoint{1.878333in}{0.929159in}}%
\pgfpathlineto{\pgfqpoint{2.188333in}{0.929159in}}%
\pgfusepath{stroke}%
\end{pgfscope}%
\begin{pgfscope}%
\pgfpathrectangle{\pgfqpoint{1.000000in}{0.440000in}}{\pgfqpoint{6.200000in}{3.080000in}}%
\pgfusepath{clip}%
\pgfsetrectcap%
\pgfsetroundjoin%
\pgfsetlinewidth{1.003750pt}%
\definecolor{currentstroke}{rgb}{0.000000,0.000000,0.000000}%
\pgfsetstrokecolor{currentstroke}%
\pgfsetdash{}{0pt}%
\pgfpathmoveto{\pgfqpoint{1.878333in}{3.380000in}}%
\pgfpathlineto{\pgfqpoint{2.188333in}{3.380000in}}%
\pgfusepath{stroke}%
\end{pgfscope}%
\begin{pgfscope}%
\pgfpathrectangle{\pgfqpoint{1.000000in}{0.440000in}}{\pgfqpoint{6.200000in}{3.080000in}}%
\pgfusepath{clip}%
\pgfsetrectcap%
\pgfsetroundjoin%
\pgfsetlinewidth{1.003750pt}%
\definecolor{currentstroke}{rgb}{0.000000,0.000000,0.000000}%
\pgfsetstrokecolor{currentstroke}%
\pgfsetdash{}{0pt}%
\pgfpathmoveto{\pgfqpoint{3.790000in}{1.844411in}}%
\pgfpathlineto{\pgfqpoint{4.410000in}{1.844411in}}%
\pgfpathlineto{\pgfqpoint{4.410000in}{2.888530in}}%
\pgfpathlineto{\pgfqpoint{3.790000in}{2.888530in}}%
\pgfpathlineto{\pgfqpoint{3.790000in}{1.844411in}}%
\pgfusepath{stroke}%
\end{pgfscope}%
\begin{pgfscope}%
\pgfpathrectangle{\pgfqpoint{1.000000in}{0.440000in}}{\pgfqpoint{6.200000in}{3.080000in}}%
\pgfusepath{clip}%
\pgfsetrectcap%
\pgfsetroundjoin%
\pgfsetlinewidth{1.003750pt}%
\definecolor{currentstroke}{rgb}{0.000000,0.000000,0.000000}%
\pgfsetstrokecolor{currentstroke}%
\pgfsetdash{}{0pt}%
\pgfpathmoveto{\pgfqpoint{4.100000in}{1.844411in}}%
\pgfpathlineto{\pgfqpoint{4.100000in}{1.112665in}}%
\pgfusepath{stroke}%
\end{pgfscope}%
\begin{pgfscope}%
\pgfpathrectangle{\pgfqpoint{1.000000in}{0.440000in}}{\pgfqpoint{6.200000in}{3.080000in}}%
\pgfusepath{clip}%
\pgfsetrectcap%
\pgfsetroundjoin%
\pgfsetlinewidth{1.003750pt}%
\definecolor{currentstroke}{rgb}{0.000000,0.000000,0.000000}%
\pgfsetstrokecolor{currentstroke}%
\pgfsetdash{}{0pt}%
\pgfpathmoveto{\pgfqpoint{4.100000in}{2.888530in}}%
\pgfpathlineto{\pgfqpoint{4.100000in}{3.084793in}}%
\pgfusepath{stroke}%
\end{pgfscope}%
\begin{pgfscope}%
\pgfpathrectangle{\pgfqpoint{1.000000in}{0.440000in}}{\pgfqpoint{6.200000in}{3.080000in}}%
\pgfusepath{clip}%
\pgfsetrectcap%
\pgfsetroundjoin%
\pgfsetlinewidth{1.003750pt}%
\definecolor{currentstroke}{rgb}{0.000000,0.000000,0.000000}%
\pgfsetstrokecolor{currentstroke}%
\pgfsetdash{}{0pt}%
\pgfpathmoveto{\pgfqpoint{3.945000in}{1.112665in}}%
\pgfpathlineto{\pgfqpoint{4.255000in}{1.112665in}}%
\pgfusepath{stroke}%
\end{pgfscope}%
\begin{pgfscope}%
\pgfpathrectangle{\pgfqpoint{1.000000in}{0.440000in}}{\pgfqpoint{6.200000in}{3.080000in}}%
\pgfusepath{clip}%
\pgfsetrectcap%
\pgfsetroundjoin%
\pgfsetlinewidth{1.003750pt}%
\definecolor{currentstroke}{rgb}{0.000000,0.000000,0.000000}%
\pgfsetstrokecolor{currentstroke}%
\pgfsetdash{}{0pt}%
\pgfpathmoveto{\pgfqpoint{3.945000in}{3.084793in}}%
\pgfpathlineto{\pgfqpoint{4.255000in}{3.084793in}}%
\pgfusepath{stroke}%
\end{pgfscope}%
\begin{pgfscope}%
\pgfpathrectangle{\pgfqpoint{1.000000in}{0.440000in}}{\pgfqpoint{6.200000in}{3.080000in}}%
\pgfusepath{clip}%
\pgfsetrectcap%
\pgfsetroundjoin%
\pgfsetlinewidth{1.003750pt}%
\definecolor{currentstroke}{rgb}{0.000000,0.000000,0.000000}%
\pgfsetstrokecolor{currentstroke}%
\pgfsetdash{}{0pt}%
\pgfpathmoveto{\pgfqpoint{5.856667in}{2.093502in}}%
\pgfpathlineto{\pgfqpoint{6.476667in}{2.093502in}}%
\pgfpathlineto{\pgfqpoint{6.476667in}{2.712755in}}%
\pgfpathlineto{\pgfqpoint{5.856667in}{2.712755in}}%
\pgfpathlineto{\pgfqpoint{5.856667in}{2.093502in}}%
\pgfusepath{stroke}%
\end{pgfscope}%
\begin{pgfscope}%
\pgfpathrectangle{\pgfqpoint{1.000000in}{0.440000in}}{\pgfqpoint{6.200000in}{3.080000in}}%
\pgfusepath{clip}%
\pgfsetrectcap%
\pgfsetroundjoin%
\pgfsetlinewidth{1.003750pt}%
\definecolor{currentstroke}{rgb}{0.000000,0.000000,0.000000}%
\pgfsetstrokecolor{currentstroke}%
\pgfsetdash{}{0pt}%
\pgfpathmoveto{\pgfqpoint{6.166667in}{2.093502in}}%
\pgfpathlineto{\pgfqpoint{6.166667in}{1.974383in}}%
\pgfusepath{stroke}%
\end{pgfscope}%
\begin{pgfscope}%
\pgfpathrectangle{\pgfqpoint{1.000000in}{0.440000in}}{\pgfqpoint{6.200000in}{3.080000in}}%
\pgfusepath{clip}%
\pgfsetrectcap%
\pgfsetroundjoin%
\pgfsetlinewidth{1.003750pt}%
\definecolor{currentstroke}{rgb}{0.000000,0.000000,0.000000}%
\pgfsetstrokecolor{currentstroke}%
\pgfsetdash{}{0pt}%
\pgfpathmoveto{\pgfqpoint{6.166667in}{2.712755in}}%
\pgfpathlineto{\pgfqpoint{6.166667in}{3.287636in}}%
\pgfusepath{stroke}%
\end{pgfscope}%
\begin{pgfscope}%
\pgfpathrectangle{\pgfqpoint{1.000000in}{0.440000in}}{\pgfqpoint{6.200000in}{3.080000in}}%
\pgfusepath{clip}%
\pgfsetrectcap%
\pgfsetroundjoin%
\pgfsetlinewidth{1.003750pt}%
\definecolor{currentstroke}{rgb}{0.000000,0.000000,0.000000}%
\pgfsetstrokecolor{currentstroke}%
\pgfsetdash{}{0pt}%
\pgfpathmoveto{\pgfqpoint{6.011667in}{1.974383in}}%
\pgfpathlineto{\pgfqpoint{6.321667in}{1.974383in}}%
\pgfusepath{stroke}%
\end{pgfscope}%
\begin{pgfscope}%
\pgfpathrectangle{\pgfqpoint{1.000000in}{0.440000in}}{\pgfqpoint{6.200000in}{3.080000in}}%
\pgfusepath{clip}%
\pgfsetrectcap%
\pgfsetroundjoin%
\pgfsetlinewidth{1.003750pt}%
\definecolor{currentstroke}{rgb}{0.000000,0.000000,0.000000}%
\pgfsetstrokecolor{currentstroke}%
\pgfsetdash{}{0pt}%
\pgfpathmoveto{\pgfqpoint{6.011667in}{3.287636in}}%
\pgfpathlineto{\pgfqpoint{6.321667in}{3.287636in}}%
\pgfusepath{stroke}%
\end{pgfscope}%
\begin{pgfscope}%
\pgfpathrectangle{\pgfqpoint{1.000000in}{0.440000in}}{\pgfqpoint{6.200000in}{3.080000in}}%
\pgfusepath{clip}%
\pgfsetbuttcap%
\pgfsetroundjoin%
\definecolor{currentfill}{rgb}{0.000000,0.000000,0.000000}%
\pgfsetfillcolor{currentfill}%
\pgfsetfillopacity{0.000000}%
\pgfsetlinewidth{1.003750pt}%
\definecolor{currentstroke}{rgb}{0.000000,0.000000,0.000000}%
\pgfsetstrokecolor{currentstroke}%
\pgfsetdash{}{0pt}%
\pgfsys@defobject{currentmarker}{\pgfqpoint{-0.041667in}{-0.041667in}}{\pgfqpoint{0.041667in}{0.041667in}}{%
\pgfpathmoveto{\pgfqpoint{0.000000in}{-0.041667in}}%
\pgfpathcurveto{\pgfqpoint{0.011050in}{-0.041667in}}{\pgfqpoint{0.021649in}{-0.037276in}}{\pgfqpoint{0.029463in}{-0.029463in}}%
\pgfpathcurveto{\pgfqpoint{0.037276in}{-0.021649in}}{\pgfqpoint{0.041667in}{-0.011050in}}{\pgfqpoint{0.041667in}{0.000000in}}%
\pgfpathcurveto{\pgfqpoint{0.041667in}{0.011050in}}{\pgfqpoint{0.037276in}{0.021649in}}{\pgfqpoint{0.029463in}{0.029463in}}%
\pgfpathcurveto{\pgfqpoint{0.021649in}{0.037276in}}{\pgfqpoint{0.011050in}{0.041667in}}{\pgfqpoint{0.000000in}{0.041667in}}%
\pgfpathcurveto{\pgfqpoint{-0.011050in}{0.041667in}}{\pgfqpoint{-0.021649in}{0.037276in}}{\pgfqpoint{-0.029463in}{0.029463in}}%
\pgfpathcurveto{\pgfqpoint{-0.037276in}{0.021649in}}{\pgfqpoint{-0.041667in}{0.011050in}}{\pgfqpoint{-0.041667in}{0.000000in}}%
\pgfpathcurveto{\pgfqpoint{-0.041667in}{-0.011050in}}{\pgfqpoint{-0.037276in}{-0.021649in}}{\pgfqpoint{-0.029463in}{-0.029463in}}%
\pgfpathcurveto{\pgfqpoint{-0.021649in}{-0.037276in}}{\pgfqpoint{-0.011050in}{-0.041667in}}{\pgfqpoint{0.000000in}{-0.041667in}}%
\pgfpathlineto{\pgfqpoint{0.000000in}{-0.041667in}}%
\pgfpathclose%
\pgfusepath{stroke,fill}%
}%
\begin{pgfscope}%
\pgfsys@transformshift{6.166667in}{0.580000in}%
\pgfsys@useobject{currentmarker}{}%
\end{pgfscope}%
\end{pgfscope}%
\begin{pgfscope}%
\pgfpathrectangle{\pgfqpoint{1.000000in}{0.440000in}}{\pgfqpoint{6.200000in}{3.080000in}}%
\pgfusepath{clip}%
\pgfsetbuttcap%
\pgfsetroundjoin%
\pgfsetlinewidth{1.003750pt}%
\definecolor{currentstroke}{rgb}{1.000000,0.498039,0.054902}%
\pgfsetstrokecolor{currentstroke}%
\pgfsetdash{}{0pt}%
\pgfpathmoveto{\pgfqpoint{1.723333in}{2.278579in}}%
\pgfpathlineto{\pgfqpoint{2.343333in}{2.278579in}}%
\pgfusepath{stroke}%
\end{pgfscope}%
\begin{pgfscope}%
\pgfpathrectangle{\pgfqpoint{1.000000in}{0.440000in}}{\pgfqpoint{6.200000in}{3.080000in}}%
\pgfusepath{clip}%
\pgfsetbuttcap%
\pgfsetroundjoin%
\pgfsetlinewidth{1.003750pt}%
\definecolor{currentstroke}{rgb}{1.000000,0.498039,0.054902}%
\pgfsetstrokecolor{currentstroke}%
\pgfsetdash{}{0pt}%
\pgfpathmoveto{\pgfqpoint{3.790000in}{2.503761in}}%
\pgfpathlineto{\pgfqpoint{4.410000in}{2.503761in}}%
\pgfusepath{stroke}%
\end{pgfscope}%
\begin{pgfscope}%
\pgfpathrectangle{\pgfqpoint{1.000000in}{0.440000in}}{\pgfqpoint{6.200000in}{3.080000in}}%
\pgfusepath{clip}%
\pgfsetbuttcap%
\pgfsetroundjoin%
\pgfsetlinewidth{1.003750pt}%
\definecolor{currentstroke}{rgb}{1.000000,0.498039,0.054902}%
\pgfsetstrokecolor{currentstroke}%
\pgfsetdash{}{0pt}%
\pgfpathmoveto{\pgfqpoint{5.856667in}{2.209221in}}%
\pgfpathlineto{\pgfqpoint{6.476667in}{2.209221in}}%
\pgfusepath{stroke}%
\end{pgfscope}%
\begin{pgfscope}%
\pgfsetrectcap%
\pgfsetmiterjoin%
\pgfsetlinewidth{0.803000pt}%
\definecolor{currentstroke}{rgb}{0.000000,0.000000,0.000000}%
\pgfsetstrokecolor{currentstroke}%
\pgfsetdash{}{0pt}%
\pgfpathmoveto{\pgfqpoint{1.000000in}{0.440000in}}%
\pgfpathlineto{\pgfqpoint{1.000000in}{3.520000in}}%
\pgfusepath{stroke}%
\end{pgfscope}%
\begin{pgfscope}%
\pgfsetrectcap%
\pgfsetmiterjoin%
\pgfsetlinewidth{0.803000pt}%
\definecolor{currentstroke}{rgb}{0.000000,0.000000,0.000000}%
\pgfsetstrokecolor{currentstroke}%
\pgfsetdash{}{0pt}%
\pgfpathmoveto{\pgfqpoint{7.200000in}{0.440000in}}%
\pgfpathlineto{\pgfqpoint{7.200000in}{3.520000in}}%
\pgfusepath{stroke}%
\end{pgfscope}%
\begin{pgfscope}%
\pgfsetrectcap%
\pgfsetmiterjoin%
\pgfsetlinewidth{0.803000pt}%
\definecolor{currentstroke}{rgb}{0.000000,0.000000,0.000000}%
\pgfsetstrokecolor{currentstroke}%
\pgfsetdash{}{0pt}%
\pgfpathmoveto{\pgfqpoint{1.000000in}{0.440000in}}%
\pgfpathlineto{\pgfqpoint{7.200000in}{0.440000in}}%
\pgfusepath{stroke}%
\end{pgfscope}%
\begin{pgfscope}%
\pgfsetrectcap%
\pgfsetmiterjoin%
\pgfsetlinewidth{0.803000pt}%
\definecolor{currentstroke}{rgb}{0.000000,0.000000,0.000000}%
\pgfsetstrokecolor{currentstroke}%
\pgfsetdash{}{0pt}%
\pgfpathmoveto{\pgfqpoint{1.000000in}{3.520000in}}%
\pgfpathlineto{\pgfqpoint{7.200000in}{3.520000in}}%
\pgfusepath{stroke}%
\end{pgfscope}%
\begin{pgfscope}%
\definecolor{textcolor}{rgb}{0.000000,0.000000,0.000000}%
\pgfsetstrokecolor{textcolor}%
\pgfsetfillcolor{textcolor}%
\pgftext[x=4.100000in,y=3.603333in,,base]{\color{textcolor}{\rmfamily\fontsize{12.000000}{14.400000}\selectfont\catcode`\^=\active\def^{\ifmmode\sp\else\^{}\fi}\catcode`\%=\active\def%{\%}CVRP – Total compactation comparison}}%
\end{pgfscope}%
\end{pgfpicture}%
\makeatother%
\endgroup%

%% file: images/CrossingBoxPlot.pgf
\begingroup%
\makeatletter%
\begin{pgfpicture}%
\pgfpathrectangle{\pgfpointorigin}{\pgfqpoint{8.000000in}{4.000000in}}%
\pgfusepath{use as bounding box, clip}%
\begin{pgfscope}%
\pgfsetbuttcap%
\pgfsetmiterjoin%
\definecolor{currentfill}{rgb}{1.000000,1.000000,1.000000}%
\pgfsetfillcolor{currentfill}%
\pgfsetlinewidth{0.000000pt}%
\definecolor{currentstroke}{rgb}{1.000000,1.000000,1.000000}%
\pgfsetstrokecolor{currentstroke}%
\pgfsetdash{}{0pt}%
\pgfpathmoveto{\pgfqpoint{0.000000in}{0.000000in}}%
\pgfpathlineto{\pgfqpoint{8.000000in}{0.000000in}}%
\pgfpathlineto{\pgfqpoint{8.000000in}{4.000000in}}%
\pgfpathlineto{\pgfqpoint{0.000000in}{4.000000in}}%
\pgfpathlineto{\pgfqpoint{0.000000in}{0.000000in}}%
\pgfpathclose%
\pgfusepath{fill}%
\end{pgfscope}%
\begin{pgfscope}%
\pgfsetbuttcap%
\pgfsetmiterjoin%
\definecolor{currentfill}{rgb}{1.000000,1.000000,1.000000}%
\pgfsetfillcolor{currentfill}%
\pgfsetlinewidth{0.000000pt}%
\definecolor{currentstroke}{rgb}{0.000000,0.000000,0.000000}%
\pgfsetstrokecolor{currentstroke}%
\pgfsetstrokeopacity{0.000000}%
\pgfsetdash{}{0pt}%
\pgfpathmoveto{\pgfqpoint{1.000000in}{0.440000in}}%
\pgfpathlineto{\pgfqpoint{7.200000in}{0.440000in}}%
\pgfpathlineto{\pgfqpoint{7.200000in}{3.520000in}}%
\pgfpathlineto{\pgfqpoint{1.000000in}{3.520000in}}%
\pgfpathlineto{\pgfqpoint{1.000000in}{0.440000in}}%
\pgfpathclose%
\pgfusepath{fill}%
\end{pgfscope}%
\begin{pgfscope}%
\pgfsetbuttcap%
\pgfsetroundjoin%
\definecolor{currentfill}{rgb}{0.000000,0.000000,0.000000}%
\pgfsetfillcolor{currentfill}%
\pgfsetlinewidth{0.803000pt}%
\definecolor{currentstroke}{rgb}{0.000000,0.000000,0.000000}%
\pgfsetstrokecolor{currentstroke}%
\pgfsetdash{}{0pt}%
\pgfsys@defobject{currentmarker}{\pgfqpoint{0.000000in}{-0.048611in}}{\pgfqpoint{0.000000in}{0.000000in}}{%
\pgfpathmoveto{\pgfqpoint{0.000000in}{0.000000in}}%
\pgfpathlineto{\pgfqpoint{0.000000in}{-0.048611in}}%
\pgfusepath{stroke,fill}%
}%
\begin{pgfscope}%
\pgfsys@transformshift{2.033333in}{0.440000in}%
\pgfsys@useobject{currentmarker}{}%
\end{pgfscope}%
\end{pgfscope}%
\begin{pgfscope}%
\definecolor{textcolor}{rgb}{0.000000,0.000000,0.000000}%
\pgfsetstrokecolor{textcolor}%
\pgfsetfillcolor{textcolor}%
\pgftext[x=2.033333in,y=0.342778in,,top]{\color{textcolor}{\rmfamily\fontsize{10.000000}{12.000000}\selectfont\catcode`\^=\active\def^{\ifmmode\sp\else\^{}\fi}\catcode`\%=\active\def%{\%}Quantum}}%
\end{pgfscope}%
\begin{pgfscope}%
\pgfsetbuttcap%
\pgfsetroundjoin%
\definecolor{currentfill}{rgb}{0.000000,0.000000,0.000000}%
\pgfsetfillcolor{currentfill}%
\pgfsetlinewidth{0.803000pt}%
\definecolor{currentstroke}{rgb}{0.000000,0.000000,0.000000}%
\pgfsetstrokecolor{currentstroke}%
\pgfsetdash{}{0pt}%
\pgfsys@defobject{currentmarker}{\pgfqpoint{0.000000in}{-0.048611in}}{\pgfqpoint{0.000000in}{0.000000in}}{%
\pgfpathmoveto{\pgfqpoint{0.000000in}{0.000000in}}%
\pgfpathlineto{\pgfqpoint{0.000000in}{-0.048611in}}%
\pgfusepath{stroke,fill}%
}%
\begin{pgfscope}%
\pgfsys@transformshift{4.100000in}{0.440000in}%
\pgfsys@useobject{currentmarker}{}%
\end{pgfscope}%
\end{pgfscope}%
\begin{pgfscope}%
\definecolor{textcolor}{rgb}{0.000000,0.000000,0.000000}%
\pgfsetstrokecolor{textcolor}%
\pgfsetfillcolor{textcolor}%
\pgftext[x=4.100000in,y=0.342778in,,top]{\color{textcolor}{\rmfamily\fontsize{10.000000}{12.000000}\selectfont\catcode`\^=\active\def^{\ifmmode\sp\else\^{}\fi}\catcode`\%=\active\def%{\%}Hybrid}}%
\end{pgfscope}%
\begin{pgfscope}%
\pgfsetbuttcap%
\pgfsetroundjoin%
\definecolor{currentfill}{rgb}{0.000000,0.000000,0.000000}%
\pgfsetfillcolor{currentfill}%
\pgfsetlinewidth{0.803000pt}%
\definecolor{currentstroke}{rgb}{0.000000,0.000000,0.000000}%
\pgfsetstrokecolor{currentstroke}%
\pgfsetdash{}{0pt}%
\pgfsys@defobject{currentmarker}{\pgfqpoint{0.000000in}{-0.048611in}}{\pgfqpoint{0.000000in}{0.000000in}}{%
\pgfpathmoveto{\pgfqpoint{0.000000in}{0.000000in}}%
\pgfpathlineto{\pgfqpoint{0.000000in}{-0.048611in}}%
\pgfusepath{stroke,fill}%
}%
\begin{pgfscope}%
\pgfsys@transformshift{6.166667in}{0.440000in}%
\pgfsys@useobject{currentmarker}{}%
\end{pgfscope}%
\end{pgfscope}%
\begin{pgfscope}%
\definecolor{textcolor}{rgb}{0.000000,0.000000,0.000000}%
\pgfsetstrokecolor{textcolor}%
\pgfsetfillcolor{textcolor}%
\pgftext[x=6.166667in,y=0.342778in,,top]{\color{textcolor}{\rmfamily\fontsize{10.000000}{12.000000}\selectfont\catcode`\^=\active\def^{\ifmmode\sp\else\^{}\fi}\catcode`\%=\active\def%{\%}Classic}}%
\end{pgfscope}%
\begin{pgfscope}%
\pgfpathrectangle{\pgfqpoint{1.000000in}{0.440000in}}{\pgfqpoint{6.200000in}{3.080000in}}%
\pgfusepath{clip}%
\pgfsetrectcap%
\pgfsetroundjoin%
\pgfsetlinewidth{0.803000pt}%
\definecolor{currentstroke}{rgb}{0.690196,0.690196,0.690196}%
\pgfsetstrokecolor{currentstroke}%
\pgfsetstrokeopacity{0.300000}%
\pgfsetdash{}{0pt}%
\pgfpathmoveto{\pgfqpoint{1.000000in}{0.489677in}}%
\pgfpathlineto{\pgfqpoint{7.200000in}{0.489677in}}%
\pgfusepath{stroke}%
\end{pgfscope}%
\begin{pgfscope}%
\pgfsetbuttcap%
\pgfsetroundjoin%
\definecolor{currentfill}{rgb}{0.000000,0.000000,0.000000}%
\pgfsetfillcolor{currentfill}%
\pgfsetlinewidth{0.803000pt}%
\definecolor{currentstroke}{rgb}{0.000000,0.000000,0.000000}%
\pgfsetstrokecolor{currentstroke}%
\pgfsetdash{}{0pt}%
\pgfsys@defobject{currentmarker}{\pgfqpoint{-0.048611in}{0.000000in}}{\pgfqpoint{-0.000000in}{0.000000in}}{%
\pgfpathmoveto{\pgfqpoint{-0.000000in}{0.000000in}}%
\pgfpathlineto{\pgfqpoint{-0.048611in}{0.000000in}}%
\pgfusepath{stroke,fill}%
}%
\begin{pgfscope}%
\pgfsys@transformshift{1.000000in}{0.489677in}%
\pgfsys@useobject{currentmarker}{}%
\end{pgfscope}%
\end{pgfscope}%
\begin{pgfscope}%
\definecolor{textcolor}{rgb}{0.000000,0.000000,0.000000}%
\pgfsetstrokecolor{textcolor}%
\pgfsetfillcolor{textcolor}%
\pgftext[x=0.763888in, y=0.441452in, left, base]{\color{textcolor}{\rmfamily\fontsize{10.000000}{12.000000}\selectfont\catcode`\^=\active\def^{\ifmmode\sp\else\^{}\fi}\catcode`\%=\active\def%{\%}$\mathdefault{10}$}}%
\end{pgfscope}%
\begin{pgfscope}%
\pgfpathrectangle{\pgfqpoint{1.000000in}{0.440000in}}{\pgfqpoint{6.200000in}{3.080000in}}%
\pgfusepath{clip}%
\pgfsetrectcap%
\pgfsetroundjoin%
\pgfsetlinewidth{0.803000pt}%
\definecolor{currentstroke}{rgb}{0.690196,0.690196,0.690196}%
\pgfsetstrokecolor{currentstroke}%
\pgfsetstrokeopacity{0.300000}%
\pgfsetdash{}{0pt}%
\pgfpathmoveto{\pgfqpoint{1.000000in}{0.850968in}}%
\pgfpathlineto{\pgfqpoint{7.200000in}{0.850968in}}%
\pgfusepath{stroke}%
\end{pgfscope}%
\begin{pgfscope}%
\pgfsetbuttcap%
\pgfsetroundjoin%
\definecolor{currentfill}{rgb}{0.000000,0.000000,0.000000}%
\pgfsetfillcolor{currentfill}%
\pgfsetlinewidth{0.803000pt}%
\definecolor{currentstroke}{rgb}{0.000000,0.000000,0.000000}%
\pgfsetstrokecolor{currentstroke}%
\pgfsetdash{}{0pt}%
\pgfsys@defobject{currentmarker}{\pgfqpoint{-0.048611in}{0.000000in}}{\pgfqpoint{-0.000000in}{0.000000in}}{%
\pgfpathmoveto{\pgfqpoint{-0.000000in}{0.000000in}}%
\pgfpathlineto{\pgfqpoint{-0.048611in}{0.000000in}}%
\pgfusepath{stroke,fill}%
}%
\begin{pgfscope}%
\pgfsys@transformshift{1.000000in}{0.850968in}%
\pgfsys@useobject{currentmarker}{}%
\end{pgfscope}%
\end{pgfscope}%
\begin{pgfscope}%
\definecolor{textcolor}{rgb}{0.000000,0.000000,0.000000}%
\pgfsetstrokecolor{textcolor}%
\pgfsetfillcolor{textcolor}%
\pgftext[x=0.763888in, y=0.802742in, left, base]{\color{textcolor}{\rmfamily\fontsize{10.000000}{12.000000}\selectfont\catcode`\^=\active\def^{\ifmmode\sp\else\^{}\fi}\catcode`\%=\active\def%{\%}$\mathdefault{12}$}}%
\end{pgfscope}%
\begin{pgfscope}%
\pgfpathrectangle{\pgfqpoint{1.000000in}{0.440000in}}{\pgfqpoint{6.200000in}{3.080000in}}%
\pgfusepath{clip}%
\pgfsetrectcap%
\pgfsetroundjoin%
\pgfsetlinewidth{0.803000pt}%
\definecolor{currentstroke}{rgb}{0.690196,0.690196,0.690196}%
\pgfsetstrokecolor{currentstroke}%
\pgfsetstrokeopacity{0.300000}%
\pgfsetdash{}{0pt}%
\pgfpathmoveto{\pgfqpoint{1.000000in}{1.212258in}}%
\pgfpathlineto{\pgfqpoint{7.200000in}{1.212258in}}%
\pgfusepath{stroke}%
\end{pgfscope}%
\begin{pgfscope}%
\pgfsetbuttcap%
\pgfsetroundjoin%
\definecolor{currentfill}{rgb}{0.000000,0.000000,0.000000}%
\pgfsetfillcolor{currentfill}%
\pgfsetlinewidth{0.803000pt}%
\definecolor{currentstroke}{rgb}{0.000000,0.000000,0.000000}%
\pgfsetstrokecolor{currentstroke}%
\pgfsetdash{}{0pt}%
\pgfsys@defobject{currentmarker}{\pgfqpoint{-0.048611in}{0.000000in}}{\pgfqpoint{-0.000000in}{0.000000in}}{%
\pgfpathmoveto{\pgfqpoint{-0.000000in}{0.000000in}}%
\pgfpathlineto{\pgfqpoint{-0.048611in}{0.000000in}}%
\pgfusepath{stroke,fill}%
}%
\begin{pgfscope}%
\pgfsys@transformshift{1.000000in}{1.212258in}%
\pgfsys@useobject{currentmarker}{}%
\end{pgfscope}%
\end{pgfscope}%
\begin{pgfscope}%
\definecolor{textcolor}{rgb}{0.000000,0.000000,0.000000}%
\pgfsetstrokecolor{textcolor}%
\pgfsetfillcolor{textcolor}%
\pgftext[x=0.763888in, y=1.164033in, left, base]{\color{textcolor}{\rmfamily\fontsize{10.000000}{12.000000}\selectfont\catcode`\^=\active\def^{\ifmmode\sp\else\^{}\fi}\catcode`\%=\active\def%{\%}$\mathdefault{14}$}}%
\end{pgfscope}%
\begin{pgfscope}%
\pgfpathrectangle{\pgfqpoint{1.000000in}{0.440000in}}{\pgfqpoint{6.200000in}{3.080000in}}%
\pgfusepath{clip}%
\pgfsetrectcap%
\pgfsetroundjoin%
\pgfsetlinewidth{0.803000pt}%
\definecolor{currentstroke}{rgb}{0.690196,0.690196,0.690196}%
\pgfsetstrokecolor{currentstroke}%
\pgfsetstrokeopacity{0.300000}%
\pgfsetdash{}{0pt}%
\pgfpathmoveto{\pgfqpoint{1.000000in}{1.573548in}}%
\pgfpathlineto{\pgfqpoint{7.200000in}{1.573548in}}%
\pgfusepath{stroke}%
\end{pgfscope}%
\begin{pgfscope}%
\pgfsetbuttcap%
\pgfsetroundjoin%
\definecolor{currentfill}{rgb}{0.000000,0.000000,0.000000}%
\pgfsetfillcolor{currentfill}%
\pgfsetlinewidth{0.803000pt}%
\definecolor{currentstroke}{rgb}{0.000000,0.000000,0.000000}%
\pgfsetstrokecolor{currentstroke}%
\pgfsetdash{}{0pt}%
\pgfsys@defobject{currentmarker}{\pgfqpoint{-0.048611in}{0.000000in}}{\pgfqpoint{-0.000000in}{0.000000in}}{%
\pgfpathmoveto{\pgfqpoint{-0.000000in}{0.000000in}}%
\pgfpathlineto{\pgfqpoint{-0.048611in}{0.000000in}}%
\pgfusepath{stroke,fill}%
}%
\begin{pgfscope}%
\pgfsys@transformshift{1.000000in}{1.573548in}%
\pgfsys@useobject{currentmarker}{}%
\end{pgfscope}%
\end{pgfscope}%
\begin{pgfscope}%
\definecolor{textcolor}{rgb}{0.000000,0.000000,0.000000}%
\pgfsetstrokecolor{textcolor}%
\pgfsetfillcolor{textcolor}%
\pgftext[x=0.763888in, y=1.525323in, left, base]{\color{textcolor}{\rmfamily\fontsize{10.000000}{12.000000}\selectfont\catcode`\^=\active\def^{\ifmmode\sp\else\^{}\fi}\catcode`\%=\active\def%{\%}$\mathdefault{16}$}}%
\end{pgfscope}%
\begin{pgfscope}%
\pgfpathrectangle{\pgfqpoint{1.000000in}{0.440000in}}{\pgfqpoint{6.200000in}{3.080000in}}%
\pgfusepath{clip}%
\pgfsetrectcap%
\pgfsetroundjoin%
\pgfsetlinewidth{0.803000pt}%
\definecolor{currentstroke}{rgb}{0.690196,0.690196,0.690196}%
\pgfsetstrokecolor{currentstroke}%
\pgfsetstrokeopacity{0.300000}%
\pgfsetdash{}{0pt}%
\pgfpathmoveto{\pgfqpoint{1.000000in}{1.934839in}}%
\pgfpathlineto{\pgfqpoint{7.200000in}{1.934839in}}%
\pgfusepath{stroke}%
\end{pgfscope}%
\begin{pgfscope}%
\pgfsetbuttcap%
\pgfsetroundjoin%
\definecolor{currentfill}{rgb}{0.000000,0.000000,0.000000}%
\pgfsetfillcolor{currentfill}%
\pgfsetlinewidth{0.803000pt}%
\definecolor{currentstroke}{rgb}{0.000000,0.000000,0.000000}%
\pgfsetstrokecolor{currentstroke}%
\pgfsetdash{}{0pt}%
\pgfsys@defobject{currentmarker}{\pgfqpoint{-0.048611in}{0.000000in}}{\pgfqpoint{-0.000000in}{0.000000in}}{%
\pgfpathmoveto{\pgfqpoint{-0.000000in}{0.000000in}}%
\pgfpathlineto{\pgfqpoint{-0.048611in}{0.000000in}}%
\pgfusepath{stroke,fill}%
}%
\begin{pgfscope}%
\pgfsys@transformshift{1.000000in}{1.934839in}%
\pgfsys@useobject{currentmarker}{}%
\end{pgfscope}%
\end{pgfscope}%
\begin{pgfscope}%
\definecolor{textcolor}{rgb}{0.000000,0.000000,0.000000}%
\pgfsetstrokecolor{textcolor}%
\pgfsetfillcolor{textcolor}%
\pgftext[x=0.763888in, y=1.886613in, left, base]{\color{textcolor}{\rmfamily\fontsize{10.000000}{12.000000}\selectfont\catcode`\^=\active\def^{\ifmmode\sp\else\^{}\fi}\catcode`\%=\active\def%{\%}$\mathdefault{18}$}}%
\end{pgfscope}%
\begin{pgfscope}%
\pgfpathrectangle{\pgfqpoint{1.000000in}{0.440000in}}{\pgfqpoint{6.200000in}{3.080000in}}%
\pgfusepath{clip}%
\pgfsetrectcap%
\pgfsetroundjoin%
\pgfsetlinewidth{0.803000pt}%
\definecolor{currentstroke}{rgb}{0.690196,0.690196,0.690196}%
\pgfsetstrokecolor{currentstroke}%
\pgfsetstrokeopacity{0.300000}%
\pgfsetdash{}{0pt}%
\pgfpathmoveto{\pgfqpoint{1.000000in}{2.296129in}}%
\pgfpathlineto{\pgfqpoint{7.200000in}{2.296129in}}%
\pgfusepath{stroke}%
\end{pgfscope}%
\begin{pgfscope}%
\pgfsetbuttcap%
\pgfsetroundjoin%
\definecolor{currentfill}{rgb}{0.000000,0.000000,0.000000}%
\pgfsetfillcolor{currentfill}%
\pgfsetlinewidth{0.803000pt}%
\definecolor{currentstroke}{rgb}{0.000000,0.000000,0.000000}%
\pgfsetstrokecolor{currentstroke}%
\pgfsetdash{}{0pt}%
\pgfsys@defobject{currentmarker}{\pgfqpoint{-0.048611in}{0.000000in}}{\pgfqpoint{-0.000000in}{0.000000in}}{%
\pgfpathmoveto{\pgfqpoint{-0.000000in}{0.000000in}}%
\pgfpathlineto{\pgfqpoint{-0.048611in}{0.000000in}}%
\pgfusepath{stroke,fill}%
}%
\begin{pgfscope}%
\pgfsys@transformshift{1.000000in}{2.296129in}%
\pgfsys@useobject{currentmarker}{}%
\end{pgfscope}%
\end{pgfscope}%
\begin{pgfscope}%
\definecolor{textcolor}{rgb}{0.000000,0.000000,0.000000}%
\pgfsetstrokecolor{textcolor}%
\pgfsetfillcolor{textcolor}%
\pgftext[x=0.763888in, y=2.247904in, left, base]{\color{textcolor}{\rmfamily\fontsize{10.000000}{12.000000}\selectfont\catcode`\^=\active\def^{\ifmmode\sp\else\^{}\fi}\catcode`\%=\active\def%{\%}$\mathdefault{20}$}}%
\end{pgfscope}%
\begin{pgfscope}%
\pgfpathrectangle{\pgfqpoint{1.000000in}{0.440000in}}{\pgfqpoint{6.200000in}{3.080000in}}%
\pgfusepath{clip}%
\pgfsetrectcap%
\pgfsetroundjoin%
\pgfsetlinewidth{0.803000pt}%
\definecolor{currentstroke}{rgb}{0.690196,0.690196,0.690196}%
\pgfsetstrokecolor{currentstroke}%
\pgfsetstrokeopacity{0.300000}%
\pgfsetdash{}{0pt}%
\pgfpathmoveto{\pgfqpoint{1.000000in}{2.657419in}}%
\pgfpathlineto{\pgfqpoint{7.200000in}{2.657419in}}%
\pgfusepath{stroke}%
\end{pgfscope}%
\begin{pgfscope}%
\pgfsetbuttcap%
\pgfsetroundjoin%
\definecolor{currentfill}{rgb}{0.000000,0.000000,0.000000}%
\pgfsetfillcolor{currentfill}%
\pgfsetlinewidth{0.803000pt}%
\definecolor{currentstroke}{rgb}{0.000000,0.000000,0.000000}%
\pgfsetstrokecolor{currentstroke}%
\pgfsetdash{}{0pt}%
\pgfsys@defobject{currentmarker}{\pgfqpoint{-0.048611in}{0.000000in}}{\pgfqpoint{-0.000000in}{0.000000in}}{%
\pgfpathmoveto{\pgfqpoint{-0.000000in}{0.000000in}}%
\pgfpathlineto{\pgfqpoint{-0.048611in}{0.000000in}}%
\pgfusepath{stroke,fill}%
}%
\begin{pgfscope}%
\pgfsys@transformshift{1.000000in}{2.657419in}%
\pgfsys@useobject{currentmarker}{}%
\end{pgfscope}%
\end{pgfscope}%
\begin{pgfscope}%
\definecolor{textcolor}{rgb}{0.000000,0.000000,0.000000}%
\pgfsetstrokecolor{textcolor}%
\pgfsetfillcolor{textcolor}%
\pgftext[x=0.763888in, y=2.609194in, left, base]{\color{textcolor}{\rmfamily\fontsize{10.000000}{12.000000}\selectfont\catcode`\^=\active\def^{\ifmmode\sp\else\^{}\fi}\catcode`\%=\active\def%{\%}$\mathdefault{22}$}}%
\end{pgfscope}%
\begin{pgfscope}%
\pgfpathrectangle{\pgfqpoint{1.000000in}{0.440000in}}{\pgfqpoint{6.200000in}{3.080000in}}%
\pgfusepath{clip}%
\pgfsetrectcap%
\pgfsetroundjoin%
\pgfsetlinewidth{0.803000pt}%
\definecolor{currentstroke}{rgb}{0.690196,0.690196,0.690196}%
\pgfsetstrokecolor{currentstroke}%
\pgfsetstrokeopacity{0.300000}%
\pgfsetdash{}{0pt}%
\pgfpathmoveto{\pgfqpoint{1.000000in}{3.018710in}}%
\pgfpathlineto{\pgfqpoint{7.200000in}{3.018710in}}%
\pgfusepath{stroke}%
\end{pgfscope}%
\begin{pgfscope}%
\pgfsetbuttcap%
\pgfsetroundjoin%
\definecolor{currentfill}{rgb}{0.000000,0.000000,0.000000}%
\pgfsetfillcolor{currentfill}%
\pgfsetlinewidth{0.803000pt}%
\definecolor{currentstroke}{rgb}{0.000000,0.000000,0.000000}%
\pgfsetstrokecolor{currentstroke}%
\pgfsetdash{}{0pt}%
\pgfsys@defobject{currentmarker}{\pgfqpoint{-0.048611in}{0.000000in}}{\pgfqpoint{-0.000000in}{0.000000in}}{%
\pgfpathmoveto{\pgfqpoint{-0.000000in}{0.000000in}}%
\pgfpathlineto{\pgfqpoint{-0.048611in}{0.000000in}}%
\pgfusepath{stroke,fill}%
}%
\begin{pgfscope}%
\pgfsys@transformshift{1.000000in}{3.018710in}%
\pgfsys@useobject{currentmarker}{}%
\end{pgfscope}%
\end{pgfscope}%
\begin{pgfscope}%
\definecolor{textcolor}{rgb}{0.000000,0.000000,0.000000}%
\pgfsetstrokecolor{textcolor}%
\pgfsetfillcolor{textcolor}%
\pgftext[x=0.763888in, y=2.970484in, left, base]{\color{textcolor}{\rmfamily\fontsize{10.000000}{12.000000}\selectfont\catcode`\^=\active\def^{\ifmmode\sp\else\^{}\fi}\catcode`\%=\active\def%{\%}$\mathdefault{24}$}}%
\end{pgfscope}%
\begin{pgfscope}%
\pgfpathrectangle{\pgfqpoint{1.000000in}{0.440000in}}{\pgfqpoint{6.200000in}{3.080000in}}%
\pgfusepath{clip}%
\pgfsetrectcap%
\pgfsetroundjoin%
\pgfsetlinewidth{0.803000pt}%
\definecolor{currentstroke}{rgb}{0.690196,0.690196,0.690196}%
\pgfsetstrokecolor{currentstroke}%
\pgfsetstrokeopacity{0.300000}%
\pgfsetdash{}{0pt}%
\pgfpathmoveto{\pgfqpoint{1.000000in}{3.380000in}}%
\pgfpathlineto{\pgfqpoint{7.200000in}{3.380000in}}%
\pgfusepath{stroke}%
\end{pgfscope}%
\begin{pgfscope}%
\pgfsetbuttcap%
\pgfsetroundjoin%
\definecolor{currentfill}{rgb}{0.000000,0.000000,0.000000}%
\pgfsetfillcolor{currentfill}%
\pgfsetlinewidth{0.803000pt}%
\definecolor{currentstroke}{rgb}{0.000000,0.000000,0.000000}%
\pgfsetstrokecolor{currentstroke}%
\pgfsetdash{}{0pt}%
\pgfsys@defobject{currentmarker}{\pgfqpoint{-0.048611in}{0.000000in}}{\pgfqpoint{-0.000000in}{0.000000in}}{%
\pgfpathmoveto{\pgfqpoint{-0.000000in}{0.000000in}}%
\pgfpathlineto{\pgfqpoint{-0.048611in}{0.000000in}}%
\pgfusepath{stroke,fill}%
}%
\begin{pgfscope}%
\pgfsys@transformshift{1.000000in}{3.380000in}%
\pgfsys@useobject{currentmarker}{}%
\end{pgfscope}%
\end{pgfscope}%
\begin{pgfscope}%
\definecolor{textcolor}{rgb}{0.000000,0.000000,0.000000}%
\pgfsetstrokecolor{textcolor}%
\pgfsetfillcolor{textcolor}%
\pgftext[x=0.763888in, y=3.331775in, left, base]{\color{textcolor}{\rmfamily\fontsize{10.000000}{12.000000}\selectfont\catcode`\^=\active\def^{\ifmmode\sp\else\^{}\fi}\catcode`\%=\active\def%{\%}$\mathdefault{26}$}}%
\end{pgfscope}%
\begin{pgfscope}%
\definecolor{textcolor}{rgb}{0.000000,0.000000,0.000000}%
\pgfsetstrokecolor{textcolor}%
\pgfsetfillcolor{textcolor}%
\pgftext[x=0.708333in,y=1.980000in,,bottom,rotate=90.000000]{\color{textcolor}{\rmfamily\fontsize{10.000000}{12.000000}\selectfont\catcode`\^=\active\def^{\ifmmode\sp\else\^{}\fi}\catcode`\%=\active\def%{\%}Average vehicle crossings}}%
\end{pgfscope}%
\begin{pgfscope}%
\pgfpathrectangle{\pgfqpoint{1.000000in}{0.440000in}}{\pgfqpoint{6.200000in}{3.080000in}}%
\pgfusepath{clip}%
\pgfsetrectcap%
\pgfsetroundjoin%
\pgfsetlinewidth{1.003750pt}%
\definecolor{currentstroke}{rgb}{0.000000,0.000000,0.000000}%
\pgfsetstrokecolor{currentstroke}%
\pgfsetdash{}{0pt}%
\pgfpathmoveto{\pgfqpoint{1.723333in}{0.986452in}}%
\pgfpathlineto{\pgfqpoint{2.343333in}{0.986452in}}%
\pgfpathlineto{\pgfqpoint{2.343333in}{1.709032in}}%
\pgfpathlineto{\pgfqpoint{1.723333in}{1.709032in}}%
\pgfpathlineto{\pgfqpoint{1.723333in}{0.986452in}}%
\pgfusepath{stroke}%
\end{pgfscope}%
\begin{pgfscope}%
\pgfpathrectangle{\pgfqpoint{1.000000in}{0.440000in}}{\pgfqpoint{6.200000in}{3.080000in}}%
\pgfusepath{clip}%
\pgfsetrectcap%
\pgfsetroundjoin%
\pgfsetlinewidth{1.003750pt}%
\definecolor{currentstroke}{rgb}{0.000000,0.000000,0.000000}%
\pgfsetstrokecolor{currentstroke}%
\pgfsetdash{}{0pt}%
\pgfpathmoveto{\pgfqpoint{2.033333in}{0.986452in}}%
\pgfpathlineto{\pgfqpoint{2.033333in}{0.760645in}}%
\pgfusepath{stroke}%
\end{pgfscope}%
\begin{pgfscope}%
\pgfpathrectangle{\pgfqpoint{1.000000in}{0.440000in}}{\pgfqpoint{6.200000in}{3.080000in}}%
\pgfusepath{clip}%
\pgfsetrectcap%
\pgfsetroundjoin%
\pgfsetlinewidth{1.003750pt}%
\definecolor{currentstroke}{rgb}{0.000000,0.000000,0.000000}%
\pgfsetstrokecolor{currentstroke}%
\pgfsetdash{}{0pt}%
\pgfpathmoveto{\pgfqpoint{2.033333in}{1.709032in}}%
\pgfpathlineto{\pgfqpoint{2.033333in}{2.567097in}}%
\pgfusepath{stroke}%
\end{pgfscope}%
\begin{pgfscope}%
\pgfpathrectangle{\pgfqpoint{1.000000in}{0.440000in}}{\pgfqpoint{6.200000in}{3.080000in}}%
\pgfusepath{clip}%
\pgfsetrectcap%
\pgfsetroundjoin%
\pgfsetlinewidth{1.003750pt}%
\definecolor{currentstroke}{rgb}{0.000000,0.000000,0.000000}%
\pgfsetstrokecolor{currentstroke}%
\pgfsetdash{}{0pt}%
\pgfpathmoveto{\pgfqpoint{1.878333in}{0.760645in}}%
\pgfpathlineto{\pgfqpoint{2.188333in}{0.760645in}}%
\pgfusepath{stroke}%
\end{pgfscope}%
\begin{pgfscope}%
\pgfpathrectangle{\pgfqpoint{1.000000in}{0.440000in}}{\pgfqpoint{6.200000in}{3.080000in}}%
\pgfusepath{clip}%
\pgfsetrectcap%
\pgfsetroundjoin%
\pgfsetlinewidth{1.003750pt}%
\definecolor{currentstroke}{rgb}{0.000000,0.000000,0.000000}%
\pgfsetstrokecolor{currentstroke}%
\pgfsetdash{}{0pt}%
\pgfpathmoveto{\pgfqpoint{1.878333in}{2.567097in}}%
\pgfpathlineto{\pgfqpoint{2.188333in}{2.567097in}}%
\pgfusepath{stroke}%
\end{pgfscope}%
\begin{pgfscope}%
\pgfpathrectangle{\pgfqpoint{1.000000in}{0.440000in}}{\pgfqpoint{6.200000in}{3.080000in}}%
\pgfusepath{clip}%
\pgfsetrectcap%
\pgfsetroundjoin%
\pgfsetlinewidth{1.003750pt}%
\definecolor{currentstroke}{rgb}{0.000000,0.000000,0.000000}%
\pgfsetstrokecolor{currentstroke}%
\pgfsetdash{}{0pt}%
\pgfpathmoveto{\pgfqpoint{3.790000in}{0.873548in}}%
\pgfpathlineto{\pgfqpoint{4.410000in}{0.873548in}}%
\pgfpathlineto{\pgfqpoint{4.410000in}{1.641290in}}%
\pgfpathlineto{\pgfqpoint{3.790000in}{1.641290in}}%
\pgfpathlineto{\pgfqpoint{3.790000in}{0.873548in}}%
\pgfusepath{stroke}%
\end{pgfscope}%
\begin{pgfscope}%
\pgfpathrectangle{\pgfqpoint{1.000000in}{0.440000in}}{\pgfqpoint{6.200000in}{3.080000in}}%
\pgfusepath{clip}%
\pgfsetrectcap%
\pgfsetroundjoin%
\pgfsetlinewidth{1.003750pt}%
\definecolor{currentstroke}{rgb}{0.000000,0.000000,0.000000}%
\pgfsetstrokecolor{currentstroke}%
\pgfsetdash{}{0pt}%
\pgfpathmoveto{\pgfqpoint{4.100000in}{0.873548in}}%
\pgfpathlineto{\pgfqpoint{4.100000in}{0.580000in}}%
\pgfusepath{stroke}%
\end{pgfscope}%
\begin{pgfscope}%
\pgfpathrectangle{\pgfqpoint{1.000000in}{0.440000in}}{\pgfqpoint{6.200000in}{3.080000in}}%
\pgfusepath{clip}%
\pgfsetrectcap%
\pgfsetroundjoin%
\pgfsetlinewidth{1.003750pt}%
\definecolor{currentstroke}{rgb}{0.000000,0.000000,0.000000}%
\pgfsetstrokecolor{currentstroke}%
\pgfsetdash{}{0pt}%
\pgfpathmoveto{\pgfqpoint{4.100000in}{1.641290in}}%
\pgfpathlineto{\pgfqpoint{4.100000in}{2.115484in}}%
\pgfusepath{stroke}%
\end{pgfscope}%
\begin{pgfscope}%
\pgfpathrectangle{\pgfqpoint{1.000000in}{0.440000in}}{\pgfqpoint{6.200000in}{3.080000in}}%
\pgfusepath{clip}%
\pgfsetrectcap%
\pgfsetroundjoin%
\pgfsetlinewidth{1.003750pt}%
\definecolor{currentstroke}{rgb}{0.000000,0.000000,0.000000}%
\pgfsetstrokecolor{currentstroke}%
\pgfsetdash{}{0pt}%
\pgfpathmoveto{\pgfqpoint{3.945000in}{0.580000in}}%
\pgfpathlineto{\pgfqpoint{4.255000in}{0.580000in}}%
\pgfusepath{stroke}%
\end{pgfscope}%
\begin{pgfscope}%
\pgfpathrectangle{\pgfqpoint{1.000000in}{0.440000in}}{\pgfqpoint{6.200000in}{3.080000in}}%
\pgfusepath{clip}%
\pgfsetrectcap%
\pgfsetroundjoin%
\pgfsetlinewidth{1.003750pt}%
\definecolor{currentstroke}{rgb}{0.000000,0.000000,0.000000}%
\pgfsetstrokecolor{currentstroke}%
\pgfsetdash{}{0pt}%
\pgfpathmoveto{\pgfqpoint{3.945000in}{2.115484in}}%
\pgfpathlineto{\pgfqpoint{4.255000in}{2.115484in}}%
\pgfusepath{stroke}%
\end{pgfscope}%
\begin{pgfscope}%
\pgfpathrectangle{\pgfqpoint{1.000000in}{0.440000in}}{\pgfqpoint{6.200000in}{3.080000in}}%
\pgfusepath{clip}%
\pgfsetrectcap%
\pgfsetroundjoin%
\pgfsetlinewidth{1.003750pt}%
\definecolor{currentstroke}{rgb}{0.000000,0.000000,0.000000}%
\pgfsetstrokecolor{currentstroke}%
\pgfsetdash{}{0pt}%
\pgfpathmoveto{\pgfqpoint{5.856667in}{1.054194in}}%
\pgfpathlineto{\pgfqpoint{6.476667in}{1.054194in}}%
\pgfpathlineto{\pgfqpoint{6.476667in}{2.702581in}}%
\pgfpathlineto{\pgfqpoint{5.856667in}{2.702581in}}%
\pgfpathlineto{\pgfqpoint{5.856667in}{1.054194in}}%
\pgfusepath{stroke}%
\end{pgfscope}%
\begin{pgfscope}%
\pgfpathrectangle{\pgfqpoint{1.000000in}{0.440000in}}{\pgfqpoint{6.200000in}{3.080000in}}%
\pgfusepath{clip}%
\pgfsetrectcap%
\pgfsetroundjoin%
\pgfsetlinewidth{1.003750pt}%
\definecolor{currentstroke}{rgb}{0.000000,0.000000,0.000000}%
\pgfsetstrokecolor{currentstroke}%
\pgfsetdash{}{0pt}%
\pgfpathmoveto{\pgfqpoint{6.166667in}{1.054194in}}%
\pgfpathlineto{\pgfqpoint{6.166667in}{0.670323in}}%
\pgfusepath{stroke}%
\end{pgfscope}%
\begin{pgfscope}%
\pgfpathrectangle{\pgfqpoint{1.000000in}{0.440000in}}{\pgfqpoint{6.200000in}{3.080000in}}%
\pgfusepath{clip}%
\pgfsetrectcap%
\pgfsetroundjoin%
\pgfsetlinewidth{1.003750pt}%
\definecolor{currentstroke}{rgb}{0.000000,0.000000,0.000000}%
\pgfsetstrokecolor{currentstroke}%
\pgfsetdash{}{0pt}%
\pgfpathmoveto{\pgfqpoint{6.166667in}{2.702581in}}%
\pgfpathlineto{\pgfqpoint{6.166667in}{3.380000in}}%
\pgfusepath{stroke}%
\end{pgfscope}%
\begin{pgfscope}%
\pgfpathrectangle{\pgfqpoint{1.000000in}{0.440000in}}{\pgfqpoint{6.200000in}{3.080000in}}%
\pgfusepath{clip}%
\pgfsetrectcap%
\pgfsetroundjoin%
\pgfsetlinewidth{1.003750pt}%
\definecolor{currentstroke}{rgb}{0.000000,0.000000,0.000000}%
\pgfsetstrokecolor{currentstroke}%
\pgfsetdash{}{0pt}%
\pgfpathmoveto{\pgfqpoint{6.011667in}{0.670323in}}%
\pgfpathlineto{\pgfqpoint{6.321667in}{0.670323in}}%
\pgfusepath{stroke}%
\end{pgfscope}%
\begin{pgfscope}%
\pgfpathrectangle{\pgfqpoint{1.000000in}{0.440000in}}{\pgfqpoint{6.200000in}{3.080000in}}%
\pgfusepath{clip}%
\pgfsetrectcap%
\pgfsetroundjoin%
\pgfsetlinewidth{1.003750pt}%
\definecolor{currentstroke}{rgb}{0.000000,0.000000,0.000000}%
\pgfsetstrokecolor{currentstroke}%
\pgfsetdash{}{0pt}%
\pgfpathmoveto{\pgfqpoint{6.011667in}{3.380000in}}%
\pgfpathlineto{\pgfqpoint{6.321667in}{3.380000in}}%
\pgfusepath{stroke}%
\end{pgfscope}%
\begin{pgfscope}%
\pgfpathrectangle{\pgfqpoint{1.000000in}{0.440000in}}{\pgfqpoint{6.200000in}{3.080000in}}%
\pgfusepath{clip}%
\pgfsetbuttcap%
\pgfsetroundjoin%
\pgfsetlinewidth{1.003750pt}%
\definecolor{currentstroke}{rgb}{1.000000,0.498039,0.054902}%
\pgfsetstrokecolor{currentstroke}%
\pgfsetdash{}{0pt}%
\pgfpathmoveto{\pgfqpoint{1.723333in}{1.392903in}}%
\pgfpathlineto{\pgfqpoint{2.343333in}{1.392903in}}%
\pgfusepath{stroke}%
\end{pgfscope}%
\begin{pgfscope}%
\pgfpathrectangle{\pgfqpoint{1.000000in}{0.440000in}}{\pgfqpoint{6.200000in}{3.080000in}}%
\pgfusepath{clip}%
\pgfsetbuttcap%
\pgfsetroundjoin%
\pgfsetlinewidth{1.003750pt}%
\definecolor{currentstroke}{rgb}{1.000000,0.498039,0.054902}%
\pgfsetstrokecolor{currentstroke}%
\pgfsetdash{}{0pt}%
\pgfpathmoveto{\pgfqpoint{3.790000in}{1.302581in}}%
\pgfpathlineto{\pgfqpoint{4.410000in}{1.302581in}}%
\pgfusepath{stroke}%
\end{pgfscope}%
\begin{pgfscope}%
\pgfpathrectangle{\pgfqpoint{1.000000in}{0.440000in}}{\pgfqpoint{6.200000in}{3.080000in}}%
\pgfusepath{clip}%
\pgfsetbuttcap%
\pgfsetroundjoin%
\pgfsetlinewidth{1.003750pt}%
\definecolor{currentstroke}{rgb}{1.000000,0.498039,0.054902}%
\pgfsetstrokecolor{currentstroke}%
\pgfsetdash{}{0pt}%
\pgfpathmoveto{\pgfqpoint{5.856667in}{1.528387in}}%
\pgfpathlineto{\pgfqpoint{6.476667in}{1.528387in}}%
\pgfusepath{stroke}%
\end{pgfscope}%
\begin{pgfscope}%
\pgfsetrectcap%
\pgfsetmiterjoin%
\pgfsetlinewidth{0.803000pt}%
\definecolor{currentstroke}{rgb}{0.000000,0.000000,0.000000}%
\pgfsetstrokecolor{currentstroke}%
\pgfsetdash{}{0pt}%
\pgfpathmoveto{\pgfqpoint{1.000000in}{0.440000in}}%
\pgfpathlineto{\pgfqpoint{1.000000in}{3.520000in}}%
\pgfusepath{stroke}%
\end{pgfscope}%
\begin{pgfscope}%
\pgfsetrectcap%
\pgfsetmiterjoin%
\pgfsetlinewidth{0.803000pt}%
\definecolor{currentstroke}{rgb}{0.000000,0.000000,0.000000}%
\pgfsetstrokecolor{currentstroke}%
\pgfsetdash{}{0pt}%
\pgfpathmoveto{\pgfqpoint{7.200000in}{0.440000in}}%
\pgfpathlineto{\pgfqpoint{7.200000in}{3.520000in}}%
\pgfusepath{stroke}%
\end{pgfscope}%
\begin{pgfscope}%
\pgfsetrectcap%
\pgfsetmiterjoin%
\pgfsetlinewidth{0.803000pt}%
\definecolor{currentstroke}{rgb}{0.000000,0.000000,0.000000}%
\pgfsetstrokecolor{currentstroke}%
\pgfsetdash{}{0pt}%
\pgfpathmoveto{\pgfqpoint{1.000000in}{0.440000in}}%
\pgfpathlineto{\pgfqpoint{7.200000in}{0.440000in}}%
\pgfusepath{stroke}%
\end{pgfscope}%
\begin{pgfscope}%
\pgfsetrectcap%
\pgfsetmiterjoin%
\pgfsetlinewidth{0.803000pt}%
\definecolor{currentstroke}{rgb}{0.000000,0.000000,0.000000}%
\pgfsetstrokecolor{currentstroke}%
\pgfsetdash{}{0pt}%
\pgfpathmoveto{\pgfqpoint{1.000000in}{3.520000in}}%
\pgfpathlineto{\pgfqpoint{7.200000in}{3.520000in}}%
\pgfusepath{stroke}%
\end{pgfscope}%
\begin{pgfscope}%
\definecolor{textcolor}{rgb}{0.000000,0.000000,0.000000}%
\pgfsetstrokecolor{textcolor}%
\pgfsetfillcolor{textcolor}%
\pgftext[x=4.100000in,y=3.603333in,,base]{\color{textcolor}{\rmfamily\fontsize{12.000000}{14.400000}\selectfont\catcode`\^=\active\def^{\ifmmode\sp\else\^{}\fi}\catcode`\%=\active\def%{\%}CVRP – Vehicle crossings comparison}}%
\end{pgfscope}%
\end{pgfpicture}%
\makeatother%
\endgroup%